\documentclass[10pt,twocolumn,letterpaper]{article}

\usepackage[pagenumbers]{iccv} % To force page numbers, e.g. for an arXiv version

\usepackage[dvipsnames]{xcolor}

\usepackage{bm}
\usepackage{float}
\usepackage{tcolorbox}
\usepackage{multirow}
\usepackage{colortbl}
\usepackage{tabulary}
\usepackage{etoolbox}
\usepackage{pifont}
\usepackage{makecell}
\usepackage{graphicx}
\usepackage{adjustbox}
\usepackage{amsmath}
\usepackage{amssymb}
\usepackage{booktabs}
\usepackage{rotating}
\usepackage{tablefootnote}
\definecolor{LightCyan}{rgb}{0.8,1,1}
\definecolor{LightGreen}{rgb}{0.6,0.9,0.7}
\definecolor{royalblue(traditional)}{rgb}{0.0, 0.14, 0.4}
\definecolor{royalblue(web)}{rgb}{0.25, 0.41, 0.88}
\definecolor{darkgreen}{rgb}{0.0, 0.5, 0.0} % Define dark green
\definecolor{darkred}{rgb}{0.5, 0.0, 0.0}
\definecolor{darkblue}{rgb}{0.0, 0.0, 0.8}
\usepackage{subcaption}
\usepackage[accsupp]{axessibility}  % Improves PDF readability for those with disabilities.

\usepackage{tcolorbox}
\usepackage{xcolor}
\usepackage{pgfplots}
\usepackage{algorithm}
\usepackage{algpseudocode}
\usepackage{pgfplotstable}
\usepackage{filecontents}
\usepackage{mdframed}
\pgfplotsset{compat=1.17}
\usetikzlibrary{backgrounds,datavisualization.formats.functions}

\newcommand{\cmark}{\textcolor{darkgreen}{\ding{51}}}%
\newcommand{\xmark}{\textcolor{red}{\ding{55}}}%

\newcommand{\ftpg}{FedTPG\xspace}
\newcommand{\method}{FedMVP\xspace}
\newcommand{\methodlong}{Federated Multimodal Visual Prompt Tuning\xspace}
\newcommand{\methodbolded}{\textbf{Fed}erated \textbf{M}ultimodal \textbf{V}isual \textbf{P}rompt Tuning\xspace}
\newcommand{\pf}{PromptFormer\xspace}
\newcommand{\sm}{Supp. Mat.\xspace}
\newcommand{\bton}{base-to-new generalization\xspace}
\newcommand{\btoncap}{Base-to-New Generalization\xspace}

\newcommand{\promptfl}{PromptFL\xspace}
\newcommand{\fedtpg}{FedTPG\xspace}
\newcommand{\fedcocoop}{FedCoCoOp\xspace}

\newcommand{\fedclip}{FedCLIP\xspace}

\newcommand{\vect}[1]{\mbox{{\boldmath $#1$}}}

\newcommand{\vectrm}[1]{\bm{\mathrm{#1}}}

\definecolor{iccvblue}{rgb}{0.21,0.49,0.74}
\usepackage[pagebackref,breaklinks,colorlinks,allcolors=iccvblue]{hyperref}

\def\paperID{****} % *** Enter the Paper ID here
\def\confName{ICCV}
\def\confYear{2025}

\title{FedMVP: Federated Multimodal Visual Prompt Tuning for Vision-Language Models}

\author{\hspace{0.5cm} Mainak Singha$^{1}$ \and Subhankar Roy$^{2}$ \and Sarthak Mehrotra$^{3}$ \and Ankit Jha$^{4}$ \hspace{0.5cm} \and \hspace{2.0cm} Moloud Abdar$^{5}$ \and Biplab Banerjee$^{3}$ \and Elisa Ricci$^{1,6}$ \hspace{2.0cm} \and
\small$^{1}$ University of Trento, Italy \and \small$^{2}$ University of Bergamo, Italy \and \small$^{3}$ Indian Institute of Technology Bombay, India \and \small$^{4}$ LNMIIT Jaipur, India \and \small$^{5}$ The University of Queensland, Australia  \and \small$^{6}$ Fondazione Bruno Kessler, Italy}

\begin{document}
\maketitle
\begin{abstract}
In federated learning, textual prompt tuning adapts Vision-Language Models (e.g., CLIP) by tuning lightweight input tokens (or prompts) on local client data, while keeping network weights frozen. After training, only the prompts are shared by the clients with the central server for aggregation. However, textual prompt tuning suffers from overfitting to known concepts, limiting its generalizability to unseen concepts. To address this limitation, we propose \textbf{M}ultimodal \textbf{V}isual \textbf{P}rompt Tuning (FedMVP) that conditions the prompts on multimodal contextual information -- derived from the input image and textual attribute features of a class. At the core of FedMVP is a PromptFormer module that synergistically aligns textual and visual features through a cross-attention mechanism. The dynamically generated multimodal visual prompts are then input to the frozen vision encoder of CLIP, and trained with a combination of CLIP similarity loss and a consistency loss. Extensive evaluation on 20 datasets, spanning three generalization settings, demonstrates that FedMVP not only preserves performance on in-distribution classes and domains, but also displays higher generalizability to unseen classes and domains, surpassing state-of-the-art methods by a notable margin of $+1.57\%-2.26\%$. Code is available at \url{https://github.com/mainaksingha01/FedMVP}.

\end{abstract} 
\vspace{-0.5cm}
\section{Introduction}
\label{sec:intro}

\begin{figure}[!t]
\centering
    \includegraphics[width=\linewidth]{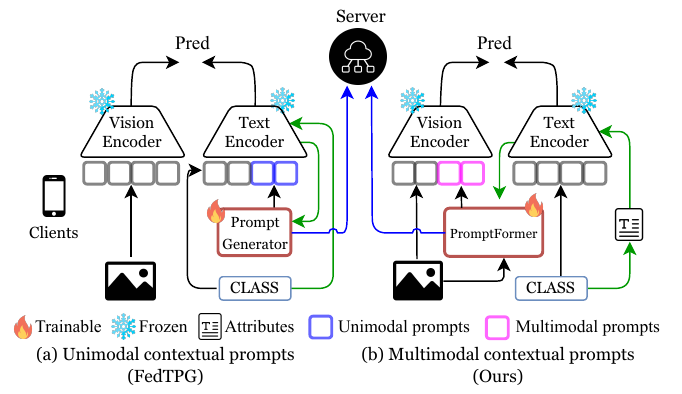}
    \vspace{-0.5cm}\caption{\textbf{Comparison of federated prompt tuning methods}: (a) a textual prompt tuning method (\eg, \fedtpg~\cite{fedtpg}) encodes \textbf{unimodal} contextual information, \ie, class names, in the learnable prompts, (b) our proposed method \method injects \textbf{multimodal} contextual information -- image and attributes -- into the prompts.}
    \label{fig:teaser}
    \vspace{-0.6cm}  
\end{figure}

A successful learning paradigm involves training neural networks on large and centralized datasets, enabling models to learn robust and generalizable representations~\cite{radford2019language,zhai2023sigmoid}. However, practical constraints often prevent centralized training (\eg, privacy regulations) creating a critical need for decentralized training frameworks that achieve high performance without central data access. As a solution, federated learning (FL) \cite{fedavg} enables multiple clients to collaboratively train a global model without requiring direct data sharing. In FL, clients train models locally and transmit only model updates (\textit{e.g.}, gradients or parameters) to a central server, which aggregates these updates to improve the global model. The research in FL is primarily aimed at: reducing communication rounds between clients and server~\cite{fedcom1,fedcom2}, faster and lightweight training algorithms on clients~\cite{fedfast,lightsecagg} and improving generalization while exhibiting high data heterogeneity across clients~\cite{fedgh,fedunder}.

Pretrained Vision-Language Models (VLMs) (\eg, CLIP~\cite{radford2019language}), have emerged as promising candidates for FL due to their strong generalization capabilities. However, the high communication overhead and computational demands of parameter-heavy VLMs pose challenges for FL, especially in bandwidth- or compute-constrained environments (\eg, robots). To make CLIP viable for the FL setting, prompt tuning (\eg, CoOp~\cite{coop}, VPT~\cite{vpt}) has been adopted as a preferred technique that appends lightweight learnable tokens (or \textit{prompts}) to the input tokens of the CLIP image and/or text encoder. Each client trains the prompts on the local data, while keeping the CLIP backbone frozen, and then dispatches these prompts to the server, where they are updated through aggregation and sent back. This speeds up client training as each client needs access to only a few samples per class (\eg, 16/class~\cite{coop}), and reduces communication overhead as only the lightweight prompts (\eg, 0.37\% of network parameters) need to be communicated, instead of all the parameters.

Although prompt tuning has proven to be effective in the FL setting, it suffers from reduced generalization to unseen classes and unseen domains -- a typical heterogeneous setup native to non independent and identically distributed (non-i.i.d.) FL~\cite{fedtpg} -- due to severe overfitting on the local training data~\cite{kgcoop}. It is attributed to the fact that prompts learn a \textit{static context} from a set of seen classes, which being fixed once learned, fail to capture the generalizable elements spread across clients. To circumvent the issue of reduced generalization, recent CLIP-based FL methods condition the prompts either on the class information, as in FedTPG~\cite{fedtpg} (see Fig.~\ref{fig:teaser}a), or on visual information, as in FedCoCoOp~\cite{cocoop}. This auxiliary conditioning on the input, or \textit{contextual information}, makes the prompts \textit{dynamic} and thus offers a better generalization to unseen distributions.

In this work, we argue that the FL setup being highly heterogeneous, where clients have data from disjoint classes and domains~\cite{fedtpg}, it is not sufficient to solely exploit either of the two modalities for contextual information. Rather, FL demands more comprehensive contextual information that encompasses all the available modalities to condition input prompts. To this end, we propose \methodbolded (\method) that conditions the prompts on \textit{multimodal contextual information}, which comprise: (i) \textit{visual} features of the input image itself, and (ii) \textit{textual} features of attributes of a class (\eg, ``two legs'', ``beak'', etc. for the class ``hen''). The \textbf{key intuition} is that (ii) promotes learning transferable prompts for unseen classes that share attributes similar to seen classes (\eg, unseen class ``seagull'' will share some attributes with the seen class ``hen''), and (i) promotes transferability for unseen classes and domains where attributes cannot be described via text (\eg, texture or abstract concepts). In a nutshell, the dual conditioning offers richer contextual information to the prompts, which helps in generalizing to classes or domains that share similar properties.

\begin{table}[!t]
    \centering 
    \caption{\textbf{Summary of federated prompt-tuning methods}. (\xmark) indicates the learnable prompts are initialized randomly, (\cmark) signifies learnable prompts are conditioned on contextual information (\textit{e.g.}, textual or visual inputs). \textit{Prompting} column denotes the CLIP encoder wherein learnable prompts are given as input (\eg, ``Textual'' means the text encoder of CLIP takes prompts as input).}
    
    \scalebox{0.8}{
    \begin{tabular}{lccc}\hline
        \multirow{2}{*}{Method} & \multicolumn{2}{c}{Contextual information} & \multirow{ 2}{*}{Prompting} \\
        \cline{2-3}
         & Textual & Visual & \\
         \hline
         FedKgCoOp~\cite{kgcoop} & \xmark & \xmark & Textual\\
         FedVPT~\cite{vpt} & \xmark & \xmark & Visual \\
         FedTPG~\cite{fedtpg} & \cmark & \xmark & Textual\\
         FedCoCoOp~\cite{cocoop} & \xmark & \cmark & Textual \\
         FedMaPLe~\cite{maple} & \xmark & \xmark & Multimodal \\
         FedMVP (Ours) & \cmark & \cmark & Visual \\
         \hline
    \end{tabular}
    }
    \vspace{-2mm}
    \label{tab:summary}
\end{table}

In detail, the proposed \method framework contains a \textbf{\pf} network (see Fig.~\ref{fig:teaser}b) that synergistically merges the information from visual and textual modalities with cross-attention mechanism~\cite{transformer} to generate the prompts. The resulting \textit{multimodal prompts} are then injected at the visual space of the frozen vision encoder of CLIP, unlike the existing methods~\cite{promptfl,fedtpg,kgcoop} that inject prompts through text encoder. The key differences between our \method and existing CLIP-based FL methods have been summarized in Tab.~\ref{tab:summary}. The \pf is trained in an end-to-end manner with the CLIP loss and a consistency loss. Once trained, each participating client shares the lightweight \pf network parameters with the server for aggregation through FedAvg~\cite{fedavg}, and thus enabling knowledge sharing among clients. During inference, the cross-attention learned by the \pf enables the model to dynamically adjust the prompts, enhancing generalizability to unseen classes and domains. We evaluated \method on three generalization settings encompassing 20 datasets that measure generalization to unseen classes, domains, and combinations of them. Experiments demonstrate that \method consistently outperforms the state-of-the-art CLIP-based FL methods by $1.57\%-2.26\%$.

In summary, our main \textbf{contributions} are: (\textbf{i}) We highlight the importance of leveraging \textbf{multimodal contextual information} in prompt tuning based FL setup in order to alleviate the issue of reduced generalizability to unseen data. (\textbf{ii}) We propose \textbf{\method} that synergistically combines the visual features of the image and textual features of the attributes of classes to generate \textbf{multimodal} prompts. This provides richer contextual information to the model and higher transferability to unseen classes and domains.
\section{Related Work}
\label{sec:related works}
\noindent \textbf{Prompt tuning of VLMs.} VLMs, such as CLIP \cite{radford2019language} and ALIGN \cite{align}, leverage large image-text datasets to enable multimodal understanding, performing well in zero-shot classification tasks. However, fully fine-tuning VLMs for downstream tasks without sacrificing their ability to generalize to unseen concepts remains a challenge \cite{wortsman2022robust}.

Prompt tuning approaches~\cite{vpt,coop,maple} have been proposed for parameter efficient and effective adaptation of CLIP by introducing a set of continuous learnable tokens (or prompts) onto the frozen CLIP backbone. These approaches can be broadly categorized into three categories depending on the manner in which CLIP is prompted: (i) \textit{textual prompting} that inject prompts in the input space of the text encoder~\cite{coop,kgcoop}, (ii) \textit{visual prompting} that injects prompts in the input space of the vision encoder~\cite{vpt}, and (iii) \textit{multimodal prompting} that injects prompts in both the text and vision encoder~\cite{xing2023dual,maple}.

In particular, MaPLe~\cite{maple}, which injects prompts to both the text and vision encoders, is not suitable for memory and compute efficient FL since it requires parts of both the encoders to be unlocked during tuning, risking overfitting. Moreover, it does not allow for image-conditioned dynamic prompting, which has shown to be effective for generalization~\cite{cocoop}. In contrast, our proposed \method enables multimodal prompting by involving only the vision encoder and supports instance-specific prompting.

\noindent \textbf{Federated prompt tuning of VLMs.} Prompt tuning has emerged as an effective technique for FL of VLMs since it allows clients to collaboratively train on a heterogeneously distributed dataset and adapt faster with fewer samples. The fact that the prompts are lightweight lends well to the FL setup as it reduces communication overhead. Due to the benefits offered by prompt tuning, several works~\cite{fedtpg,promptfl,fedcola,diprompt} have specifically adapted prompt tuning approaches to the FL setup. For example, PromptFL~\cite{promptfl} extended CoOp to FL via sharing prompts with the server, enhancing the generalization performance of client-server interactions in vision tasks. FedCLIP \cite{fedclip} learns a lightweight adapter in each client, similar to the CLIP-Adapter \cite{clip-adapter}, and shares parameters with the server. FedOTP \cite{fedotp} introduced global and local prompts, optimizing shared parameters through optimal transport, while DiPrompt \cite{diprompt} studied multi-domain learning by allowing multiple domains per client.

Unlike existing methods, a couple of recent works~\cite{fedtpg,xing2023dual} argue that prompt tuning is prone to overfitting to the seen classes due to learning static context. The closest work to ours is FedTPG \cite{fedtpg}, where a prompt generation network is learned across multiple clients. Additionally, this network leverages textual inputs, enhancing its ability to generalize to unseen classes. Our work significantly differs from previous methods in two key ways: (i) we exploit both attribute information derived from an LLM and image features to condition the prompts, and (ii) we leverage multimodal prompts from the proposed PromptFormer for visual prompt tuning, instead of textual prompt tuning.

\noindent\textbf{Classification via textual descriptions.} Zero-shot accuracy of CLIP on downstream tasks is sensitive to the quality of text prompts~\cite{radford2019language}. In detail, some works rely on simple hand-crafted prompts (\eg, ``a photo of a [\texttt{CLS}]'')~\cite{radford2019language} or others augment the hand-crafted prompts with richer attributes obtained from LLMs~\cite{desc,pratt2023does} (\eg, a ``hen'' has \textit{two legs}, \textit{white feathers}, etc.). There also exists a family of methods, such as LaCLIP~\cite{laclip}, LaBo~\cite{labo} and VFC~\cite{momeni2023verbs}, that have proposed to further refine CLIP weights with enriched captions from LLMs for improved performance. Recognizing the importance of class attributes for improved generalization to unseen concepts and domains -- an important desideratum in the FL set up -- we, for the first time in FL, introduce LLM generated class attributes for enriching the image-conditioned multimodal prompts through the \pf network.
\section{Methods}
\label{methodology}

We introduce FL setup in Sec.~\ref{sec:setup}, preliminaries on prompt tuning in Sec.~\ref{sec:prelim} and our proposed solution in Sec.~\ref{sec:proposedmethod}.

\begin{figure*}
\centering
    \includegraphics[width=1.0\linewidth]{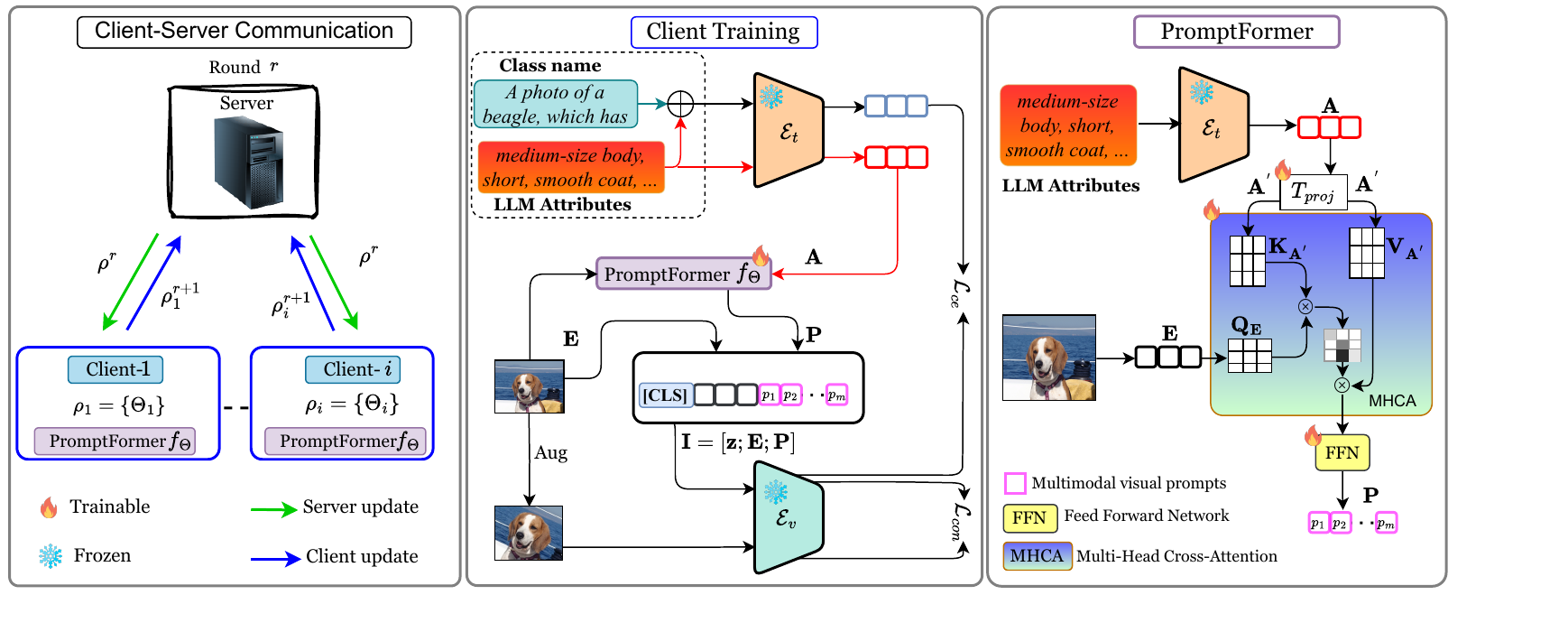}
    \vspace{-1.1cm}  
    \caption{\textbf{Overview of Federated Multimodal Visual Prompt Tuning} (\textbf{\method}). Each client $i$ trains the parameters of the \pf network $f_\Theta$ and multi-head self-attention on its local data. At each round, the server aggregates the trainable parameters from the clients and sends the updated parameters back to the clients. The \pf uses cross-attention mechanism to fuse the visual and textual attribute feature to generate multimodal visual prompts $\vectrm{P}$, which are then used for visual prompt tuning. }
    \label{fig:pipeline}
    \vspace{-0.5cm}  
\end{figure*}

\subsection{Problem formulation}
\label{sec:setup}
We consider a FL setup in which multiple remote clients train models on local data, and a central server aggregates models from the clients. Formally, let each remote client $i$ has access to a dataset $\mathcal{D}_i = \{\vectrm{x}_{i,j}, y_{i,j}\}^{n_i}_{j=1}$, where $\vectrm{x}_{i,j}$ is the $j\textsuperscript{th}$ image from the $i\textsuperscript{th}$ client, $y_{i,j}$ is the corresponding class label and $n_i$ is the number of samples in client $i$. Let the set of text class names in client $i$ be denoted as $\mathcal{C}_i = \{c_{i,k}\}^{K_i}_{k=1}$, where $K_i$ is the total number of classes.  As in \ftpg~\cite{fedtpg}, we follow a non-IID FL setup, where the samples in each client are drawn from a disjoint set of classes, \ie, $\mathcal{C}_i \cap \mathcal{C}_l = \emptyset$. As an example, the training data on each mobile device usually depend on the user, which may not be representative of the general population~\cite{fedavg}.

In particular, we investigate two kinds of non-IID FL setups, where, in addition to the in-distribution performance, we also evaluate the generalization ability of the aggregated server model on (i) unseen \textit{classes}, and (ii) unseen \textit{domains} or \textit{datasets}. The need for generalization to unseen classes and domains makes the non-IID FL setup more challenging than the traditional FL setting~\cite{kairouz2021advances}.

\subsection{Preliminaries on Prompt Tuning}
\label{sec:prelim}
Textual prompt tuning (TPT), \ie, CoOp~\cite{coop}, has emerged as an efficient technique in CLIP-based FL that keeps the CLIP backbone frozen and locally learns lightweight prompt vectors on each client. The locally learned prompts are then communicated to the server, as in PromptFL~\cite{promptfl}. Specifically, the hand-crafted text prompts (\eg, \textit{``a photo of a} \texttt{[CLS]}") input to the text encoder $\mathcal{E}_t$ is augmented with the learnable soft prompts $v_1, v_2, \dots, v_m$. However, as shown in~\cite{cocoop}, TPT can often lead to overfitting and reduced generalization due to the static context learned on the seen classes. This can be exacerbated in the non-IID FL setup due to the high data heterogeneity among the clients~\cite{fedtpg}.

Alternatively, visual prompt tuning (VPT)~\cite{vpt} adapts pre-trained vision encoder \( \mathcal{E}_v \) by injecting small number of visual learnable parameters into ViT's input space. More formally, let an input image $\vectrm{x}$ be divided in $b$ patches, yielding a collection of patch embeddings (augmented with positional encodings) at the input of the vision encoder as \(\bm{\mathrm{E}} \in \mathbb{R}^{b \times d_v} \), where $d_v$ is the embedding dimension. A [\texttt{CLS}] token $\vectrm{z}$ is concatenated with \( \vectrm{E} \), encapsulating the initial input representation for the ViT layers of \( \mathcal{E}_v \). VPT further augments the input by concatenating a set of randomly initialized learnable visual prompts \( \vectrm{P} = \{p_1, p_2, \dots, p_m\}\), where \( m \) is the prompt length. The first layer input of \( \mathcal{E}_v \) is then redefined as \( \vectrm{I} = [\vectrm{z}; \vectrm{E}; \vectrm{P}] \in \mathbb{R}^{(1 + b + m) \times d_v} \),  where “;” means concatenation. Although VPT enables more nuanced understanding of the visual space, it suffers from the same issues as TPT, \ie, limited generalization (see Sec.~\ref{sec:experiments}).

\subsection{\methodlong}
\label{sec:proposedmethod}

Our proposed method \method deviates from the TPT and VPT paradigms and incorporates \textbf{multimodal contextual information} from both the visual and textual modalities to improve generalization in the FL setup. As shown in Fig.~\ref{fig:pipeline}, at the core of our novel architecture is a prompt generator, called \textbf{\pf}, which conditions the visual prompts $\vectrm{P}$ on the task-related information: input image and text attributes corresponding to the class names, rather than randomly initializing them. This makes $\vectrm{P}$ \textit{context-aware}, generalizing better to unseen classes and domains. Once trained on each client, only the parameters of the lightweight PromptFormer network are communicated to the server, thus, keeping the communication overhead low. Next, we describe PromptFormer (Sec.~\ref{visualprompting}), training objectives (Sec.~\ref{training}) and how we conduct client-side training and server-side aggregation (Sec.~\ref{sec:client-server}).

\subsubsection{Visual Prompt Generation with PromptFormer}
 \label{visualprompting} 
We introduce the \pf network \( f_\Theta \), parameterized by $\Theta$, designed to generate contextually rich visual tokens (or prompts) for each client $i$ by leveraging both attribute features and visual patch embeddings (see Fig.~\ref{fig:pipeline}). Concretely, we use the CLIP text encoder $\mathcal{E}_t$ to extract embeddings  $\vectrm{A}_i = \{\mathcal{E}_t (\texttt{LLM}(c_k))\}^K_{k=1}$ of the text attributes\footnote{For example, for the class name ``giraffe'' the LLM outputs ``Exceptionally long neck, unique coat pattern with irregular brown patches, ...''} $\mathcal{A}_i$ corresponding to the text class names $\mathcal{C}_i$, generated using a large lanuage model (LLM)~\cite{gpt4} (see Sec. A.1 of Supp. Mat. for Attribute Generation process). Going forward, we drop the client index $i$ for notational clarity. The \pf takes as input $b$ visual patch embeddings $\vectrm{E}$, corresponding to an image $\vect{x}$, and the text embeddings $\vectrm{A}$ to generate a set of enriched $m$ multimodal visual prompts $\vectrm{P} \in \mathrm{R}^{m \times d_v}$:
\vspace{-0.1cm}
\begin{equation}
\label{eqn:attr-vis-prompts}
    \vectrm{P} = \{p_j\}^m_{j=1} = f_\Theta(\vectrm{A}, \vectrm{E}).
\end{equation}

Consequently, the input to the first layer of the visual encoder \( \mathcal{E}_v \) is redefined as \( [\vectrm{z}; \vectrm{E}; \vectrm{P}] \in \mathbb{R}^{(1 + b + m) \times d_v} \). Intuitively, multimodal visual prompts carry richer contextual information, and helps the network in generalizing to  unseen classes and domains that share commonalities.

\vspace{-0.2cm}
\paragraph{Architecture of \pf.} The \pf architecture comprises: lightweight multi-head cross-attention (MHCA) modules, feed-forward network (FFN) and a linear projection layer; transforming multimodal features into comprehensive visual prompts. In detail, the visual patch embedding \( \vectrm{E} \) interacts with the linearly projected attribute features \(\vectrm{A}' = T_\text{proj}(\vectrm{A}) \) to construct multimodal visual prompts as: 
\begin{equation}
\label{eqn:ca}
\begin{array}{c}
\vectrm{P}(\vectrm{A}', \vectrm{E}) = \text{FFN}(\text{CrossAttention}(\vectrm{Q}_{\vectrm{E}}, \vectrm{K}_{\vectrm{A}^{'}}, \vectrm{V}_{\vectrm{A}^{'}})); \\[0.5em]
\text{with} \; \vectrm{Q}_{\vectrm{E}} = \vectrm{E} W_{\vectrm{Q}}, \vectrm{K}_{\vectrm{A}^{'}} = \vectrm{A}' W_{\vectrm{K}}, \vectrm{V}_{\vectrm{A}^{'}} = \vectrm{A}' W_{\vectrm{V}}, 
\end{array}
\end{equation}
where, $\vectrm{E}$ are transformed into query vectors $\vectrm{Q}_{\vectrm{E}}$ using query matrices $W_{\vectrm{Q}}$; and the projected attribute features $\vectrm{A}'$ are transformed into key and value vectors $\vectrm{K}_{\vectrm{A}'}$ and  $\vectrm{V}_{\vectrm{A}'}$, using key and value matrices $W_{\vectrm{K}}$ and $W_{\vectrm{V}}$, respectively. 

Intuitively, through the cross-attention mechanism, visual features learn to attend to the relevant parts of the corresponding attribute features. For example, the patches depicting the \textit{legs} of the class ``dog'' will learn to attend to the attributes \textit{four legged}. When an unseen class (\eg, ``giraffe'') comprises an animal with \textit{four legs}, the final prompts $\vectrm{P}$ will contain this relevant information derived from the attributes \textit{four legged}. Additional details of \pf architecture are reported in Sec. A.3 of the \sm

\subsubsection{Training of \method} 
\label{training}
 
After the visual prompts are obtained with \pf as per Eq.~\ref{eqn:ca}, we concatenate the [\texttt{CLS}] token, the patch embeddings, and the visual prompts to get $\vectrm{I} = [\vectrm{z}; \vectrm{E}; \vectrm{P}]$ following Sec.~\ref{sec:prelim}, and feed it to the visual encoder $\mathcal{E}_v$. 

The parameters of \pf \( \rho = \{\Theta\} \) are the only trainable parameters of \method, and the rest of the weights are kept frozen. To train the model, we employ the CLIP cross-entropy loss~\cite{radford2019language}, formulated as:
\vspace{-0.1cm}  
\begin{equation}
\begin{split}
    \mathcal{L}_\text{ce} = - \underset{(\vectrm{x},y) \sim \mathcal{D}}{\mathbb{E}} y\log p(y | \vectrm{I}),
\end{split}
\label{loss}
\end{equation}
\noindent where \( y \) denotes the ground truth label, and the prediction probability of CLIP between the visual features $\vectrm{v} = \mathcal{E}_v(\vectrm{I})$ and text features $\vectrm{t}_k = \mathcal{E}_t([\text{A photo of [CLASS]} ; \texttt{LLM}(c_k)])$ for class $k$ is computed as:
\vspace{-0.25cm}  
\begin{equation}
    p(y=k|\vectrm{I}) = \frac{\exp(\text{cos}(\vectrm{v}, \vectrm{t}_k))/\tau)}{\sum^{K}_{k'=1} \exp(\text{cos}(\vectrm{v}, \vectrm{t}_{k'}))/\tau)},
\label{eqprob}
\end{equation}
where \( \text{cos}(\cdot, \cdot) \) as a similarity function, \( \tau \) as the temperature parameter, and $[;]$ denotes concatenation. Details on how $\vectrm{t}_k$ is computed are provided in Sec. A.1 of the \sm

In addition to the $\mathcal{L}_\text{ce}$, we employ a regularization constraint to ensure consistency between two augmented views of an image. We define the consistency loss \( \mathcal{L}_\text{con} \) as:
\vspace{-0.1mm}
\begin{equation}
    \mathcal{L}_\text{con} = 1 - \text{cos}(\mathcal{E}_{v}(\vectrm{I}), \mathcal{E}_{v}(\vectrm{x}')),
\label{loss_con}
\end{equation}
where $\vectrm{x}'$ is the augmented version of $\vectrm{x}$. Finally, the total loss function is given as:

\vspace{-0.7cm}
\begin{equation}
\begin{split}
    \mathcal{L}_\text{total} = \mathcal{L}_\text{ce} + \alpha \cdot \mathcal{L}_\text{con},
\end{split}
\label{eqn:toal_loss}
\end{equation}
\noindent where \( \alpha \) is a balancing coefficient for the regularization term, ensuring both alignment and consistency.
  
\subsubsection{Client-side training and server-side aggregation}
\label{sec:client-server}
In our approach, multiple remote clients participate in distinct image classification tasks, collaboratively training the PromptFormer network, parameterized by \( \rho = \{\Theta\} \). In each communication round, the client transmits $\rho$ to the server, which is aggregated at the server. Through training on a diverse set of classification tasks distributed across clients, and aggregating the parameters at the server, the resulting model achieves generalization capability to classes and domains unseen during training. Pseudo-code of \method can be found in Sec. A.2 of the \sm

Initially, the server randomly initializes the parameters \( \rho^{0} \) and in each communication round only a random subset of remote clients $\mathcal{S}^r$ will be chosen by the server for sharing the updated parameters $\rho$. Each communication round $r$ involves the following steps:

\noindent (i) \textbf{Model initialization:} Each client \( i \) in \( S_r \) receives the latest model parameters \( \rho^{r} \) to configure its local model.\\
\noindent (ii) \textbf{Local optimization:} Following Eq.~(\ref{eqn:toal_loss}), client \( i \) optimizes its local parameters $\rho^r$ using an SGD optimizer over \( J \) iterations with learning rate \( \eta^{r} \). Updated parameters:
\vspace{-1mm}
    \begin{equation}
    \begin{split}
        \rho^{r+1}_{i} = \texttt{SGD}_{J}(\eta^{r}, \rho^{r}, \mathcal{C}_i, \mathcal{D}_i).
    \end{split}
    \label{sgd_agg}
    \end{equation}

\noindent (iii) \textbf{Parameter aggregation:} After local training, each client \( i \) in \( S_r \) transmits its updated parameters \( \rho^{r+1}_{i} \) back to the server, where they are aggregated~\cite{fedavg} as:
    \begin{equation}
    \begin{split}
        \rho^{r+1} = \frac{1}{|S_r|} \sum_{i \in S_r} \rho^{r+1}_i.
    \end{split}
    \label{server_agg}
    \end{equation}

The above steps are repeated for \( R \) communication rounds, after which the server converges on the final model parameters \( \rho^{R} \). This iterative process facilitates collaborative and distributed optimization of \({\rho} \) across client data.

\noindent \textbf{Lightweight fine-tuning in clients}. As the FL setup exhibits high data heterogeneity, different clients possess varying numbers of samples. Thus, treating every client as equal and repeatedly fine-tuning the parameters \( \rho \) can lead to severe overfitting in some clients, especially those with very limited training data. To prevent this, we propose to inject low-rank decomposition matrices, or LoRA~\cite{lora}, to the \pf module, and fine-tune these LoRA matrices if training loss $\mathcal{L}_\text{total}$ in a client starts below a certain loss value $\sigma$, which we set to 0.5. In particular, only the LoRA weights corresponding to the query, key and value matrices of the attention layers are trained, while freezing the parameters \( \rho \). In that case, the client shares only the LoRA matrices with the server. This not only helps in boosting generalization, but also brings down the communication overhead of clients by around $\times 267$.

\section{Experiments}
\label{sec:experiments}

We evaluate our proposed \method in three distinct settings that are aimed to gauge the (i) generalization to unseen \textit{categories} (or Base-to-New Generalization) in Sec.~\ref{sec:ID_DG}, (ii) generalization to unseen \textit{domains} (or Domain Generalization) in Sec.~\ref{sec:OOD_DG}, and (iii) generalization to unseen \textit{datasets} (or Cross-Dataset Generalization) in Sec.~\ref{sec:cross-dataset}. Finally, we report the results of our ablation study in Sec.~\ref{sec:ablation} that are aimed to assess impact of the design choices made in \method. Due to the lack of space, we report additional experimental results in Sec. B of the \sm

\noindent\textbf{Benchmarks and datasets.} Following FedTPG~\cite{fedtpg}, we have reported results on the Base-to-New Generalization setting that encompasses nine medium-sized image recognition datasets. For the Domain Generalization setting we have used the ImageNet~\cite{imagenet} and DomainBed~\cite{gulrajani2020search} benchmarks. In addition, we have also included EuroSAT~\cite{eurosat} for the Cross-Dataset Generalization setting, alongside all nine datasets of the Base-to-New Generalization setting. In total, we have used 20 datasets for a comprehensive experimental evaluation. Detailed dataset descriptions and statistics can be found in Sec. B.1 of the \sm

\noindent\textbf{Baselines.} We have compared our method with the state-of-the-art prompt tuning approaches that use the CLIP backbone~\cite{radford2019language}: KgCoOp \cite{kgcoop}, CoCoOp \cite{cocoop}, VPT~\cite{vpt}, and MaPLe \cite{maple}, which were originally designed for offline and non-FL setup, but adapted to the federated setting, as in FedTPG~\cite{fedtpg}. Furthermore, we have compared with CLIP-based methods specifically designed to work in the FL setting: FedTPG~\cite{fedtpg}, FedOTP~\cite{fedotp}, PromptFL~\cite{promptfl}, and FedCLIP~\cite{fedclip}. We also included zero-shot CLIP (ZS-CLIP).

\noindent\textbf{Implementation details.} We have employed the frozen ViT-B/16 CLIP backbone as the vision encoder, and GPT-4o \cite{gpt4} as the LLM in our experiments. Each MHCA module of $f_\Theta$, consists of a 4-head cross-attention layers, complemented by layer normalization and each FFN comprising of a two-layer bottleneck structure (Linear-GeLU-Linear). The query prompt $\vectrm{Q}$ has a length \( m = 4 \) and matches the patch embedding dimension \( d_v = 768 \). The projection layer $T_\text{proj}$ maps the textual dimension of $512$ to the patch embedding dimension via a linear transformation. In Eq.~(\ref{eqn:toal_loss}), we keep $\alpha=10$. We have set the batch size to 128, number of shots per class to 8, and the learning rate equal to 0.003 with a decay rate of $1\text{e-}5$ and momentum of 0.9.

\subsection{Base-to-New Generalization}
\label{sec:ID_DG}

\noindent\textbf{Experimental setup.} To recap, the \bton setting encompasses nine different datasets. The setting is characterized by splitting the classes from each dataset into two groups -- base and new. The samples of the base classes are used for training, and the samples of the new classes are held out to test the generalization performance~\cite{fedtpg}. Furthermore, the samples of bases classes from all nine datasets are distributed to the clients, such that each client does not see samples from more than 20 disjoint classes (referred to as \textit{local} classes). We report the model's classification accuracy on: (i) the client's local classes, (ii) all the base classes (spread across clients), and (iii) the held-out new classes that were never seen by any client. Note that (ii) and (iii) is computed using the aggregated model on the server, and (i) is computed using the model on the client (which is the same server model after convergence).
\begin{table}[t]

    \centering
    \caption{{\textbf{\btoncap setting.}} Accuracies (\%) on client's local (seen) classes, base (seen) classes and new (unseen) classes. Harmonic mean (HM) is computed with Base and New class accuracies. Best number in each column is in bold and second-best underlined. Improvement row reports the absolute difference between the best and second-best numbers.}
    \vspace{-.2cm}
    \scalebox{0.69}{
    \begin{tabular}{lccc|cc}
    \toprule
        
        \textbf{Methods} & {Local} & {Base} & {New} & {HM}\\ 
        \midrule
         
        ZS-CLIP \cite{radford2019language} &76.72	&70.51	&75.78	&\cellcolor[gray]{0.9}74.24 \\ 
        \midrule
        FedOTP \cite{fedotp} &74.82 &65.22 &57.04 &\cellcolor[gray]{0.9}64.89 \\
        FedCoCoOp \cite{cocoop} &81.46	&73.76	&66.00	&\cellcolor[gray]{0.9}73.20 \\
        FedVPT \cite{vpt} &76.29 &70.43 &74.89 &\cellcolor[gray]{0.9}73.79  \\
        FedCLIP \cite{fedclip} &76.87 &71.04 &75.06 &\cellcolor[gray]{0.9}74.24  \\
        FedMaPLe \cite{maple} &81.63	&74.44	&70.62	&\cellcolor[gray]{0.9}75.29 \\
        FedKgCoOp \cite{kgcoop} &78.38	&72.18	&75.87	&\cellcolor[gray]{0.9}75.39 \\
        PromptFL \cite{promptfl} &\underline{81.75}	& \underline{74.47}	&71.70	&\cellcolor[gray]{0.9}75.74 \\
        FedTPG \cite{fedtpg} &80.75	&73.68	& \underline{76.02}	&\cellcolor[gray]{0.9}\underline{76.70} \\

        \method (Ours) &\textbf{81.89}	&\textbf{75.37}	&\textbf{77.82}	&\cellcolor[gray]{0.9}\textbf{78.27} \\
        \midrule
        Improvement & \textcolor{darkgreen}{+\textbf{0.14}} & \textcolor{darkgreen}{+\textbf{0.90}} & \textcolor{darkgreen}{+\textbf{1.79}} & \cellcolor[gray]{0.9}\textcolor{darkgreen}{+\textbf{1.57}}\\

\bottomrule
    \end{tabular}}
    \label{tab:base2new}
    \vspace{-0.5cm}
\end{table}

\noindent\textbf{Main results.} In Table~\ref{tab:base2new} we report the performance in the \bton setting, summarized across nine datasets. The results show that the textual and multimodal prompting methods, such as PromptFL, FedKgCoOp, and FedMaPLe, excel on the seen local and base class accuracy, but have at-par or worse generalization on the new classes than the ZS-CLIP model. This hints at the potential overfitting caused by learning static context from the seen classes. In contrast, our proposed \method that considers multimodal contextual information and dynamic conditioning demonstrates superior performance in both the seen and unseen classes, achieving the highest HM accuracy. While \fedtpg also generates prompts based on task-related input, it does not outperform our \method due to lack of comprehensive visual and attribute information in the prompts. In summary, \method surpasses the state-of-the-art methods by a notable margin of $+1.57\%$ in the \bton setting. Detailed breakdown for each of the nine datasets is reported in Tab. A.4 in the \sm

\begin{table*}[!ht]
    \centering
    \caption{\textbf{Domain Generalization setting on the DomainBed~\cite{gulrajani2020search} benchmark.} Accuracies (\%) on the unseen target domain(s) for the Multi-source Single-target (MSST) and Single-source Multi-target (SSMT) settings. Best number in each column is in bold and second-best underlined. Improvement row reports the absolute difference between the best and second-best numbers.}
    \vspace{-0.2cm}
    \scalebox{0.65}{
    \begin{tabular}{lccccc|c||ccccc|c}
    \toprule
        & \multicolumn{6}{c}{\textbf{Multi-source Single-target}} & \multicolumn{6}{c}{\textbf{Single-source Multi-target}} \\ \cmidrule(lr){2-7} \cmidrule(lr){8-13}
        
        \multirow{-2}{*}{\textbf{Method}} & {PACS} & {OfficeHome} & {VLCS} & {Terra Inc.} & {DomainNet} & {Average} & {PACS} & {OfficeHome} & {VLCS} & {Terra Inc.} & {DomainNet} & {Average} \\ 
        \midrule

        ZS-CLIP \cite{radford2019language} &96.16	&81.49	&\underline{83.29}	&33.98	&57.13	&\cellcolor[gray]{0.9}70.41 &96.16	&81.49	&\underline{83.29}	&33.98	&57.13	&\cellcolor[gray]{0.9}70.41 \\ 
        \midrule
        
         FedOTP \cite{fedotp} &90.71	&76.42	&67.41	&13.24	&49.67	&\cellcolor[gray]{0.9}59.49 &91.17	&74.53	&65.16	&15.11	&39.37	&\cellcolor[gray]{0.9}57.07 \\

         FedCoCoOp \cite{cocoop} &85.06	&81.42	&61.73	&23.68	&57.08	&\cellcolor[gray]{0.9}61.79 &84.56	&50.87	&61.25	&23.07	&58.28	&\cellcolor[gray]{0.9}55.61 \\

         FedTPG \cite{fedtpg} &90.99	&\underline{82.78}	&69.77	&26.79	&56.82	&\cellcolor[gray]{0.9}65.43 &90.71	&\underline{82.60}	&67.63	&22.51	&\underline{58.34}	&\cellcolor[gray]{0.9}64.36 \\

         PromptFL \cite{promptfl} &95.37	&81.74	&74.87	&25.02	&56.87	&\cellcolor[gray]{0.9}66.77 &95.83	&81.72	&66.97	&24.17	&58.27	&\cellcolor[gray]{0.9}65.39 \\
         
        FedKgCoOp \cite{kgcoop} &95.42	&81.82	&74.90	&25.03	&56.92	&\cellcolor[gray]{0.9}66.82 &95.81	&81.70	&67.49	&24.20	&58.30	&\cellcolor[gray]{0.9}65.50 \\

        FedMaPLe \cite{maple}  &94.51	&82.03	&71.79	&36.30	&\underline{58.88}	&\cellcolor[gray]{0.9}68.70 &94.25	&81.48	&72.08	&33.59	&57.62	&\cellcolor[gray]{0.9}67.80 \\

        FedVPT \cite{vpt} &95.36	&81.76	&83.19	&33.62	&55.98	&\cellcolor[gray]{0.9}69.98 &94.79	&80.99	&82.44	&32.23	&55.66	&\cellcolor[gray]{0.9}69.22 \\

        FedCLIP \cite{fedclip} &\underline{96.29}	&81.74	&82.70	&\underline{36.58}	&57.85	&\cellcolor[gray]{0.9}\underline{71.03} &\underline{96.29}	&81.62	&81.81	&\underline{36.44}	&57.42	&\cellcolor[gray]{0.9}\underline{70.72} \\

        \method (Ours) &\textbf{97.28}	&\textbf{84.15}	&\textbf{85.12}	&\textbf{37.36}	&\textbf{61.17}	&\cellcolor[gray]{0.9}\textbf{73.02} &\textbf{97.17}	&\textbf{83.89}	&\textbf{84.61}	&\textbf{36.92}	&\textbf{60.56}	&\cellcolor[gray]{0.9}\textbf{72.63} \\

        \midrule
        Improvement &\textcolor{darkgreen}{+\textbf{0.99}}	&\textcolor{darkgreen}{+\textbf{1.37}}	&\textcolor{darkgreen}{+\textbf{1.83}}	&\textcolor{darkgreen}{+\textbf{0.78}}	&\textcolor{darkgreen}{+\textbf{2.29}}	&\cellcolor[gray]{0.9}\textcolor{darkgreen}{+\textbf{1.99}} &\textcolor{darkgreen}{+\textbf{0.88}}	&\textcolor{darkgreen}{+\textbf{1.29}}	&\textcolor{darkgreen}{+\textbf{1.32}}	&\textcolor{darkgreen}{+\textbf{0.48}}	&\textcolor{darkgreen}{+\textbf{2.22}}	&\cellcolor[gray]{0.9}\textcolor{darkgreen}{+\textbf{1.91}} \\

    \bottomrule
        
    \end{tabular}}
    \label{tab:combined}
\vspace{-0.5cm}  
\end{table*}

\begin{table}[t]

    \centering
    \caption{\textbf{Domain Generalization setting on the ImageNet benchmark.} Accuracies (\%) on the seen source domain ImageNet (IN) and unseen target domains: ImageNet-V2 (INV2), ImageNet-Sketch (IN-S), ImageNet-A (IN-A) and ImageNet-R (IN-R). Best number in each column is in bold and second-best underlined. Improvement row reports the absolute difference between the best and second-best numbers.}
    \vspace{-0.2cm}
    \scalebox{0.63}{
    \begin{tabular}{lccccc|c}
    \toprule
        & \multicolumn{1}{c}{\textbf{Source}} & \multicolumn{5}{c}{\textbf{Target}}  \\ \cmidrule(lr){2-2} \cmidrule(lr){3-7}
        
        \multirow{-2}{*}{\textbf{Method}} & {IN} & {INV2} & {IN-S} & {IN-A} & {IN-R} & {Average} \\ 
        \midrule
         ZS-CLIP \cite{radford2019language} &66.75 &60.79 &46.12 &47.79 &74.00 &\cellcolor[gray]{0.9}57.18 \\
         \midrule
          FedOTP \cite{fedotp} &51.68 &45.53 &34.73 &16.64 &63.98 &\cellcolor[gray]{0.9}40.22 \\

          FedMaPLe \cite{maple} &66.96 &60.65 &44.69 &46.24 &74.62 &\cellcolor[gray]{0.9}56.55 \\

          FedVPT \cite{vpt} &66.92 &60.34 &46.43 &48.03 &74.56 &\cellcolor[gray]{0.9}57.34 \\
          
          PromptFL \cite{promptfl} &67.80 &61.59 &45.61 &48.78 &74.49 &\cellcolor[gray]{0.9}57.62 \\

          FedCLIP \cite{fedclip} &67.26 &61.35 &46.66 &48.24 &74.36 &\cellcolor[gray]{0.9}57.65 \\
          
          FedKgCoOp \cite{kgcoop} &67.53 &61.60 &46.69 &48.37 &74.71 &\cellcolor[gray]{0.9}57.84 \\

          FedCoCoOp \cite{cocoop} &68.51 &62.29 &46.90 &\underline{50.33} &\underline{76.49} &\cellcolor[gray]{0.9}59.00 \\

          FedTPG \cite{fedtpg} &\underline{69.51} &\underline{62.90} &\underline{47.65} &49.97 &76.35 &\cellcolor[gray]{0.9}\underline{59.22} \\

        \method (Ours) &\textbf{70.87}	&\textbf{63.72}	&\textbf{50.93}	&\textbf{51.76}	&\textbf{77.23}	&\cellcolor[gray]{0.9}\textbf{60.91}\\

        \midrule
        Improvement &\textcolor{darkgreen}{+\textbf{1.36}} &\textcolor{darkgreen}{+\textbf{0.82}}	&\textcolor{darkgreen}{+\textbf{3.28}}	&\textcolor{darkgreen}{+\textbf{1.43}}	&\textcolor{darkgreen}{+\textbf{0.74}}	&\cellcolor[gray]{0.9}\textcolor{darkgreen}{+\textbf{1.69}}\\
\bottomrule

    \end{tabular}}
    \label{tab:DG_Imagenet}
\vspace{-0.5cm}  
\end{table}

\subsection{Domain Generalization}
\label{sec:OOD_DG}

\noindent\textbf{Experimental setup.} We conduct experiments under two distinct domain generalization (DG) setups: Multi-source Single-target (MSST) and Single-source Multi-target (SSMT). In the MSST DG setting for a given benchmark, we follow a leave-one-domain-out protocol for evaluation, where we hold out one domain for measuring generalization and use the rest of the domains for training the clients. Training classes are split among clients under the constraint that each client can see images from a single domain. This setting mirrors a practical scenario where images taken with a smartphone uniquely reflect the taste of the user or socio-economic biases. In a similar manner, in the SSMT DG setting, we use one domain for training the clients in a distributed manner, while the rest of the domains are used for evaluation. More details on this experimental setup can be found in Sec. B.4 of the \sm

\noindent\textbf{Main results.} In Table~\ref{tab:combined}, we report the performance of \method and its competitors in the MSST and SSMT DG settings. In particular, SSMT DG is a more challenging setting than the MSST DG setting, as the clients are trained only on a single domain, and expected to generalize to multiple unseen domains. We observe that ZS-CLIP is a strong baseline that outperforms all textual and multimodal prompting methods, with the exception of \fedclip and our proposed \method. This suggests that suboptimal prompt tuning can lead to overfitting on the source training domains, thus hurting generalization. Our \method owing to the use of contextual information learns more transferable representations that generalize better to unseen target domains, thus outperforming all competitor methods in most of the benchmarks. More detailed results in Sec. B.4 of the \sm

In Table.~\ref{tab:DG_Imagenet} we report the results in the SSMT DG setting on ImageNet benchmark. We observe that \fedtpg and \fedcocoop come close to our \method in terms of average performance, with \method outperforming all baselines by $+1.69\%$. In particular, we believe \method improves the performance on ImageNet-Sketch (IN-S) by a large margin of $+3.28\%$ due to the use of attribute information, which remains unchanged between real and sketch images.

\begin{table*}[!ht]

    \centering
    \caption{\textbf{Cross-Dataset Generalization setting.} Accuracies (\%) on the seen classes of the source dataset, \ie, ImageNet, and the unseen classes of the target datasets. Best number in each column is in bold and second-best underlined. Improvement row reports the absolute difference between the best and second-best numbers.}
    \vspace{-0.2cm}
    \scalebox{0.62}{
    \begin{tabular}{lccccccccccc|c}
    \toprule

       & \multicolumn{1}{c}{\textbf{Source}} & \multicolumn{11}{c}{\textbf{Target}}  \\ \cmidrule(lr){2-2} \cmidrule(lr){3-13}
        
       \multirow{-2}{*}{\textbf{Method}} & {ImageNet} & {Caltech101} & {Flowers102} & {FGVCAircraft} & {UCF101} & {OxfordPets} & {Food101} & {DTD} & {StanfordCars} & {SUN397} & {EuroSAT} & {Average} \\ 
       
        \midrule
        ZS-CLIP \cite{radford2019language} &66.75 &92.90 &\underline{71.29} &\underline{24.72} &\underline{66.75} &89.15 &\underline{86.09} &44.33 &65.29 &62.59 &47.68 &\cellcolor[gray]{0.9}65.08 \\
        \midrule
        
        FedOTP \cite{fedotp} &51.68	&88.60	&48.36	&10.95	&46.10	&73.43	&61.01	&41.55	&29.61	&48.34	&50.26	&\cellcolor[gray]{0.9}49.82 \\

        FedMaPLe \cite{maple} &66.96 &92.49 &68.25 &23.52 &60.32 &\underline{89.67} &83.52 &44.68 &60.16 &61.85 &45.38 &\cellcolor[gray]{0.9}62.98 \\

        PromptFL \cite{promptfl} &67.80 &91.87 &68.13 &21.44 &64.13 &88.70 &85.85 &42.43 &63.59 &62.77 &43.26 &\cellcolor[gray]{0.9}63.22 \\

        FedCLIP \cite{fedclip} &67.26	&93.39	&67.19	&24.03	&65.64	&88.28	&85.14	&44.44	&\underline{65.50}	&63.42	&41.67	&\cellcolor[gray]{0.9}63.87 \\

        FedKgCoOp \cite{kgcoop} &67.53 &93.63 &69.31 &23.06 &64.46 &88.55 &85.37 &44.74 &64.99 &63.85 &43.29 &\cellcolor[gray]{0.9}64.13 \\

        FedVPT \cite{vpt} &68.56	&91.42	&69.24	&21.09	&65.70	&87.79	&85.45	&44.57	&65.00	&64.68	&47.82	&\cellcolor[gray]{0.9}64.28 \\

        FedCoCoOp \cite{cocoop} &68.51 &94.11 &66.34 &20.79 &62.75 &89.04 &85.40 &43.20 &63.98 &64.02 &\textbf{55.40} &\cellcolor[gray]{0.9}64.50 \\

        FedTPG \cite{fedtpg} &\underline{69.51} &\underline{93.75} &70.04 &23.22 &64.72 &\textbf{90.60} &85.91 &\underline{46.25} &63.98 &\underline{66.78} &47.86 &\cellcolor[gray]{0.9}\underline{65.31} \\

        \method (Ours) &\textbf{70.87} 	&\textbf{95.37}	&\textbf{72.80}	&\textbf{25.94}	&\textbf{70.58}	&89.27	&\textbf{87.06}	&\textbf{49.78}	&\textbf{65.83}	&\textbf{68.19}	&\underline{50.84}	&\cellcolor[gray]{0.9}\textbf{67.57} \\

        \midrule
        Improvement &\textcolor{darkgreen}{+\textbf{1.36}} &\textcolor{darkgreen}{+\textbf{1.62}} &\textcolor{darkgreen}{+\textbf{1.51}}	&\textcolor{darkgreen}{+\textbf{1.22}}	&\textcolor{darkgreen}{+\textbf{3.83}}	&\textcolor{red}{-\textbf{1.33}}	
        &\textcolor{darkgreen}{+\textbf{0.97}}	&\textcolor{darkgreen}{+\textbf{3.53}}	&\textcolor{darkgreen}{+\textbf{0.33}}	
        &\textcolor{darkgreen}{+\textbf{1.41}}	&\textcolor{red}{-\textbf{4.56}}
        &\cellcolor[gray]{0.9}\textcolor{darkgreen}{+\textbf{2.26}} \\
\bottomrule

    \end{tabular}}
    \label{tab:CDT}
    \vspace{-0.5cm}  
\end{table*}
\vspace{-0.1cm}  
\subsection{Cross-dataset Generalization} 
\label{sec:cross-dataset}
\vspace{-0.15cm}  
\noindent\textbf{Experimental setup.} In the Cross-Dataset Generalization setting the classes from ImageNet are split among the clients for training, in a disjoint manner, and the resulting network is evaluated on 10 held out datasets not seen during training. In other words, it is a combination of \bton and DG settings, since it requires generalization to both unseen class and data distributions.

\noindent\textbf{Main results.} In Table.~\ref{tab:CDT} we report the results of the Cross-Dataset Generalization setting. This setting is considerably challenging, as the model should generalize to different datasets.  We find that the results are consistent with the previous findings, where our proposed \method generalizes better to unseen classes and data distributions, outperforming all the text-based and multimodal-based prompting techniques by $+2.26\%$ on average. However, with \method we notice a drop in performance with respect to the ZS-CLIP baseline for some fine-grained datasets, such as OxfordPets and StanfordCars, which is potentially due to overlapping attributes for visually similar categories.
\vspace{-0.1cm}
\subsection{Ablation Study}
\label{sec:ablation}
\vspace{-0.15cm}  
\noindent\textbf{Impact of proposed components in \method.} In Tab.~\ref{tab:ablation} we ablate each component of \method on the base-to-new and MSST DG settings. We observe that each component contributes positively to the final performance. First, PromptFormer $f_\Theta$ plays a major role in combining textual information with visual features, thus improving semantic alignment between the two modalities. Second, adding $\mathcal{L}_\text{con}$ (as in Eq.~(\ref{loss_con})) improves the performance by a small margin. Finally, the impact of LoRA can be observed in the penultimate row, where we fine-tune all the parameters of \method. It shows the dangers of overfitting in the FL setup, as some clients may have limited data.

\begin{table}[!ht]
    \centering
    \caption{\textbf{Ablating \method components.} We report numbers for the Base-to-New and MMST DG (on DomainBed) settings.}
    \vspace{-0.2cm}
    \scalebox{0.7}{
    \begin{tabular}{l|ccc}
    \toprule
        
        {Components} & {Base-to-New} & {MSST DG} \\ 
        \midrule
        \centering
         ZS-CLIP & 74.24 & 70.41\\
         \midrule
         $f_\Theta$ \textit{only}	&75.94 & 71.85\\
         $f_\Theta$ + $\mathcal{L}_\text{con}$ &76.27 & 72.14\\
         w/o LoRA &77.41 & 72.58\\
\midrule
         \method (Full)	&\textbf{78.27} & \textbf{73.02}\\
        
\bottomrule
        
    \end{tabular}}
    \label{tab:components}
    
    \label{tab:ablation}
    \vspace{-0.4cm}  
\end{table}

\begin{figure}[h]
    \centering
    \begin{subfigure}[t]{0.48\linewidth}
        \centering
        \includegraphics[width=\linewidth]{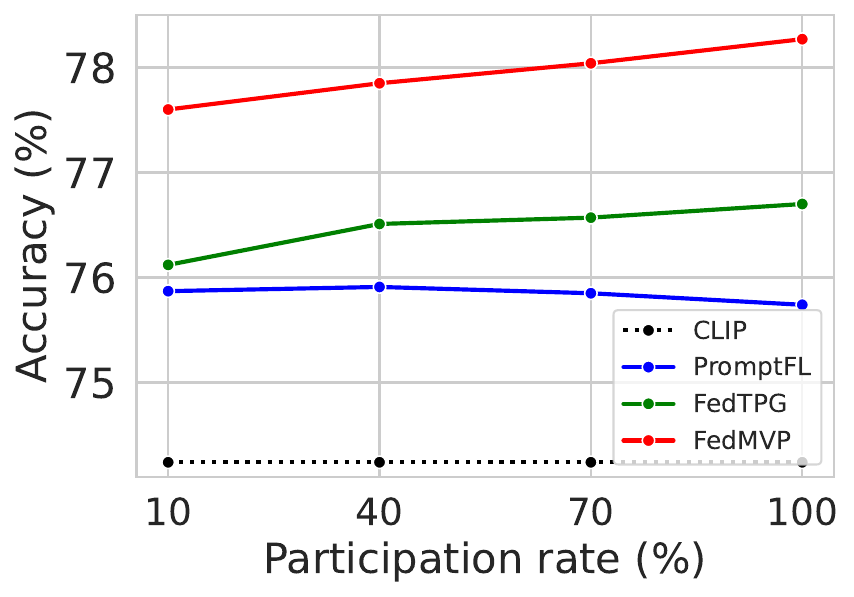}
        \caption{Base-to-New}
        \label{fig:participation_b2n}
    \end{subfigure}
    \hfill
    \begin{subfigure}[t]{0.48\linewidth}
        \centering
        \includegraphics[width=\linewidth]{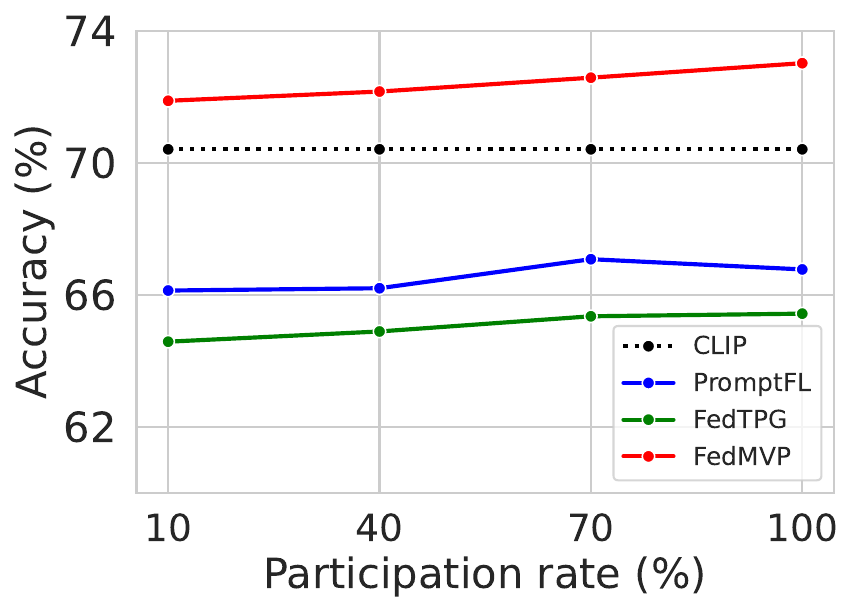}
        \caption{MSST DG}
        \label{fig:participation_msst}
    \end{subfigure}
    \vspace{-0.2cm}
    \caption{\textbf{Sensitivity to participation rate} of the clients, on (a) Base-to-New and (b) MSST DG (DomainBed) settings.}
    \label{fig:participation}
    \vspace{-0.3cm}  
\end{figure}

\noindent\textbf{Participation Rate of the Clients.} In Fig.~\ref{fig:participation} we vary the participation rate from 10\% to 100\% and show how the participation rate of the clients impacts performance on both seen and unseen classes and domains. We observe that \method always exhibits a gentle, yet positive slope in both settings, indicating that continuous parameter aggregation improves generalization. In contrast, for some competitor methods, such as \promptfl, increased participation leads to saturated performance or negative slope in the plot.

\begin{figure}[h]
    \centering
    \begin{subfigure}[t]{0.48\linewidth}
        \centering
        \includegraphics[width=\linewidth]{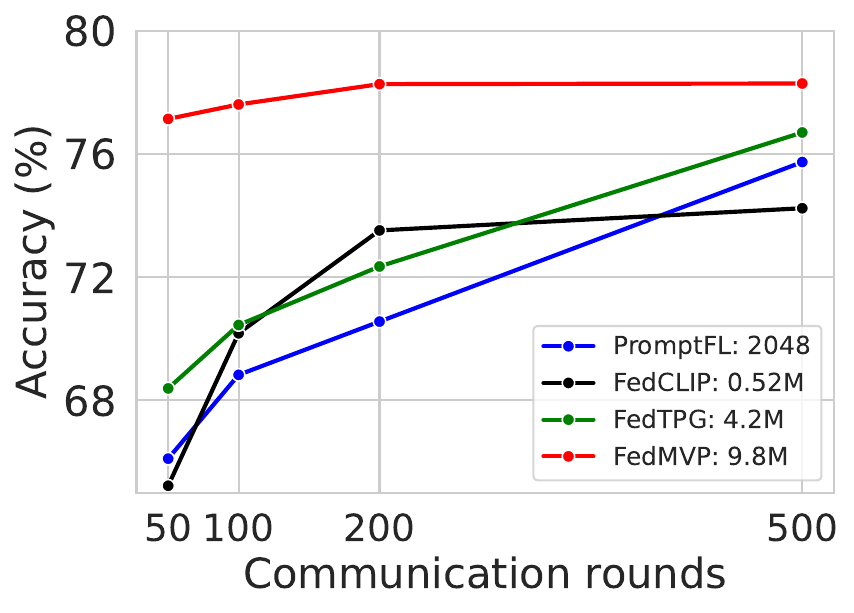}
        \caption{Base-to-New}
        \label{fig:com_b2n}
    \end{subfigure}
    \hfill
    \begin{subfigure}[t]{0.48\linewidth}
        \centering
        \includegraphics[width=\linewidth]{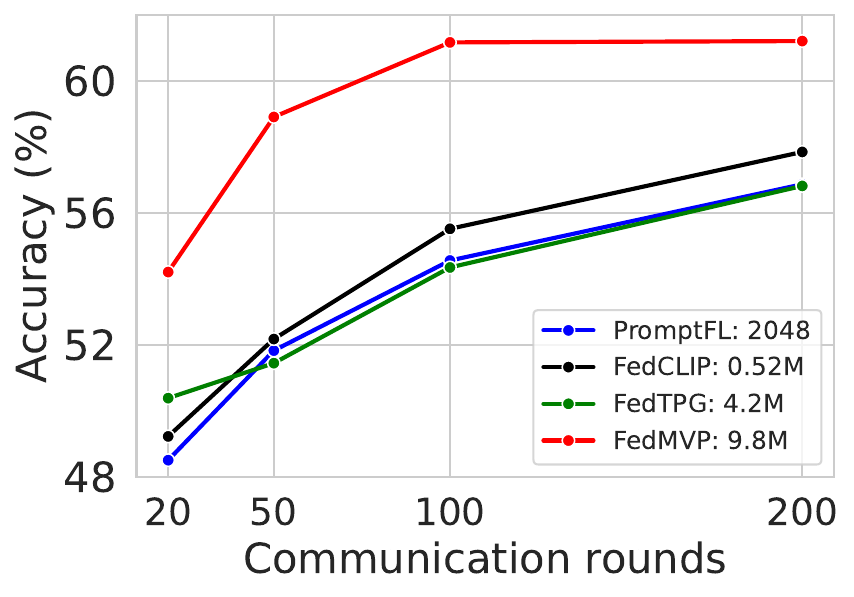}
        \caption{MSST DG}
        \label{fig:com_msst}
    \end{subfigure}
    \vspace{-0.2cm}
    \caption{\textbf{Computational analysis of the methods} on (a) Base-to-New and (b) MSST DG (DomainNet) setting. }
    \label{fig:comround}
    \vspace{-0.3cm}  
\end{figure}

\noindent \textbf{Computational analysis.} In Fig.~\ref{fig:comround} we plot the accuracy vs. communication rounds for FedMVP and competitors in base-to-new and MSST settings. While FedMVP transmits 2$\times$ more parameters than FedTPG, it converges nearly 10$\times$ faster, requiring fewer rounds. In particular, as training progresses, only the LoRA parameters in \pf are fine-tuned in a client, thus reducing the trainable parameter count by 267$\times$. Overall, FedMVP has lower total computational cost with respect to other methods when normalized by the number of communication rounds.

\begin{figure}[h]
    \centering
    \begin{subfigure}[t]{0.48\linewidth}
        \centering
        \includegraphics[width=\linewidth]{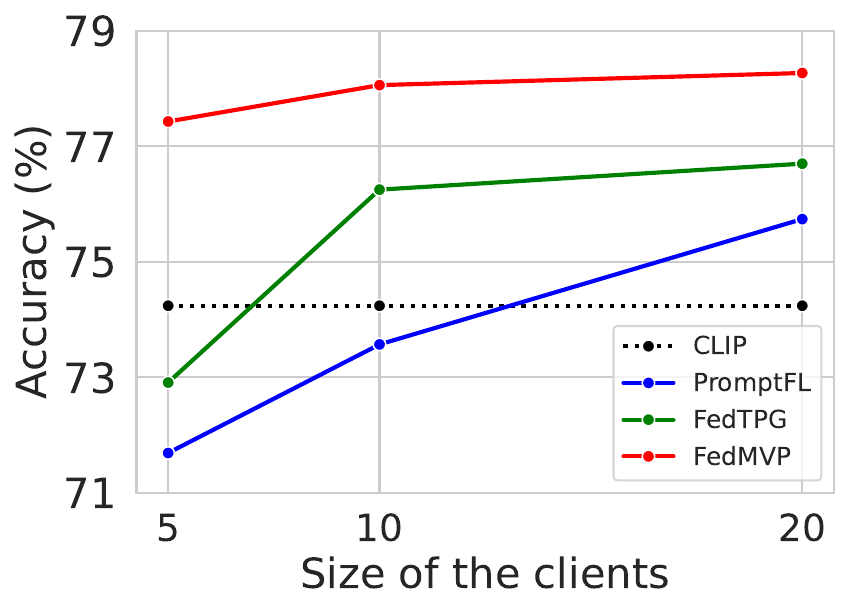}
        \caption{Base-to-New}
        \label{fig:clientsize_b2n}
    \end{subfigure}
    \hfill
    \begin{subfigure}[t]{0.48\linewidth}
        \centering
        \includegraphics[width=\linewidth]{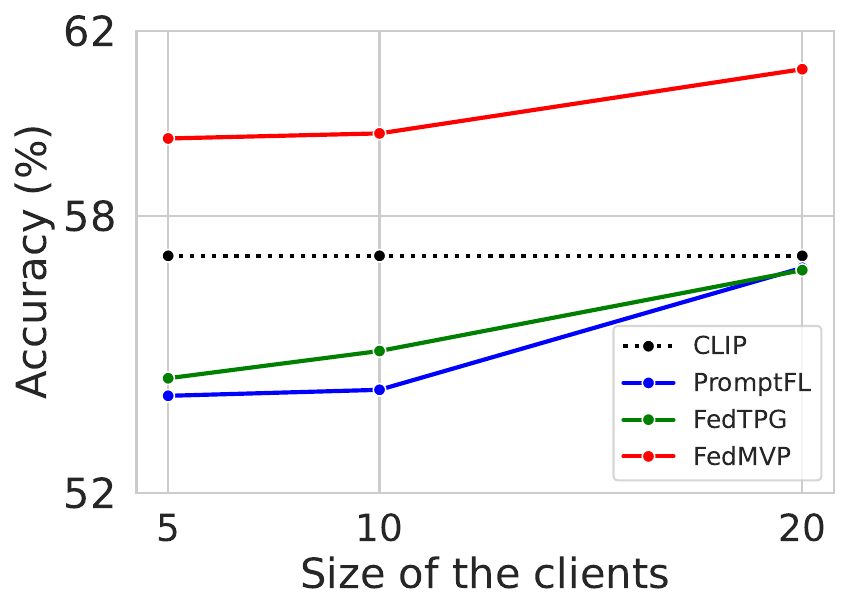}
        \caption{MSST DG}
        \label{fig:clients_msst}
    \end{subfigure}
    \vspace{-0.2cm}
    \caption{\textbf{Sensitivity to number of classes per client} on (a) Base-to-New and (b) MSST DG (DomainNet) setting.}
    \label{fig:clientsize}
    \vspace{-0.6cm}
\end{figure}

\noindent \textbf{Size of the clients.} In Fig. \ref{fig:clientsize}, we vary the number of training classes in each client $K =\{5, 10, 20\}$ while keeping the size of the dataset fixed. In other words, decreasing the number of classes in each client results in an increase in the total number of clients participating in the federated setup. We observe that for all the baselines a higher heterogeneity (or fewer classes per client) leads to lower performance and vice versa.  Nevertheless, \method still exhibits a much superior performance when compared to the baselines, outperforming them while having 4x more clients.

Additional analyses relating to number of shots, choice of LLMs, hyperparameters, and performance in non-federated setting are reported in Sec. B.5 of the \sm

\vspace{-0.2cm}
\section{Conclusion}
\label{conclusion}
\vspace{-0.2cm}  
In this work, we proposed FedMVP, a novel federated prompt tuning framework for CLIP, to address the challenge of reduced generalization to unseen domains and classes. FedMVP combines visual embeddings with textual attribute features to generate multimodal visual prompts, which are tuned using visual prompt tuning. Our findings revealed that the multimodal contextual information in the prompts enhances the generalizability to unseen classes and domains. FedMVP outperforms all its competitors by a notable margin of $+1.57\%-2.26\%$ across benchmarks.

\noindent \textbf{Acknowledgements.} This work was supported by: JPNP23019, subsidized by the New Energy and Industrial Technology Development Organization (NEDO); SERICS (PE00000014) under the NRRP MUR program funded by the EU - NGEU; EU project ELIAS (no: 101120237), ANT (no:  101169439), ELLIOT (no: 101214398) and Ministero delle Imprese e del Made in Italy (IPCEI Cloud DM 27 giugno 2022 - IPCEI-CL-0000007).
{
    \small
    \bibliographystyle{ieeenat_fullname}
    \bibliography{main}

\begin{thebibliography}{63}
\providecommand{\natexlab}[1]{#1}
\providecommand{\url}[1]{\texttt{#1}}
\expandafter\ifx\csname urlstyle\endcsname\relax
  \providecommand{\doi}[1]{doi: #1}\else
  \providecommand{\doi}{doi: \begingroup \urlstyle{rm}\Url}\fi

\bibitem[Abdin et~al.(2024)Abdin, Aneja, Behl, Bubeck, Eldan, Gunasekar, Harrison, Hewett, Javaheripi, Kauffmann, et~al.]{phi4}
Marah Abdin, Jyoti Aneja, Harkirat Behl, S{\'e}bastien Bubeck, Ronen Eldan, Suriya Gunasekar, Michael Harrison, Russell~J Hewett, Mojan Javaheripi, Piero Kauffmann, et~al.
\newblock Phi-4 technical report.
\newblock \emph{arXiv preprint arXiv:2412.08905}, 2024.

\bibitem[Achiam et~al.(2023)Achiam, Adler, Agarwal, Ahmad, Akkaya, Aleman, Almeida, Altenschmidt, Altman, Anadkat, et~al.]{gpt4}
Josh Achiam, Steven Adler, Sandhini Agarwal, Lama Ahmad, Ilge Akkaya, Florencia~Leoni Aleman, Diogo Almeida, Janko Altenschmidt, Sam Altman, Shyamal Anadkat, et~al.
\newblock Gpt-4 technical report.
\newblock \emph{arXiv preprint arXiv:2303.08774}, 2023.

\bibitem[Bai et~al.(2024)Bai, Zhang, Guo, Li, Guo, Hou, Han, and Lu]{diprompt}
Sikai Bai, Jie Zhang, Song Guo, Shuaicheng Li, Jingcai Guo, Jun Hou, Tao Han, and Xiaocheng Lu.
\newblock Diprompt: Disentangled prompt tuning for multiple latent domain generalization in federated learning.
\newblock In \emph{Proceedings of the IEEE/CVF Conference on Computer Vision and Pattern Recognition}, pages 27284--27293, 2024.

\bibitem[Beery et~al.(2018)Beery, Van~Horn, and Perona]{terraincognita}
Sara Beery, Grant Van~Horn, and Pietro Perona.
\newblock Recognition in terra incognita.
\newblock In \emph{Proceedings of the European conference on computer vision (ECCV)}, pages 456--473, 2018.

\bibitem[Bose et~al.(2024)Bose, Jha, Fini, Singha, Ricci, and Banerjee]{stylip}
Shirsha Bose, Ankit Jha, Enrico Fini, Mainak Singha, Elisa Ricci, and Biplab Banerjee.
\newblock Stylip: Multi-scale style-conditioned prompt learning for clip-based domain generalization.
\newblock In \emph{Proceedings of the IEEE/CVF Winter Conference on Applications of Computer Vision}, pages 5542--5552, 2024.

\bibitem[Bossard et~al.(2014)Bossard, Guillaumin, and Gool]{food101}
Lukas Bossard, Matthieu Guillaumin, and Luc~Van Gool.
\newblock Food-101 - mining discriminative components with random forests.
\newblock In \emph{European Conference on Computer Vision}, 2014.

\bibitem[Cimpoi et~al.(2014)Cimpoi, Maji, Kokkinos, Mohamed, and Vedaldi]{dtd}
Mircea Cimpoi, Subhransu Maji, Iasonas Kokkinos, Sammy Mohamed, and Andrea Vedaldi.
\newblock Describing textures in the wild.
\newblock In \emph{Proceedings of the IEEE Conference on Computer Vision and Pattern Recognition}, pages 3606--3613, 2014.

\bibitem[Deng et~al.(2009)Deng, Dong, Socher, Li, Li, and Fei-Fei]{imagenet}
Jia Deng, Wei Dong, Richard Socher, Li-Jia Li, Kai Li, and Li Fei-Fei.
\newblock Imagenet: A large-scale hierarchical image database.
\newblock In \emph{2009 IEEE Conference on Computer Vision and Pattern Recognition}, pages 248--255. Ieee, 2009.

\bibitem[Fan et~al.(2024)Fan, Krishnan, Isola, Katabi, and Tian]{laclip}
Lijie Fan, Dilip Krishnan, Phillip Isola, Dina Katabi, and Yonglong Tian.
\newblock Improving clip training with language rewrites.
\newblock \emph{Advances in Neural Information Processing Systems}, 36, 2024.

\bibitem[Fang et~al.(2013)Fang, Xu, and Rockmore]{vlcs}
Chen Fang, Ye Xu, and Daniel~N. Rockmore.
\newblock Unbiased metric learning: On the utilization of multiple datasets and web images for softening bias.
\newblock In \emph{Proceedings of the IEEE International Conference on Computer Vision (ICCV)}, 2013.

\bibitem[Fei-Fei et~al.(2004)Fei-Fei, Fergus, and Perona]{caltech}
Li Fei-Fei, Rob Fergus, and Pietro Perona.
\newblock Learning generative visual models from few training examples: An incremental bayesian approach tested on 101 object categories.
\newblock In \emph{Conference on Computer Vision and Pattern Recognition Workshop}, pages 178--178. IEEE, 2004.

\bibitem[Gao et~al.(2024)Gao, Geng, Zhang, Ma, Fang, Zhang, Li, and Qiao]{clip-adapter}
Peng Gao, Shijie Geng, Renrui Zhang, Teli Ma, Rongyao Fang, Yongfeng Zhang, Hongsheng Li, and Yu Qiao.
\newblock Clip-adapter: Better vision-language models with feature adapters.
\newblock \emph{International Journal of Computer Vision}, 132\penalty0 (2):\penalty0 581--595, 2024.

\bibitem[Grattafiori et~al.(2024)Grattafiori, Dubey, Jauhri, Pandey, Kadian, Al-Dahle, Letman, Mathur, Schelten, Vaughan, et~al.]{llama3}
Aaron Grattafiori, Abhimanyu Dubey, Abhinav Jauhri, Abhinav Pandey, Abhishek Kadian, Ahmad Al-Dahle, Aiesha Letman, Akhil Mathur, Alan Schelten, Alex Vaughan, et~al.
\newblock The llama 3 herd of models.
\newblock \emph{arXiv preprint arXiv:2407.21783}, 2024.

\bibitem[Gulrajani and Lopez-Paz(2021)]{gulrajani2020search}
Ishaan Gulrajani and David Lopez-Paz.
\newblock In search of lost domain generalization.
\newblock In \emph{International Conference on Learning Representations}, 2021.

\bibitem[Guo et~al.(2023)Guo, Guo, Wang, Tang, and Xu]{promptfl}
Tao Guo, Song Guo, Junxiao Wang, Xueyang Tang, and Wenchao Xu.
\newblock Promptfl: Let federated participants cooperatively learn prompts instead of models-federated learning in age of foundation model.
\newblock \emph{IEEE Transactions on Mobile Computing}, 2023.

\bibitem[Helber et~al.(2019)Helber, Bischke, Dengel, and Borth]{eurosat}
Patrick Helber, Benjamin Bischke, Andreas Dengel, and Damian Borth.
\newblock Eurosat: A novel dataset and deep learning benchmark for land use and land cover classification.
\newblock \emph{IEEE Journal of Selected Topics in Applied Earth Observations and Remote Sensing}, 12\penalty0 (7):\penalty0 2217--2226, 2019.

\bibitem[Hendrycks et~al.(2021{\natexlab{a}})Hendrycks, Basart, Mu, Kadavath, Wang, Dorundo, Desai, Zhu, Parajuli, Guo, et~al.]{img_R}
Dan Hendrycks, Steven Basart, Norman Mu, Saurav Kadavath, Frank Wang, Evan Dorundo, Rahul Desai, Tyler Zhu, Samyak Parajuli, Mike Guo, et~al.
\newblock The many faces of robustness: A critical analysis of out-of-distribution generalization.
\newblock In \emph{Proceedings of the IEEE/CVF International Conference on Computer Vision}, pages 8340--8349, 2021{\natexlab{a}}.

\bibitem[Hendrycks et~al.(2021{\natexlab{b}})Hendrycks, Zhao, Basart, Steinhardt, and Song]{img_A}
Dan Hendrycks, Kevin Zhao, Steven Basart, Jacob Steinhardt, and Dawn Song.
\newblock Natural adversarial examples.
\newblock In \emph{Proceedings of the IEEE/CVF Conference on Computer Vision and Pattern Recognition (CVPR)}, pages 15262--15271, 2021{\natexlab{b}}.

\bibitem[Hu et~al.(2022)Hu, Shen, Wallis, Allen-Zhu, Li, Wang, Wang, Chen, et~al.]{lora}
Edward~J Hu, Yelong Shen, Phillip Wallis, Zeyuan Allen-Zhu, Yuanzhi Li, Shean Wang, Lu Wang, Weizhu Chen, et~al.
\newblock Lora: Low-rank adaptation of large language models.
\newblock In \emph{International Conference on Learning Representations}, page~3, 2022.

\bibitem[Jia et~al.(2021)Jia, Yang, Xia, Chen, Parekh, Pham, Le, Sung, Li, and Duerig]{align}
Chao Jia, Yinfei Yang, Ye Xia, Yi-Ting Chen, Zarana Parekh, Hieu Pham, Quoc Le, Yun-Hsuan Sung, Zhen Li, and Tom Duerig.
\newblock Scaling up visual and vision-language representation learning with noisy text supervision.
\newblock In \emph{International Conference on Machine Learning}, pages 4904--4916. PMLR, 2021.

\bibitem[Jia et~al.(2022)Jia, Tang, Chen, Cardie, Belongie, Hariharan, and Lim]{vpt}
Menglin Jia, Luming Tang, Bor-Chun Chen, Claire Cardie, Serge Belongie, Bharath Hariharan, and Ser-Nam Lim.
\newblock Visual prompt tuning.
\newblock In \emph{European Conference on Computer Vision}, pages 709--727. Springer, 2022.

\bibitem[Kairouz et~al.(2021)Kairouz, McMahan, Avent, Bellet, Bennis, Bhagoji, Bonawitz, Charles, Cormode, Cummings, et~al.]{kairouz2021advances}
Peter Kairouz, H~Brendan McMahan, Brendan Avent, Aur{\'e}lien Bellet, Mehdi Bennis, Arjun~Nitin Bhagoji, Kallista Bonawitz, Zachary Charles, Graham Cormode, Rachel Cummings, et~al.
\newblock Advances and open problems in federated learning.
\newblock \emph{Foundations and trends{\textregistered} in machine learning}, 14\penalty0 (1--2):\penalty0 1--210, 2021.

\bibitem[Khattak et~al.(2023{\natexlab{a}})Khattak, Rasheed, Maaz, Khan, and Khan]{maple}
Muhammad~Uzair Khattak, Hanoona Rasheed, Muhammad Maaz, Salman Khan, and Fahad~Shahbaz Khan.
\newblock Maple: Multi-modal prompt learning.
\newblock In \emph{Proceedings of the IEEE/CVF Conference on Computer Vision and Pattern Recognition (CVPR)}, pages 19113--19122, 2023{\natexlab{a}}.

\bibitem[Khattak et~al.(2023{\natexlab{b}})Khattak, Wasim, Naseer, Khan, Yang, and Khan]{promptsrc}
Muhammad~Uzair Khattak, Syed~Talal Wasim, Muzammal Naseer, Salman Khan, Ming-Hsuan Yang, and Fahad~Shahbaz Khan.
\newblock Self-regulating prompts: Foundational model adaptation without forgetting.
\newblock In \emph{Proceedings of the IEEE/CVF International Conference on Computer Vision}, pages 15190--15200, 2023{\natexlab{b}}.

\bibitem[Krause et~al.(2013)Krause, Stark, Deng, and Fei-Fei]{stanfordcars}
Jonathan Krause, Michael Stark, Jia Deng, and Li Fei-Fei.
\newblock 3d object representations for fine-grained categorization.
\newblock In \emph{IEEE International Conference on Computer Vision Workshops}, pages 554--561, 2013.

\bibitem[Li et~al.(2017)Li, Yang, Song, and Hospedales]{li2017deeper}
Da Li, Yongxin Yang, Yi-Zhe Song, and Timothy~M Hospedales.
\newblock Deeper, broader and artier domain generalization.
\newblock In \emph{Proceedings of the IEEE international conference on computer vision}, pages 5542--5550, 2017.

\bibitem[Li et~al.(2024)Li, Huang, Wang, and Shi]{fedotp}
Hongxia Li, Wei Huang, Jingya Wang, and Ye Shi.
\newblock Global and local prompts cooperation via optimal transport for federated learning.
\newblock In \emph{Proceedings of the IEEE/CVF Conference on Computer Vision and Pattern Recognition}, pages 12151--12161, 2024.

\bibitem[Li et~al.(2025)Li, Cheng, Han, and Song]{dekgtcp}
Yilun Li, Miaomiao Cheng, Xu Han, and Wei Song.
\newblock Divergence-enhanced knowledge-guided context optimization for visual-language prompt tuning.
\newblock In \emph{The Thirteenth International Conference on Learning Representations}, 2025.

\bibitem[Lu et~al.(2023)Lu, Xixu, Wang, and Xie]{fedclip}
Wang Lu, HU Xixu, Jindong Wang, and Xing Xie.
\newblock Fedclip: Fast generalization and personalization for clip in federated learning.
\newblock In \emph{ICLR Workshop on Trustworthy and Reliable Large-Scale Machine Learning Models}, 2023.

\bibitem[Maji et~al.(2013)Maji, Rahtu, Kannala, Blaschko, and Vedaldi]{fgvcaircraft}
Subhransu Maji, Esa Rahtu, Juho Kannala, Matthew~B. Blaschko, and Andrea Vedaldi.
\newblock Fine-grained visual classification of aircraft.
\newblock \emph{CoRR}, abs/1306.5151, 2013.

\bibitem[McMahan et~al.(2017)McMahan, Moore, Ramage, Hampson, and y~Arcas]{fedavg}
Brendan McMahan, Eider Moore, Daniel Ramage, Seth Hampson, and Blaise~Aguera y Arcas.
\newblock Communication-efficient learning of deep networks from decentralized data.
\newblock In \emph{Artificial intelligence and statistics}, pages 1273--1282. PMLR, 2017.

\bibitem[Menon and Vondrick(2023)]{desc}
Sachit Menon and Carl Vondrick.
\newblock Visual classification via description from large language models.
\newblock In \emph{International Conference on Learning Representations}, 2023.

\bibitem[Momeni et~al.(2023)Momeni, Caron, Nagrani, Zisserman, and Schmid]{momeni2023verbs}
Liliane Momeni, Mathilde Caron, Arsha Nagrani, Andrew Zisserman, and Cordelia Schmid.
\newblock Verbs in action: Improving verb understanding in video-language models.
\newblock In \emph{Proceedings of the IEEE/CVF International Conference on Computer Vision}, pages 15579--15591, 2023.

\bibitem[Muhammad et~al.(2020)Muhammad, Wang, O'Reilly-Morgan, Tragos, Smyth, Hurley, Geraci, and Lawlor]{fedfast}
Khalil Muhammad, Qinqin Wang, Diarmuid O'Reilly-Morgan, Elias Tragos, Barry Smyth, Neil Hurley, James Geraci, and Aonghus Lawlor.
\newblock Fedfast: Going beyond average for faster training of federated recommender systems.
\newblock In \emph{Proceedings of the 26th ACM SIGKDD international conference on knowledge discovery \& data mining}, pages 1234--1242, 2020.

\bibitem[Nilsback and Zisserman(2008)]{oxfordflowers}
Maria-Elena Nilsback and Andrew Zisserman.
\newblock Automated flower classification over a large number of classes.
\newblock In \emph{2008 Sixth Indian Conference on Computer Vision, Graphics \& Image Processing}, pages 722--729, 2008.

\bibitem[Parkhi et~al.(2012)Parkhi, Vedaldi, Zisserman, and Jawahar]{oxfordpets}
Omkar~M. Parkhi, Andrea Vedaldi, Andrew Zisserman, and C.~V. Jawahar.
\newblock Cats and dogs.
\newblock In \emph{IEEE Conference on Computer Vision and Pattern Recognition}, 2012.

\bibitem[Peng et~al.(2019)Peng, Bai, Xia, Huang, Saenko, and Wang]{domainnet}
Xingchao Peng, Qinxun Bai, Xide Xia, Zijun Huang, Kate Saenko, and Bo Wang.
\newblock Moment matching for multi-source domain adaptation.
\newblock In \emph{Proceedings of the IEEE/CVF international conference on computer vision}, pages 1406--1415, 2019.

\bibitem[Pratt et~al.(2023)Pratt, Covert, Liu, and Farhadi]{pratt2023does}
Sarah Pratt, Ian Covert, Rosanne Liu, and Ali Farhadi.
\newblock What does a platypus look like? generating customized prompts for zero-shot image classification.
\newblock In \emph{Proceedings of the IEEE/CVF International Conference on Computer Vision}, pages 15691--15701, 2023.

\bibitem[Qiu et~al.(2024)Qiu, Li, Mummadi, Ganesh, Li, Peng, and Lin]{fedtpg}
Chen Qiu, Xingyu Li, Chaithanya~Kumar Mummadi, Madan~Ravi Ganesh, Zhenzhen Li, Lu Peng, and Wan-Yi Lin.
\newblock Federated text-driven prompt generation for vision-language models.
\newblock In \emph{The Twelfth International Conference on Learning Representations}, 2024.

\bibitem[Radford et~al.(2021)Radford, Kim, Hallacy, Ramesh, Goh, Agarwal, Sastry, Askell, Mishkin, Clark, et~al.]{radford2019language}
Alec Radford, Jong~Wook Kim, Chris Hallacy, Aditya Ramesh, Gabriel Goh, Sandhini Agarwal, Girish Sastry, Amanda Askell, Pamela Mishkin, Jack Clark, et~al.
\newblock Learning transferable visual models from natural language supervision.
\newblock In \emph{International Conference on Machine Learning}, pages 8748--8763. PMLR, 2021.

\bibitem[Recht et~al.(2019)Recht, Roelofs, Schmidt, and Shankar]{imgv2}
Benjamin Recht, Rebecca Roelofs, Ludwig Schmidt, and Vaishaal Shankar.
\newblock Do imagenet classifiers generalize to imagenet?
\newblock In \emph{International Conference on Machine Learning}, pages 5389--5400. PMLR, 2019.

\bibitem[Roy and Etemad(2024)]{coprompt}
Shuvendu Roy and Ali Etemad.
\newblock Consistency-guided prompt learning for vision-language models.
\newblock In \emph{International Conference on Learning Representations}, 2024.

\bibitem[So et~al.(2022)So, He, Yang, Li, Yu, E~Ali, Guler, and Avestimehr]{lightsecagg}
Jinhyun So, Chaoyang He, Chien-Sheng Yang, Songze Li, Qian Yu, Ramy E~Ali, Basak Guler, and Salman Avestimehr.
\newblock Lightsecagg: a lightweight and versatile design for secure aggregation in federated learning.
\newblock \emph{Proceedings of Machine Learning and Systems}, 4:\penalty0 694--720, 2022.

\bibitem[Soomro et~al.(2012)Soomro, Zamir, and Shah]{ucf101}
Khurram Soomro, Amir~Roshan Zamir, and Mubarak Shah.
\newblock {UCF101:} {A} dataset of 101 human actions classes from videos in the wild.
\newblock \emph{CoRR}, abs/1212.0402, 2012.

\bibitem[Sun et~al.(2024{\natexlab{a}})Sun, Mendieta, Dutta, Li, and Chen]{fedcola}
Guangyu Sun, Matias Mendieta, Aritra Dutta, Xin Li, and Chen Chen.
\newblock Towards multi-modal transformers in federated learning.
\newblock In \emph{European Conference on Computer Vision}, pages 229--246. Springer, 2024{\natexlab{a}}.

\bibitem[Sun et~al.(2024{\natexlab{b}})Sun, Niu, and Wei]{fedunder}
Zhenyu Sun, Xiaochun Niu, and Ermin Wei.
\newblock Understanding generalization of federated learning via stability: Heterogeneity matters.
\newblock In \emph{International Conference on Artificial Intelligence and Statistics}, pages 676--684. PMLR, 2024{\natexlab{b}}.

\bibitem[Vaswani et~al.(2017)Vaswani, Shazeer, Parmar, Uszkoreit, Jones, Gomez, Kaiser, and Polosukhin]{transformer}
Ashish Vaswani, Noam Shazeer, Niki Parmar, Jakob Uszkoreit, Llion Jones, Aidan~N Gomez, {\L}ukasz Kaiser, and Illia Polosukhin.
\newblock Attention is all you need.
\newblock In \emph{Advances in Neural Information Processing Systems}, 2017.

\bibitem[Venkateswara et~al.(2017)Venkateswara, Eusebio, Chakraborty, and Panchanathan]{officehome}
Hemanth Venkateswara, Jose Eusebio, Shayok Chakraborty, and Sethuraman Panchanathan.
\newblock Deep hashing network for unsupervised domain adaptation.
\newblock In \emph{Proceedings of the IEEE Conference on Computer Vision and Pattern Recognition}, pages 5018--5027, 2017.

\bibitem[Wang et~al.(2019)Wang, Ge, Lipton, and Xing]{img_sketch}
Haohan Wang, Songwei Ge, Zachary Lipton, and Eric~P Xing.
\newblock Learning robust global representations by penalizing local predictive power.
\newblock In \emph{Advances in Neural Information Processing Systems}, 2019.

\bibitem[Wortsman et~al.(2022)Wortsman, Ilharco, Kim, Li, Kornblith, Roelofs, Lopes, Hajishirzi, Farhadi, Namkoong, et~al.]{wortsman2022robust}
Mitchell Wortsman, Gabriel Ilharco, Jong~Wook Kim, Mike Li, Simon Kornblith, Rebecca Roelofs, Raphael~Gontijo Lopes, Hannaneh Hajishirzi, Ali Farhadi, Hongseok Namkoong, et~al.
\newblock Robust fine-tuning of zero-shot models.
\newblock In \emph{Proceedings of the IEEE/CVF conference on computer vision and pattern recognition}, pages 7959--7971, 2022.

\bibitem[Xiao et~al.(2010)Xiao, Hays, Ehinger, Oliva, and Torralba]{sundataset}
Jianxiong Xiao, James Hays, Krista~A Ehinger, Aude Oliva, and Antonio Torralba.
\newblock Sun database: Large-scale scene recognition from abbey to zoo.
\newblock In \emph{2010 IEEE computer society conference on computer vision and pattern recognition}, pages 3485--3492. IEEE, 2010.

\bibitem[Xing et~al.(2023)Xing, Wu, Cheng, Zhang, Liang, Wang, and Zhang]{xing2023dual}
Yinghui Xing, Qirui Wu, De Cheng, Shizhou Zhang, Guoqiang Liang, Peng Wang, and Yanning Zhang.
\newblock Dual modality prompt tuning for vision-language pre-trained model.
\newblock \emph{IEEE Transactions on Multimedia}, 26:\penalty0 2056--2068, 2023.

\bibitem[Yang et~al.(2024)Yang, Yang, Zhang, Hui, Zheng, Yu, Li, Liu, Huang, Wei, et~al.]{qwen2}
An Yang, Baosong Yang, Beichen Zhang, Binyuan Hui, Bo Zheng, Bowen Yu, Chengyuan Li, Dayiheng Liu, Fei Huang, Haoran Wei, et~al.
\newblock Qwen2. 5 technical report.
\newblock \emph{arXiv preprint arXiv:2412.15115}, 2024.

\bibitem[Yang et~al.(2023)Yang, Panagopoulou, Zhou, Jin, Callison-Burch, and Yatskar]{labo}
Yue Yang, Artemis Panagopoulou, Shenghao Zhou, Daniel Jin, Chris Callison-Burch, and Mark Yatskar.
\newblock Language in a bottle: Language model guided concept bottlenecks for interpretable image classification.
\newblock In \emph{Proceedings of the IEEE/CVF Conference on Computer Vision and Pattern Recognition}, pages 19187--19197, 2023.

\bibitem[Yao et~al.(2023)Yao, Zhang, and Xu]{kgcoop}
Hantao Yao, Rui Zhang, and Changsheng Xu.
\newblock Visual-language prompt tuning with knowledge-guided context optimization.
\newblock In \emph{Proceedings of the IEEE/CVF Conference on Computer Vision and Pattern Recognition}, pages 6757--6767, 2023.

\bibitem[Yao et~al.(2024)Yao, Zhang, and Xu]{tcp}
Hantao Yao, Rui Zhang, and Changsheng Xu.
\newblock Tcp: Textual-based class-aware prompt tuning for visual-language model.
\newblock In \emph{Proceedings of the IEEE/CVF Conference on Computer Vision and Pattern Recognition}, pages 23438--23448, 2024.

\bibitem[Yao et~al.(2018)Yao, Huang, and Sun]{fedcom1}
Xin Yao, Chaofeng Huang, and Lifeng Sun.
\newblock Two-stream federated learning: Reduce the communication costs.
\newblock In \emph{2018 IEEE Visual Communications and Image Processing (VCIP)}, pages 1--4. IEEE, 2018.

\bibitem[Yao et~al.(2019)Yao, Huang, Wu, Zhang, and Sun]{fedcom2}
Xin Yao, Tianchi Huang, Chenglei Wu, Rui-Xiao Zhang, and Lifeng Sun.
\newblock Federated learning with additional mechanisms on clients to reduce communication costs.
\newblock \emph{arXiv preprint arXiv:1908.05891}, 2019.

\bibitem[Yi et~al.(2023)Yi, Wang, Liu, Shi, and Yu]{fedgh}
Liping Yi, Gang Wang, Xiaoguang Liu, Zhuan Shi, and Han Yu.
\newblock Fedgh: Heterogeneous federated learning with generalized global header.
\newblock In \emph{Proceedings of the 31st ACM International Conference on Multimedia}, pages 8686--8696, 2023.

\bibitem[Zhai et~al.(2023)Zhai, Mustafa, Kolesnikov, and Beyer]{zhai2023sigmoid}
Xiaohua Zhai, Basil Mustafa, Alexander Kolesnikov, and Lucas Beyer.
\newblock Sigmoid loss for language image pre-training.
\newblock In \emph{Proceedings of the IEEE/CVF international conference on computer vision}, pages 11975--11986, 2023.

\bibitem[Zhang et~al.(2024)Zhang, Wu, Gao, Shen, and Song]{dept}
Ji Zhang, Shihan Wu, Lianli Gao, Heng~Tao Shen, and Jingkuan Song.
\newblock Dept: Decoupled prompt tuning.
\newblock In \emph{Proceedings of the IEEE/CVF Conference on Computer Vision and Pattern Recognition}, pages 12924--12933, 2024.

\bibitem[Zhou et~al.(2022{\natexlab{a}})Zhou, Yang, Loy, and Liu]{cocoop}
Kaiyang Zhou, Jingkang Yang, Chen~Change Loy, and Ziwei Liu.
\newblock Conditional prompt learning for vision-language models.
\newblock In \emph{Proceedings of the IEEE/CVF Conference on Computer Vision and Pattern Recognition}, pages 16816--16825, 2022{\natexlab{a}}.

\bibitem[Zhou et~al.(2022{\natexlab{b}})Zhou, Yang, Loy, and Liu]{coop}
Kaiyang Zhou, Jingkang Yang, Chen~Change Loy, and Ziwei Liu.
\newblock Learning to prompt for vision-language models.
\newblock \emph{International Journal of Computer Vision}, 130\penalty0 (9):\penalty0 2337--2348, 2022{\natexlab{b}}.

\end{thebibliography}
}

\newpage
\twocolumn[{
\vspace{1em}
\begin{center}
\section*{FedMVP: Federated Multimodal Visual Prompt Tuning for Vision-Language Models (Supplementary Material)}
\end{center}
\vspace{1em}
}]

\appendix
The supplementary material is organized as follows: in Sec.~\ref{sec:addn_details} we lay down additional implementation details of our proposed \method. In Sec.~\ref{sec:addn_exp} we provide a detailed breakdown of the experimental results that have been reported in the main paper.

\section{Additional implementation details}
\label{sec:addn_details}

\subsection{Attribute generation and usage}
\label{attributes}

\paragraph{Attribute generation.} As discussed in Sec. 3.3.1 of the main paper, one of the core proposals in our proposed \method is to integrate the class attribute information into the multimodal prompt generation process orchestrated by the \pf network.

The attributes for a given class are generated using a large language model (LLM), such as GPT-4o~\cite{gpt4} in our case. For the $k\textsuperscript{th}$ class name in the $i\textsuperscript{th}$ client $c_{i,k}$, we query GPT-4o using a structured instructional prompt, following~\cite{desc} as:

\begin{tcolorbox}[width=\linewidth, colback=white!95!black, title=LLM Prompt]
``What are the most useful detailed generic visual features for distinguishing a [class name] in an image? Please act as an expert with comprehensive knowledge of all aspects of generic objects.''
\end{tcolorbox}
\noindent where the ``class name'' is replaced with the value of $c_{i,k}$. For instance, when we prompt GPT-4o with the class name ``giraffe'' we get a comma separated list of attributes:
\begin{tcolorbox}[width=\linewidth, colback=white!95!black, title=Attributes generated by LLM]

    ``Distinctive coat pattern with large, irregular brown patches'',
    ``unique coat pattern with large, irregular brown patches'',
    ``exceptionally long neck, a primary distinguishing feature'',
    ``small, rounded ossicones or horns on the head'',
    ``slender, elongated legs, emphasizing their height'',
    ``tall, narrow body frame with prominent shoulders.''

\end{tcolorbox}

\begin{figure*}[!t]
\centering
    \includegraphics[width=1.0\linewidth]{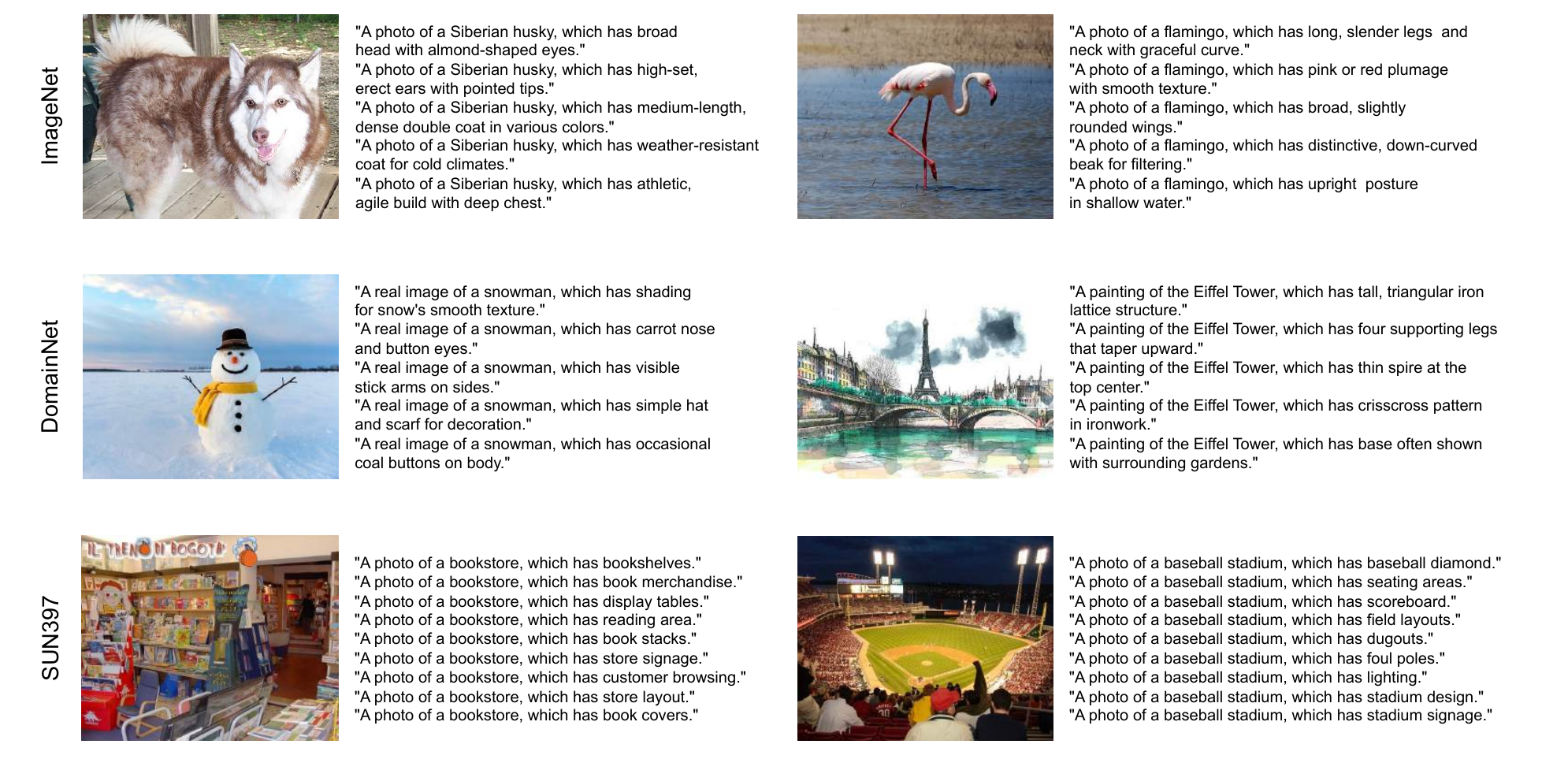}
    
    \caption{\textbf{Examples of LLM-generated attributes and complete text prompts.} We report some example text prompts used in \method for the datasets ImageNet, DomainNet and SUN397.}
    \label{fig:attribute}
     
\end{figure*}

\paragraph{Composing text prompts.}
% ------
In addition to the attributes, we utilize generic hand-crafted prefixes (\eg ``a photo of a [class name]''), as used in \cite{radford2019language}, or domain-specific prefixes (\eg ``a sketch of a [class name]'') tailored to each dataset. Details of the prefix templates for each benchmark are reported in Table \ref{tab:datasets}.
We then combine these prefixes with GPT-4o generated attributes using connector phrases ``which is a/an'' or ``which has'', to form composite text prompts for the CLIP text-encoder feature extractor. A complete example of text prompt for the class ``giraffe'' that is used for CLIP text feature attribute extraction is given as follows:

\begin{tcolorbox}[width=\linewidth, colback=white!95!black, title=Composite prompt for CLIP text encoder]
``A photo of a giraffe, which has a distinctive coat pattern with large, irregular brown patches.''
\end{tcolorbox}
We provide more examples of the LLM-generated attributes and the complete text prompts in Fig. \ref{fig:attribute}. 

\paragraph{Using attributes during training.} Note from Fig. 2 of the main paper that there is a distinction between how attributes are used by the \pf network and for training the \method. To recap, the \pf network takes as input only the LLM-generated attributes for constructing the multimodal visual prompts. Whereas, for training the \method we use the composite text prompts, which has just been described above. Given a class name $c_{k}$\footnote{Omitting client index $i$ for brevity.} we obtain the text feature $\vectrm{t}_k$ (used in Eq. 5 of the main paper) from the CLIP text encoder $\mathcal{E}_t$ by using a composite text prompt. Then we repeat this step for all the attributes generated by the LLM for a given class $k$, which gives us a matrix of CLIP text embeddings $\vectrm{T}_k$, where the entry $\vectrm{T}_{k,j}$ is the text embedding corresponding the $j\textsuperscript{th}$ attribute for class $k$. Given all the text embeddings for a class, we are now ready to compute the CLIP similarity in Eq. 5 of the main paper.

We follow the scoring proposed in DESC~\cite{desc}, where for a given image, belonging to class label $y$ and class name $c_k$, the final prediction probability $p(y=k \mid \vectrm{I})$ are calculated by averaging the score for each attribute $j$ as $p(y=k|\vectrm{I}) =$:

\begin{equation}
    \frac{1}{\text{dim}_2(\vectrm{T}_{k,J})}\sum_{j \in J}\frac{\exp(\text{cos}(\vectrm{v}, \vectrm{T}_{k,j}))/\tau)}{\sum^{K}_{k'=1} \exp(\text{cos}(\vectrm{v}, \vectrm{T}_{k',j}))/\tau)},
\label{eqprob_desc}
\end{equation}
where $\text{dim}_2(\cdot)$ denotes the dimension or total number of attributes $J$ for a given class $k$.

\begin{table}[t]
% \vspace{-0.3cm}
% \scriptsize{
    \centering
    \caption{\textbf{Comparison of our \method with zero-shot CLIP and DESC.} Average harmonic mean is reported for Base-to-New generalization and average accuracy of DomainBed benchmarks of multi-source single-target generalization.}
    \vspace{-.2cm}
    \scalebox{0.69}{
    \begin{tabular}{lc|cc}
    \toprule
        
        \textbf{Methods} & \textbf{Prompts} & {Base-to-New} & {MSST}\\ 
        \midrule
         
        ZS-CLIP \cite{radford2019language} &\multirow{2}{*}{hand-crafted} &74.24	&70.41\\ 

        DESC \cite{desc} & &75.18 &69.89 \\
        \midrule
         
        \method (Ours) & textual+visual &\textbf{77.52}	&\textbf{72.24} \\
\bottomrule
    \end{tabular}}
    \label{tab:clip-desc}
    \vspace{-0.5cm}
\end{table}

\paragraph{Using attributes during inference.} For inference, we follow the same scoring as per Eq.~\ref{eqprob_desc}. The predicted class is given by taking an argmax of the probability distribution over all the classes. Since we utilize the scoring method of DESC in our method, we also compare the performance of DESC with \method on two generalization settings -- base-to-new generalization and multi-source single-target generalization -- in Tab.~\ref{tab:clip-desc}. From DESC numbers we can observe that using LLM-generated attributes alone is not sufficient to improve much over the ZS-CLIP baseline. Interestingly, DESC leads to a drop in performance over the ZS-CLIP baseline that does not use any attributes on the DomainBed benchmark. Contrarily, our \method can better utilize the LLM attributes through the cross-attention mechanism of the proposed \pf network, as described in Sec. 3.3.1 of the main paper.

\begin{algorithm*}[t!]
\caption{\methodlong (\method) algorithm}
\label{pseudocode}
\begin{algorithmic}[1]
\Procedure{Server Execution}{}
       \State \( \rho^{0} = \{\Theta^{0}\} \)
       \Comment{Initialize \( \rho^{0}\) parameters of \pf module}
       \For{\( r \) $\gets$ 0 \textbf{to} \( R \)} 
       \Comment{For total communication rounds \( R \)}
       \State Choose a random subset of remote clients as \(S_r \)
       \For{ \(i \) $\in$ \(S_r \) in parallel} 
       \Comment{For a client \(i \), belongs to \(S_r \)}
       \State Send the current global model \( \rho^{r}\) to client \(i \)
       \State Receive locally updated \( \rho^{r+1}_{i}\) from \textbf{Client Training}
       \EndFor
       \State Aggregate the updated model parameters, \( \rho^{r+1} = \frac{1}{|S_r|} \sum_{i \in S_r} \rho^{r+1}_i \)
       \EndFor
       \State Get the final model parameter \( \rho^{R} = \{\Theta^{R}, \theta^{R}\} \)
\EndProcedure
\Procedure{Client Training}{}
       \State Generate the attributes of set of classes, $\mathcal{C}_i = \{c_{i,k}\}^{K_i}_{k=1}$, by a LLM
       \Comment{$K_i$ is the total number of classes of $i$}
       \State Extract the attribute embeddings, $\vectrm{A}_{i} = \{\mathcal{E}_t (\texttt{LLM}(\mathcal{C}_i))\}$
       \For{\( l \) $\gets$ 0 \textbf{to} \( L \)} 
       \Comment{For local epochs \( L \)}
       \If{loss \textgreater threshold}
       \State Generate the visual prompts $\vectrm{P}$ using eq. 2 and eq. 3
       \Comment{Follow eq. 2 and eq. 3 from the main paper}
       \State Concatenate the [\texttt{CLS}] token, the patch embeddings, and the visual prompts, as $\vectrm{I} = [\vectrm{z}; \vectrm{E}; \vectrm{P}]$
       \State Extract the visual features from $\mathcal{E}_v$
       
       \Else
       \State Start LoRA fine-tuning of the \pf module
       \EndIf
       \State Estimate the prediction scores using Eq. \ref{eqprob_desc}
       \State Calculate and update the losses using eq. 4 to eq. 7
       \Comment{Follow eq. 4 to eq. 7 from the main paper}
       \State Update the parameters \( \rho^{r} \) to \( \rho^{r+1}_{i} \) locally using eq. 8 on \( (x,y) \sim \mathcal{D}_i\)
       \Comment{Follow eq. 8 from the main paper}
       \EndFor

\EndProcedure
\end{algorithmic}
\end{algorithm*}

Since all the evaluations are done on the server, except for local accuracy, as described in Sec. 4.1 of the main paper, each client sends the LLM-generated attributes to the server, which constitutes a relatively tiny communication overhead. Alternatively, the server can also generate the attributes for all the classes, since the clients will have no knowledge about the disjoint classes in other clients.

\subsection{Pseudo-code of \method}
In Algorithm~\ref{pseudocode} we provide a pseudo-code of the full \method algorithm. We split the algorithm into two parts: one for server execution and another for client training.

\subsection{Architecture details of \method}
The only trainable parameters of \method are composed of \pf network parameters. Below we describe additional architecture details of each trainable network. Note that we do not tune the vision and text encoder backbones of CLIP, and hence refer the reader to the original paper~\cite{radford2019language} for the architecture details of CLIP.

\paragraph{PromptFormer.} The PromptFormer network \( f_\Theta \) consists of two multi-head cross-attention (MHCA) modules, two feed-forward networks (FFN), a projection layer $T_\text{proj}$ and a learnable query prompt $\vectrm{Q}$. Each MHCA module consists of a 4-head cross-attention mechanism, followed by LayerNorm. Each FFN comprises of a two-layer bottleneck structure (Linear-GeLU-Linear). $T_\text{proj}$ performs a linear transformation to convert the textual feature space of dimension 512 into the patch embedding space with a dimension of 768. $\vectrm{Q}$ is comprised of prompt length 4, initialized with a Gaussian distribution of $\sigma = 0.02$.

\section{Additional experimental results}
\label{sec:addn_exp}

\subsection{Dataset details.}

\begin{table*}[t]
    \centering
    \begin{minipage}[t]{0.95\textwidth}
        \centering
        \caption{\textbf{Dataset Details}}
        \label{tab:datasets}
        
        % First Table
        \begin{minipage}[t]{\textwidth}
            \centering
            \subcaption{Domain Generation dataset statistical details on class, training and test splits, prefix template.} 
            \scalebox{0.69}{
            \begin{tabular}{lc|ccccc}
    \toprule
        
        \textbf{Dataset} & \textbf{Domain} & {Classes} & {Train} & {Test} & {Prefix template}\\ 
        \midrule

         \multirow{4}{*}{PACS \cite{li2017deeper}} &Art Painting &\multirow{4}{*}{7} &1,024	&614	&An art painting of a [CLASS]\\
         &Cartoon & &1,171	&704	&A cartoon of a [CLASS]\\
         &Photo & &835	&502	&A photo of a [CLASS]\\
         &Sketch & &1,964	&1,179	&A sketch of a [CLASS]\\
         \midrule
         
         \multirow{4}{*}{OfficeHome \cite{officehome}} &Art &\multirow{4}{*}{65} &1,214	&728	&An art of a [CLASS]\\
         &Clipart & &2,191	&1,298	&A clipart of a [CLASS]\\
         &Product & &2,226	&1,324	&A product image of a [CLASS]\\
         &RealWorld & &2,180	&1,304	&A realworld image of a [CLASS]\\
         \midrule
         
         \multirow{4}{*}{VLCS \cite{vlcs}} &CALTECH &\multirow{4}{*}{5} &891	&424	&A high quality photo of a [CLASS], as a standalone object\\
         &LABELME & &1,672	&797	&A realworld photo of the [CLASS]\\
         &PASCAL-VOC & &2,127	&1,013	&A realworld photo of a [CLASS]\\
         &SUN & &2,067	&985	&A photo of a [CLASS], in diverse scenic environments\\
         \midrule
         
         \multirow{4}{*}{Terra Incognita \cite{terraincognita}} &Location-38 &\multirow{4}{*}{10} &4,883	&2,930	&A photo of a [CLASS]\\
         &Location-43 & &2,009	&1,207	&A photo of a [CLASS]\\
         &Location-46 & &3,061	&1,836	&A photo of a [CLASS]\\
         &Location-100 & &2,439	&1,466	&A photo of a [CLASS]\\

         \midrule
         
         \multirow{6}{*}{DomainNet \cite{domainnet}} &Clipart &\multirow{6}{*}{345} &24,417	&14,647	&A clipart of a [CLASS]\\
         &Infograph & &26,609	&15,948	&An infograph of a [CLASS]\\
         &Painting & &37,873	&22,744	&A painting of a [CLASS]\\
         &Quickdraw & &86,250	&51,750	&A quickdraw image of a [CLASS]\\
         &Real & &87,663	&52,604	&A real image of a [CLASS]\\
         &Sketch & &35,195	&21,109	&A sketch of a [CLASS]\\

\bottomrule
            \end{tabular}}
            \label{tab:first_sub_table}
        \end{minipage}
        
        \vspace{0.2cm}  % Space between the two tables

        % Second Table
        \begin{minipage}[t]{\textwidth}
            \centering
            \subcaption{Dataset statistical details on class, training and test splits, prefix template.}
            \scalebox{0.69}{
           \begin{tabular}{lc|cccc}
    \toprule
        
        \textbf{Dataset} & {Classes} & {Train} & {Test} & {Prefix template}\\ 
        \midrule
         
         Caltech101 \cite{caltech} &101 &4,128 &2,465 &A photo of a [CLASS] \\
         Flowers102 \cite{oxfordflowers} &102 &4,093 &2,463 &A photo of a [CLASS], a type of flower \\
         FGVCAircraft \cite{fgvcaircraft} &100 &3,334 &3,333 &A photo of a [CLASS], a type of aircraft \\
         UCF101 \cite{ucf101} &101 &7,639 &3,783 &A photo of a person doing [CLASS]\\
         OxfordPets \cite{oxfordpets} &37 &2,944 &3,369 &A photo of a [CLASS], a type of pet \\
         Food101 \cite{food101} &101 &50,500 &30,300 &A photo of a [CLASS], a type of food\\
         DTD \cite{dtd} & 47 &2,820 &1,692 &A photo of a [CLASS], a type of texture\\
         StanfordCars \cite{stanfordcars} &196 &6,509 &8,041 &A photo of a [CLASS]\\
         SUN397 \cite{sundataset} &397 &15,880 &19,850 &A photo of a [CLASS]\\
         EuroSAT \cite{eurosat} &10 &13500 &8,100 &A centered satellite photo of [CLASS]\\

\midrule
         ImageNet \cite{imagenet} &1000 &1.28M &50,000 &A photo of a [CLASS]\\
         ImageNetV2 \cite{imgv2} &1000 &N/A &10,000 &A photo of a [CLASS]\\
         ImageNet-Sketch \cite{img_sketch} &1000 &N/A &50,889 &A photo of a [CLASS]\\
         ImageNet-A \cite{img_A} &200 &N/A &7500 &A photo of a [CLASS]\\
         ImageNet-R \cite{img_R} &200 &N/A &30,000 &A photo of a [CLASS]\\
               
\bottomrule
            \end{tabular}}
            \label{tab:second_sub_table}
        \end{minipage}
    \end{minipage}
\end{table*}

In Tabs.~\ref{tab:datasets}(a) and (b) we provide detailed information of all the datasets used in the experiments of the main paper and the associated statistics, such as the number of classes, number of samples, and the prefix templates. In detail, the Tab.~\ref{tab:datasets}(a) includes the datasets used in the experiments corresponding to Tab. 2 of the main paper. The Tab.~\ref{tab:datasets}(a) includes the datasets used in Tabs. 1, 2, and 4 of the main paper. We refer the reader to the corresponding papers that have proposed the original datasets for further details and example images.

\begin{figure*}[!t]
\centering
    \includegraphics[width=1.0\linewidth]{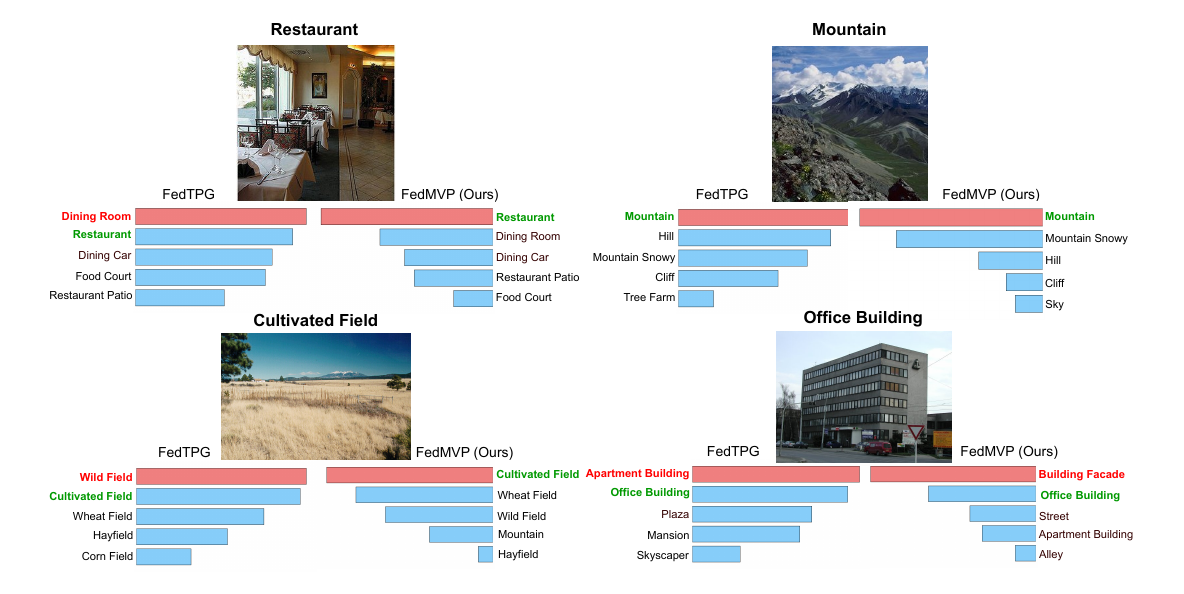}
    \vspace{-0.5cm}  
    \caption{\textbf{Qualitative comparison} of top-5 predictions on SUN397 dataset in Base-to-New Generalization setting. The correct predictions (and annotations) are highlighted with green and the incorrect predictions are highlighted with red.}
    \label{fig:top5_sun}
    
\end{figure*}

\begin{figure*}[!t]
\centering
    \includegraphics[width=1.0\linewidth]{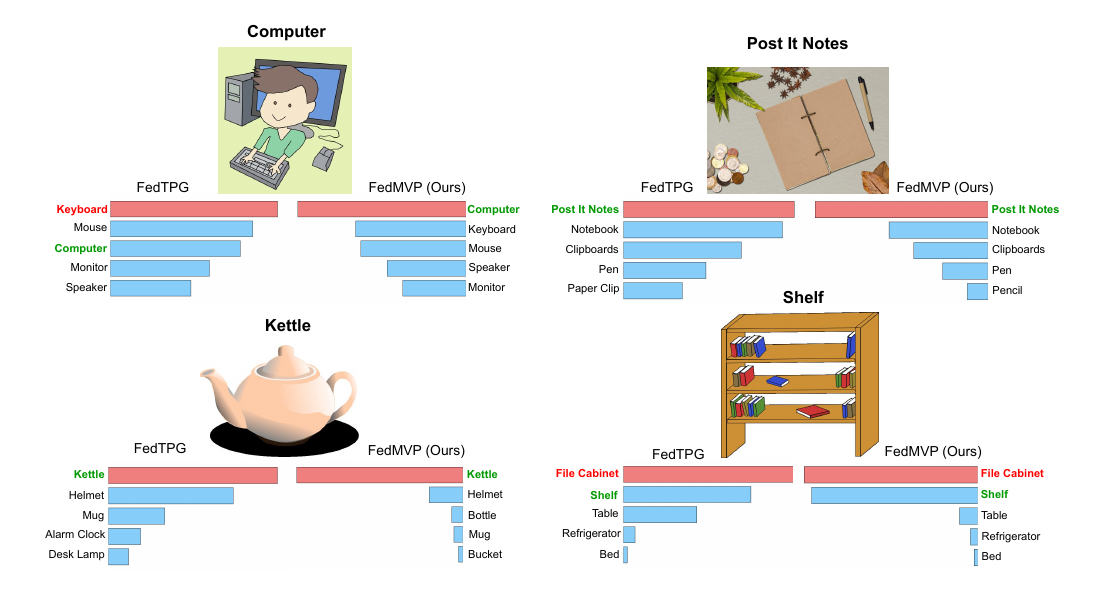}
    \vspace{-0.5cm}  
    \caption{\textbf{Qualitative comparison} of top-5 predictions on Clipart domain of OfficeHome dataset in MSST DG setting. The correct predictions (and annotations) are highlighted with green and the incorrect predictions are highlighted with red.}
    \label{fig:top5_office}
\end{figure*}

\subsection{Experimental setup}

\paragraph{Metrics.} We assess the performance of all methods using classification accuracy. In the base-to-novel generalization setting, we additionally report the harmonic mean (HM) of the accuracies on base and new classes. All the performances are reported on the test split of each dataset, unless stated otherwise.

\paragraph{Base-to-New Generalization.} In this setup (which corresponds to Sec. 4.1 of the main paper), we keep the participation of clients to 100\% and the number of classes per client, $K = 20$, similar to \cite{fedtpg} that produces 30 remote clients over 9 datasets. We set the batch size to 128, the training sample per class to 8, and the number of communication rounds to 200. 

\paragraph{Domain Generalization.} For both the Multi-source Single-target (MSST) and Single-source Multi-target (SSMT) domain generalization settings (corresponding to Sec. 4.2 of the main paper) on DomainBed benchmark, we keep the participation of clients to 100\%, shots to 8 and batch size to 128. However, we set the number of classes per client, $K = 2$, and global communication round to 20 for PACS, VLCS, and Terra Incognita datasets. In contrast, for OfficeHome and DomainNet datasets, we fix the number of classes per client, $K = 20$, and global communication rounds to 100.

For the DG setting on ImageNet benchmark, we follow the setup of \cite{fedtpg}, keeping participation of clients to 10\%, shots to 8, and number of classes per client, $K = 5$, for ImageNet training. In this case, we fix the batch size to 128 and the number of communication rounds to 200.

\paragraph{Cross-Dataset Generalization.} Similar to \cite{fedtpg}, we keep the participation of clients to 10\%, shots to 8, and the number of classes per client, $K = 5$ for ImageNet training, and perform the evaluation on 10 datasets. The batch size and number of communication rounds are fixed to 128 and 200, respectively.

\subsection{Detailed Results}

\paragraph{Qualitative results.} In Figs.~\ref{fig:top5_sun} and~\ref{fig:top5_office} we randomly pick few test samples and visualize the top-5 prediction probabilities of our \method and compare it with that of \fedtpg~\cite{fedtpg}.

In Fig.~\ref{fig:top5_sun} we report the top-5 predictions of our \method and \fedtpg for a few samples corresponding to the \bton generalization setting. On the top-left example, we notice that \fedtpg confuses a ``Restaurant'' with a ``Dining room'', two classes that share a lot of visual similarities, whereas our \method correctly classifies it as the ``Restaurant''. This behaviour can be attributed to the usage of attributes of our method that imparts a more fine-grained knowledge into the visual prompts. Interestingly, for the example of ``Mountain'' on top-right, while both the methods can predict the correct class, we see that for our \method, the second most confident class is ``Mountain Snowy'', which is more accurate for the given example. This can be attributed due to the use of attributes during visual prompt tuning, as a LLM would generate ``\textit{snow on mountains}'' as a characteristic attribute of a mountain. Thus, given enough training samples of mountains with snow, the model will even start recognizing snowy mountains, which is not the case for \fedtpg, where the second most confident prediction is ``Hill'' -- a more generic form of a mountainous landscape.

In Fig.~\ref{fig:top5_office} bottom-left example of the class ``Kettle'', both \fedtpg and \method predicts the correct class. However, the top-5 classes predicted by \fedtpg include completely unrelated classes such as ``Helmet'', ``Alarm Clock'' and ``Desk Lamp''. Whereas, in our \method we notice classes semantically similar to the class ``Kettle'' such as ``Bottle'', ``Mug'' and ``Bucket''. This indicates that the presence of the attributes helps the model to generalize across clients or domains unseen during training, and make more reasonable predictions, as long as objects share similar visual parts. However, in the bottom right example of Fig.~\ref{fig:top5_office} we observe that both the models get the prediction incorrect, but it is a more reasonable mistake as a ``File Cabinet'' shares a lot of visual similarity with the class ``Shelf'' which is the ground truth annotation. 

This underscores the importance of incorporating attributes during the visual prompt tuning step, enhancing the model's accuracy and ensuring that its errors remain reasonable even when the top-1 predictions are incorrect.

\subsection{Detailed quantitative results} In this section we report the detailed experimental results corresponding to the Tabs. 1, and 2 of the main paper, which are essentially the summarized versions of the tables in the supplementary material.

\paragraph{Base-to-New Generalization.} In Tab.~\ref{tab:b2n_detailed}, we present the expanded version of Tab. 1 of the main paper. Here, we showcase the base-to-new generalization performances of 9 datasets. In detail, we report the seen class accuracies, \ie, the local and base accuracy, the unseen class accuracy or the new accuracy, and the harmonic mean (HM) of the base and new class accuracies, separately for each dataset. Note that all the 9 datasets participates in the federated training set up. The Tab.~\ref{tab:b2n_detailed}(a) is the same as Tab. 1 of the main paper, and the Tabs.~\ref{tab:b2n_detailed}(b)-(h) reports the performance on each dataset separately.

From these tables on individual datasets we observe that our proposed \method outperforms the baselines in majority of the datasets, with a few exceptions. This demonstrates that the improvement brought by \method is consistent across datasets, and the average performance in Tab.~\ref{tab:b2n_detailed}(a) or Tab. 1 of the main paper is not dominated by some particular dataset. In summary, we have demonstrated that \method can successfully generalize on the base, \ie, combined classes from multiple clients, and completely new classes. Notably, \method has achieved better performances with significant margin on unseen classes, where other FL methods fail to do so.

\begin{figure}[h]
    \centering
    \begin{subfigure}[t]{0.48\linewidth}
        \centering
        \includegraphics[width=\linewidth]{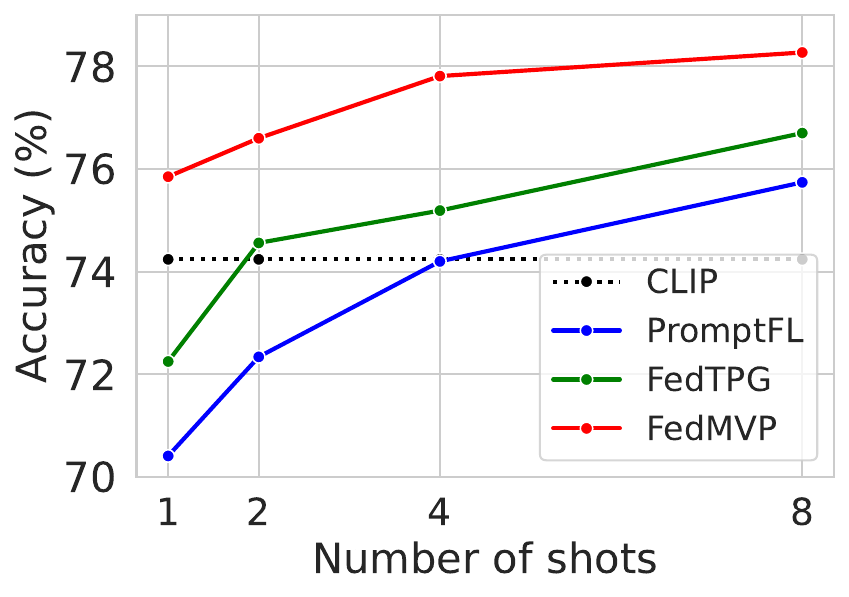}
        \caption{Base-to-New}
        \label{fig:shots_b2n}
    \end{subfigure}
    \hfill
    \begin{subfigure}[t]{0.48\linewidth}
        \centering
        \includegraphics[width=\linewidth]{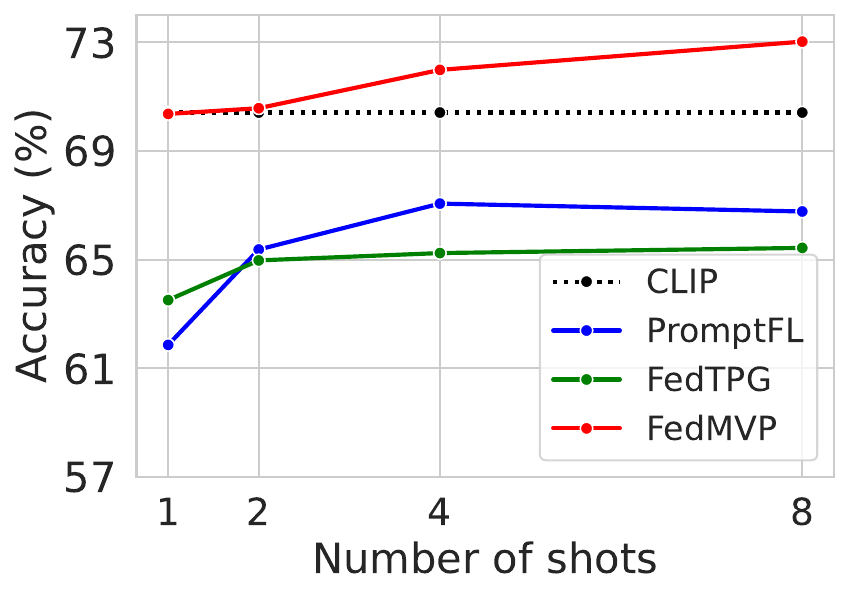}
        \caption{MSST DG}
        \label{fig:shots_msst}
    \end{subfigure}
    
    \caption{\textbf{Sensitivity to number of shots of images} on (a) Base-to-New and (b) MSST DG (DomainNet) setting. }
    \label{fig:shots}
    \vspace{-0.6cm}  
\end{figure}

\paragraph{Domain Generalization.} In Tab.~\ref{tab:pacsoffc_msst}, \ref{tab:vlcsterra_msst} \& \ref{tab:domainnet_msst} we present the detailed results of the summarized results of Tab. 2 of the main paper on the DomainBed benchmark in MSST DG setting. Here, the accuracy of a particular domain refer that the model is tested on that domain, while trained on rest of the domains. \method has shown superior performances over other FL methods in all of the datasets, except Terra Incognita, where FedCLIP and FedMaPLe are able to classify fine-grained animal classes better than others. 

In addition, we have provided the detailed results on the SSMT DG setting (reported in Tab. 2 of the main paper) for the datasets PACS (in Tab.~\ref{tab:pacs_ssmt}), OfficeHome in (Tab.~\ref{tab:office_ssmt}), VLCS (in Tab.~\ref{tab:vlcs_ssmt}), Terra Incognita (in Tab.~\ref{tab:terra_ssmt}) and DomainNet (in Tab.~\ref{tab:domainnet_ssmt}). We also notice similar trend, with our \method outperforming the baselines consistently on several datasets, with a few exceptions.

\begin{table}[!ht]
    \centering
    \caption{\textbf{Comparison of effects of different prompting methods used in \method on the Base-to-New, Multi-source Single-target (MSST) and Single-source Multi-target (SSMT) Domain Generalization settings}.}
    \scalebox{0.75}{\begin{tabular}{c|c|c|c}

\toprule
\textbf{Method}  &B2N &MSST & SSMT \\ \midrule
Textual Prompting &73.56 &66.26 &65.89 \\
Multi-modal Prompting &75.95 &69.14 &68.76 \\
Visual Prompting (ours) &\textbf{78.27} &\textbf{73.02} &\textbf{72.63} \\ \hline
\end{tabular}}
\label{tab:multimodal}
\end{table}
\vspace{-3.6mm}

\begin{table*}[!ht]
\centering
\caption{\textbf{Comparison of methods on the Base-to-new generalization setting}. Tables (b)-(k) report the performance on each dataset.}
\scalebox{0.78}{
    \begin{tabular}{lcccclcccc}
         \multicolumn{5}{c}{(a) \textbf{Average over 9 datasets}}&\multicolumn{5}{c}{(b) \textbf{Caltech101}}\\\cmidrule(lr){1-5}\cmidrule(lr){6-10}

         Method &Local &Base &\multicolumn{1}{c|}{New} &HM &Method &Local &Base &\multicolumn{1}{c|}{New} &HM\\\cmidrule(lr){1-5}\cmidrule(lr){6-10}
         
         ZS-CLIP \cite{radford2019language} &76.72 &70.51 &\multicolumn{1}{c|}{75.78} &\cellcolor[gray]{0.9}74.24 &ZS-CLIP \cite{radford2019language} &97.57 &96.97 &\multicolumn{1}{c|}{93.89} &\cellcolor[gray]{0.9}96.12 \\
         \cmidrule(lr){1-5}\cmidrule(lr){6-10}
        FedOTP \cite{fedotp} &74.82 &65.22 &\multicolumn{1}{c|}{57.04} &\cellcolor[gray]{0.9}64.89 &FedOTP \cite{fedotp}  &95.72 &94.83 &\multicolumn{1}{c|}{86.46} &\cellcolor[gray]{0.9}92.14 \\
        FedCoCoOp \cite{cocoop} &81.46	&73.76	&\multicolumn{1}{c|}{66.00}	&\cellcolor[gray]{0.9}73.20 &FedCoCoOp \cite{cocoop} &96.71 &94.41 &\multicolumn{1}{c|}{91.59} &\cellcolor[gray]{0.9}94.19 \\
        FedVPT \cite{vpt} &76.29	&70.43	&\multicolumn{1}{c|}{74.89}	&\cellcolor[gray]{0.9}73.79 &FedVPT \cite{vpt} &96.23	&95.31	&\multicolumn{1}{c|}{94.53}	&\cellcolor[gray]{0.9}95.35 \\

        FedCLIP \cite{fedclip} &76.87 &71.04 &\multicolumn{1}{c|}{75.06} &\cellcolor[gray]{0.9}74.24 &FedCLIP \cite{fedclip} &97.71 &97.29 &\multicolumn{1}{c|}{94.21} &\cellcolor[gray]{0.9}96.38 \\

        FedMaPLe \cite{maple} &81.63	&74.44	&\multicolumn{1}{c|}{70.62}	&\cellcolor[gray]{0.9}75.29 &FedMaPLe \cite{maple} &97.00 &95.41 &\multicolumn{1}{c|}{90.06} &\cellcolor[gray]{0.9}94.06 \\
        
        FedKgCoOp \cite{kgcoop} &78.38	&72.18	&\multicolumn{1}{c|}{75.87}	&\cellcolor[gray]{0.9}75.39 &FedKgCoOp \cite{kgcoop} &97.65 &97.24 &\multicolumn{1}{c|}{94.79} &\cellcolor[gray]{0.9}96.54 \\
        PromptFL \cite{promptfl} &81.75	&74.47	&\multicolumn{1}{c|}{71.70}	&\cellcolor[gray]{0.9}75.74 &PromptFL \cite{promptfl} &96.97 & 96.69 &\multicolumn{1}{c|}{92.79} &\cellcolor[gray]{0.9}95.44 \\
        FedTPG \cite{fedtpg} &80.75	&73.68	&\multicolumn{1}{c|}{76.02}	&\cellcolor[gray]{0.9}76.70 &FedTPG \cite{fedtpg} &97.59 &97.08 &\multicolumn{1}{c|}{95.24} &\cellcolor[gray]{0.9}96.63 \\
        
        \method (Ours) &\textbf{81.89}	&\textbf{75.37}	&\multicolumn{1}{c|}{\textbf{77.82}}	&\cellcolor[gray]{0.9}\textbf{78.27} &\method (Ours) &\textbf{97.85} &\textbf{97.73} &\multicolumn{1}{c|}{\textbf{95.48}} &\cellcolor[gray]{0.9}\textbf{97.01} \\
        
        \cmidrule(lr){1-5}\cmidrule(lr){6-10}

        &&&&&&&&&\\

        \multicolumn{5}{c}{(c) \textbf{Flowers102}}&\multicolumn{5}{c}{(d) \textbf{FGVCAircraft}}\\\cmidrule(lr){1-5}\cmidrule(lr){6-10}

         Method &Local &Base &\multicolumn{1}{c|}{New} &HM &Method &Local &Base &\multicolumn{1}{c|}{New} &HM\\\cmidrule(lr){1-5}\cmidrule(lr){6-10}
         
         ZS-CLIP \cite{radford2019language} &82.58 &72.18 &\multicolumn{1}{c|}{77.94} &\cellcolor[gray]{0.9}77.33 &ZS-CLIP \cite{radford2019language}  &30.59 &27.55 &\multicolumn{1}{c|}{35.81} &\cellcolor[gray]{0.9}30.96 \\
         \cmidrule(lr){1-5}\cmidrule(lr){6-10}
        FedOTP \cite{fedotp} &86.95 &65.90 &\multicolumn{1}{c|}{62.06} &\cellcolor[gray]{0.9}70.11 &FedOTP \cite{fedotp}  &28.35 &24.01 &\multicolumn{1}{c|}{15.53} &\cellcolor[gray]{0.9}21.23 \\

        FedCoCoOp \cite{cocoop} &94.00	&\textbf{77.49}	&\multicolumn{1}{c|}{65.63}	&\cellcolor[gray]{0.9}77.36 &FedCoCoOp \cite{cocoop} &35.21 &31.93 &\multicolumn{1}{c|}{22.67} &\cellcolor[gray]{0.9}28.89 \\

        FedVPT \cite{vpt} &81.67	&73.09	&\multicolumn{1}{c|}{76.10}	&\cellcolor[gray]{0.9}76.79 &FedVPT \cite{vpt} &31.36	&27.92	&\multicolumn{1}{c|}{32.67}	&\cellcolor[gray]{0.9}30.51 \\

        FedCLIP \cite{fedclip} &79.72 &71.51 &\multicolumn{1}{c|}{75.96} &\cellcolor[gray]{0.9}75.58 &FedCLIP \cite{fedclip} &31.41 &28.45 &\multicolumn{1}{c|}{34.07} &\cellcolor[gray]{0.9}31.14 \\
             
        FedMaPLe \cite{maple} &94.28	&76.44	&\multicolumn{1}{c|}{68.51}	&\cellcolor[gray]{0.9}78.36 &FedMaPLe \cite{maple} &35.83 &31.39 &\multicolumn{1}{c|}{32.34} &\cellcolor[gray]{0.9}33.08 \\
        
        FedKgCoOp \cite{kgcoop} &84.59	&72.11	&\multicolumn{1}{c|}{77.06}	&\cellcolor[gray]{0.9}77.59 &FedKgCoOp \cite{kgcoop} &33.68 &29.79 &\multicolumn{1}{c|}{34.01} &\cellcolor[gray]{0.9}32.38 \\
        PromptFL \cite{promptfl} &\textbf{94.44}	&76.40	&\multicolumn{1}{c|}{70.12}	&\cellcolor[gray]{0.9}79.07 &PromptFL \cite{promptfl} &\textbf{36.29} & 32.41 &\multicolumn{1}{c|}{30.95} &\cellcolor[gray]{0.9}33.07 \\
        FedTPG \cite{fedtpg} &90.76	&71.80	&\multicolumn{1}{c|}{77.76}	&\cellcolor[gray]{0.9}79.35 &FedTPG \cite{fedtpg} &34.68 &30.82 &\multicolumn{1}{c|}{35.18} &\cellcolor[gray]{0.9}33.44 \\

        \method (Ours) &94.05	&76.34	&\multicolumn{1}{c|}{\textbf{78.24}}	&\cellcolor[gray]{0.9}\textbf{82.16} &\method (Ours) &35.50 &\textbf{32.54} &\multicolumn{1}{c|}{\textbf{37.32}} &\cellcolor[gray]{0.9}\textbf{35.01} \\
        \cmidrule(lr){1-5}\cmidrule(lr){6-10}

        &&&&&&&&&\\

        \multicolumn{5}{c}{(e) \textbf{UCF101}}&\multicolumn{5}{c}{(f) \textbf{OxfordPets}}\\\cmidrule(lr){1-5}\cmidrule(lr){6-10}

         Method &Local &Base &\multicolumn{1}{c|}{New} &HM &Method &Local &Base &\multicolumn{1}{c|}{New} &HM\\\cmidrule(lr){1-5}\cmidrule(lr){6-10}
         
         ZS-CLIP \cite{radford2019language} &80.75 &70.58 &\multicolumn{1}{c|}{77.50} &\cellcolor[gray]{0.9}76.04 &ZS-CLIP \cite{radford2019language}  &91.33 &91.33 &\multicolumn{1}{c|}{97.04} &\cellcolor[gray]{0.9}93.16 \\
         \cmidrule(lr){1-5}\cmidrule(lr){6-10}
        FedOTP \cite{fedotp} &70.99 &59.61 &\multicolumn{1}{c|}{51.54} &\cellcolor[gray]{0.9}59.68 &FedOTP \cite{fedotp}  &88.62 &88.62 &\multicolumn{1}{c|}{71.92} &\cellcolor[gray]{0.9}82.26 \\

        FedCoCoOp \cite{cocoop} &84.92	&75.23	&\multicolumn{1}{c|}{64.25}	&\cellcolor[gray]{0.9}73.83 &FedCoCoOp \cite{cocoop} &92.34 &92.34 &\multicolumn{1}{c|}{87.36} &\cellcolor[gray]{0.9}90.62 \\

        FedVPT \cite{vpt} &82.43	&71.32	&\multicolumn{1}{c|}{77.05}	&\cellcolor[gray]{0.9}76.66 &FedVPT \cite{vpt} &89.70	&89.70	&\multicolumn{1}{c|}{96.73}	&\cellcolor[gray]{0.9}91.93 \\		
        
        FedCLIP \cite{fedclip} &79.67 &70.22 &\multicolumn{1}{c|}{75.66} &\cellcolor[gray]{0.9}74.98 &FedCLIP \cite{fedclip} &91.17 &91.18 &\multicolumn{1}{c|}{93.12} &\cellcolor[gray]{0.9}91.81 \\
         
        FedMaPLe \cite{maple} &84.17	&75.12	&\multicolumn{1}{c|}{68.68}	&\cellcolor[gray]{0.9}75.47 &FedMaPLe \cite{maple} &\textbf{95.00} &\textbf{95.00} &\multicolumn{1}{c|}{\textbf{97.09}} &\cellcolor[gray]{0.9}\textbf{95.69} \\
        FedKgCoOp \cite{kgcoop} &82.66	&73.14	&\multicolumn{1}{c|}{76.36}	&\cellcolor[gray]{0.9}77.19 &FedKgCoOp \cite{kgcoop} &91.58 &91.58 &\multicolumn{1}{c|}{96.53} &\cellcolor[gray]{0.9}93.17 \\
        PromptFL \cite{promptfl} &\textbf{86.13}	&75.65	&\multicolumn{1}{c|}{70.60}	&\cellcolor[gray]{0.9}76.94 &PromptFL \cite{promptfl} &93.31 & 93.32 &\multicolumn{1}{c|}{95.39} &\cellcolor[gray]{0.9}94.00 \\
        FedTPG \cite{fedtpg} &85.64	&74.89	&\multicolumn{1}{c|}{76.64}	&\cellcolor[gray]{0.9}78.79 &FedTPG \cite{fedtpg} &94.70 &94.69 &\multicolumn{1}{c|}{95.79} &\cellcolor[gray]{0.9}95.06 \\

        \method (Ours) &85.41	&\textbf{75.92}	&\multicolumn{1}{c|}{\textbf{80.25}}	&\cellcolor[gray]{0.9}\textbf{80.34} &\method (Ours) &94.05 &94.05 &\multicolumn{1}{c|}{96.12} &\cellcolor[gray]{0.9}94.73 \\
        \cmidrule(lr){1-5}\cmidrule(lr){6-10}

        &&&&&&&&&\\

        \multicolumn{5}{c}{(g) \textbf{Food101}}&\multicolumn{5}{c}{(h) \textbf{DTD}}\\\cmidrule(lr){1-5}\cmidrule(lr){6-10}

         Method &Local &Base &\multicolumn{1}{c|}{New} &HM &Method &Local &Base &\multicolumn{1}{c|}{New} &HM\\\cmidrule(lr){1-5}\cmidrule(lr){6-10}
         
         ZS-CLIP \cite{radford2019language} &94.39 &90.16 &\multicolumn{1}{c|}{91.25} &\cellcolor[gray]{0.9}91.90 &ZS-CLIP \cite{radford2019language}  &53.13 &53.01 &\multicolumn{1}{c|}{58.21} &\cellcolor[gray]{0.9}54.68 \\
         \cmidrule(lr){1-5}\cmidrule(lr){6-10}
        FedOTP \cite{fedotp} &87.06 &77.12 &\multicolumn{1}{c|}{69.09} &\cellcolor[gray]{0.9}77.07 &FedOTP \cite{fedotp}  &\textbf{69.21} &\textbf{69.21} &\multicolumn{1}{c|}{55.31} &\cellcolor[gray]{0.9}63.86 \\

        FedCoCoOp \cite{cocoop} &93.24	&87.57	&\multicolumn{1}{c|}{84.95}	&\cellcolor[gray]{0.9}88.45 &FedCoCoOp \cite{cocoop} &68.63 &68.63 &\multicolumn{1}{c|}{45.77} &\cellcolor[gray]{0.9}58.83 \\

        FedVPT \cite{vpt} &90.26	&88.39	&\multicolumn{1}{c|}{89.45}	&\cellcolor[gray]{0.9}89.36 &FedVPT \cite{vpt} &52.06	&52.06	&\multicolumn{1}{c|}{60.13}	&\cellcolor[gray]{0.9}54.50 \\
        
        FedCLIP \cite{fedclip} &94.23 &89.57 &\multicolumn{1}{c|}{90.67} &\cellcolor[gray]{0.9}91.45 &FedCLIP \cite{fedclip} &56.48 &56.48 &\multicolumn{1}{c|}{60.39} &\cellcolor[gray]{0.9}57.72 \\
        
        FedMaPLe \cite{maple} &93.95	&89.43	&\multicolumn{1}{c|}{89.60}	&\cellcolor[gray]{0.9}90.95 &FedMaPLe \cite{maple} &68.28 &68.28 &\multicolumn{1}{c|}{46.61} &\cellcolor[gray]{0.9}59.12 \\
        
        FedKgCoOp \cite{kgcoop} &94.19	&89.94	&\multicolumn{1}{c|}{91.81}	&\cellcolor[gray]{0.9}91.95 &FedKgCoOp \cite{kgcoop} &58.76 &58.75 &\multicolumn{1}{c|}{59.61} &\cellcolor[gray]{0.9}59.04 \\
        PromptFL \cite{promptfl} &93.52	&88.63	&\multicolumn{1}{c|}{88.47}	&\cellcolor[gray]{0.9}90.15 &PromptFL \cite{promptfl} &68.67 & 68.67 &\multicolumn{1}{c|}{52.74} &\cellcolor[gray]{0.9}62.39 \\
        FedTPG \cite{fedtpg} &94.09	&89.87	&\multicolumn{1}{c|}{91.64}	&\cellcolor[gray]{0.9}91.83 &FedTPG \cite{fedtpg} &63.62 &63.62 &\multicolumn{1}{c|}{60.51} &\cellcolor[gray]{0.9}62.55 \\
         
        \method (Ours) &\textbf{95.06}	&\textbf{91.89}	&\multicolumn{1}{c|}{\textbf{92.57}}	&\cellcolor[gray]{0.9}\textbf{93.15} &\method (Ours) &67.32 &67.32 &\multicolumn{1}{c|}{\textbf{64.96}} &\cellcolor[gray]{0.9}\textbf{66.51} \\
        \cmidrule(lr){1-5}\cmidrule(lr){6-10}

        &&&&&&&&&\\

        \multicolumn{5}{c}{(i) \textbf{StanfordCars}}&\multicolumn{5}{c}{(j) \textbf{SUN397}}\\\cmidrule(lr){1-5}\cmidrule(lr){6-10}

         Method &Local &Base &\multicolumn{1}{c|}{New} &HM &Method &Local &Base &\multicolumn{1}{c|}{New} &HM\\\cmidrule(lr){1-5}\cmidrule(lr){6-10}
         
         ZS-CLIP \cite{radford2019language} &71.51 &63.44 &\multicolumn{1}{c|}{74.90} &\cellcolor[gray]{0.9}69.61 &ZS-CLIP \cite{radford2019language}  &88.66 &69.41 &\multicolumn{1}{c|}{75.46} &\cellcolor[gray]{0.9}77.05 \\
         \cmidrule(lr){1-5}\cmidrule(lr){6-10}
        FedOTP \cite{fedotp} &58.89 &45.25 &\multicolumn{1}{c|}{44.00} &\cellcolor[gray]{0.9}48.54 &FedOTP \cite{fedotp} &87.54 &62.39 &\multicolumn{1}{c|}{57.42} &\cellcolor[gray]{0.9}66.87 \\

        FedCoCoOp \cite{cocoop} &\textbf{76.62}	&\textbf{66.51}	&\multicolumn{1}{c|}{66.40}	&\cellcolor[gray]{0.9}69.53 &FedCoCoOp \cite{cocoop} &91.44 &69.76 &\multicolumn{1}{c|}{65.36} &\cellcolor[gray]{0.9}73.94 \\

        FedVPT \cite{vpt} &73.47	&65.98	&\multicolumn{1}{c|}{71.47}	&\cellcolor[gray]{0.9}70.16 &FedVPT \cite{vpt} &89.43	&70.13	&\multicolumn{1}{c|}{75.92}	&\cellcolor[gray]{0.9}77.69 \\

        FedCLIP \cite{fedclip} &72.32 &64.42 &\multicolumn{1}{c|}{75.04} &\cellcolor[gray]{0.9}70.30 &FedCLIP \cite{fedclip} &89.10 &70.22 &\multicolumn{1}{c|}{76.42} &\cellcolor[gray]{0.9}77.82 \\
              
        FedMaPLe \cite{maple}&74.76	&66.26	&\multicolumn{1}{c|}{71.33}	&\cellcolor[gray]{0.9}70.61 &FedMaPLe \cite{maple} &91.40 &72.66 &\multicolumn{1}{c|}{71.33} &\cellcolor[gray]{0.9}77.47 \\
        
        FedKgCoOp \cite{kgcoop} &71.89	&64.33	&\multicolumn{1}{c|}{75.71}	&\cellcolor[gray]{0.9}70.32 &FedKgCoOp \cite{kgcoop} &90.38 &72.72 &\multicolumn{1}{c|}{76.94} &\cellcolor[gray]{0.9}79.34 \\
        PromptFL \cite{promptfl} &74.53	&66.16	&\multicolumn{1}{c|}{72.32}	&\cellcolor[gray]{0.9}70.82 &PromptFL \cite{promptfl} &91.93 &72.34 &\multicolumn{1}{c|}{71.89} &\cellcolor[gray]{0.9}77.70 \\
        FedTPG \cite{fedtpg} &74.54	&66.34	&\multicolumn{1}{c|}{74.26}	&\cellcolor[gray]{0.9}71.50 &FedTPG \cite{fedtpg} &91.11 &74.01 &\multicolumn{1}{c|}{77.13} &\cellcolor[gray]{0.9}80.10 \\
         
        \method (Ours) &75.30	&66.45	&\multicolumn{1}{c|}{\textbf{75.94}}	&\cellcolor[gray]{0.9}\textbf{72.29} &\method (Ours) &\textbf{92.44} &\textbf{76.09} &\multicolumn{1}{c|}{\textbf{79.51}} &\cellcolor[gray]{0.9}\textbf{82.11} \\
        \cmidrule(lr){1-5}\cmidrule(lr){6-10}
        
    \end{tabular}}
    \label{tab:b2n_detailed}
\end{table*}

\begin{table*}[!ht]
    \centering
    \caption{\textbf{Comparison of methods on the Multi-source Single-target (MSST) Domain Generalization setting}. The results are reported for the PACS and OfficeHome datasets.}
    \scalebox{0.75}{
    \begin{tabular}{lcccc|c||cccc|c}
    \toprule
        & \multicolumn{5}{c}{\textbf{PACS}} & \multicolumn{5}{c}{\textbf{OfficeHome}} \\ \cmidrule(lr){2-6} \cmidrule(lr){7-11}
        
        \multirow{-2}{*}{\textbf{Method}}  & {A. Painting} & {Cartoon} & {Photo} & {Sketch} & {Average} & {Art} & {Clipart} & {Product} & {RealWorld} & {Average} \\ 
        \midrule

        ZS-CLIP \cite{radford2019language}  &97.55	&98.72	&100.00	&88.38	&\cellcolor[gray]{0.9}96.16	&80.63	&67.25	&87.99	&90.10	&\cellcolor[gray]{0.9}81.49 \\ 
        \midrule
        
         FedOTP \cite{fedotp}  &93.49	&93.75	&98.00	&77.61	&\cellcolor[gray]{0.9}90.71 &71.98	&65.33	&83.08	&85.28	&\cellcolor[gray]{0.9}76.42 \\

         FedCoCoOp \cite{cocoop}  &81.43	&92.05	&81.67	&85.07	&\cellcolor[gray]{0.9}85.06 &78.54	&66.73	&90.18	&90.24	&\cellcolor[gray]{0.9}81.42 \\

         FedTPG \cite{fedtpg} &90.55	&95.17	&89.84	&88.38	&\cellcolor[gray]{0.9}90.99 &80.63	&68.56	&90.78	&91.14	&\cellcolor[gray]{0.9}82.78 \\

         PromptFL \cite{promptfl} &96.42	&97.72	&99.80	&87.53	&\cellcolor[gray]{0.9}95.37 &79.61	&66.70	&90.42	&90.22	&\cellcolor[gray]{0.9}81.74 \\
         
        FedKgCoOp \cite{kgcoop} &96.48	&97.77	&99.83 &87.58	&\cellcolor[gray]{0.9}95.42 &79.67	&66.87	&90.48	&90.26	&\cellcolor[gray]{0.9}81.82 \\

        FedMaPLe \cite{maple} &90.72	&97.44	&99.80	&90.08	&\cellcolor[gray]{0.9}94.51 &79.67	&68.34	&90.33	&89.80	&\cellcolor[gray]{0.9}82.03 \\

        FedVPT \cite{vpt} &95.73	&96.37	&100.00	&89.32	&\cellcolor[gray]{0.9}95.36 &80.94	&68.21	&88.56	&89.34	&\cellcolor[gray]{0.9}81.76 \\

        FedCLIP \cite{fedclip}  &\textbf{97.56}	&98.72	&100.00	&88.89	&\cellcolor[gray]{0.9}96.29 &80.50	&67.26	&89.05	&90.15	&\cellcolor[gray]{0.9}81.74 \\
       
        \method (Ours) &96.92	&\textbf{99.35}	&\textbf{100.00} &\textbf{92.86}	&\cellcolor[gray]{0.9}\textbf{97.28} &\textbf{82.20}	&\textbf{70.05}	&\textbf{91.78}	&\textbf{92.56}	&\cellcolor[gray]{0.9}\textbf{84.15} \\
    \bottomrule
        
    \end{tabular}}
    \label{tab:pacsoffc_msst}
\end{table*}

\begin{table*}[!ht]
    \centering
    \caption{\textbf{Comparison of methods on the Multi-source Single-target (MSST) Domain Generalization setting}. The results are reported for the VLCS and Terra Incognita datasets.}
    \scalebox{0.75}{
    \begin{tabular}{lcccc|c||cccc|c}
    \toprule
        & \multicolumn{5}{c}{\textbf{VLCS}} & \multicolumn{5}{c}{\textbf{Terra Incognita}} \\ \cmidrule(lr){2-6} \cmidrule(lr){7-11}
        
        \multirow{-2}{*}{\textbf{Method}}  & {Caltech} & {Labelme} & {Pascal-VOC} & {Sun} & {Average} & {L38} & {L43} & {L46} & {L100} & {Average} \\ 
        \midrule

        ZS-CLIP \cite{radford2019language}  &100.00	&\textbf{68.88}	&89.24	&75.02	&\cellcolor[gray]{0.9}83.29 &20.14	&33.64	&29.19	&52.93	&\cellcolor[gray]{0.9}33.98 \\ 
        \midrule
        
         FedOTP \cite{fedotp}  &86.79	&57.21	&62.98	&62.64	&\cellcolor[gray]{0.9}67.41 &4.30	&27.92	&16.94	&3.82	&\cellcolor[gray]{0.9}13.24 \\

         FedCoCoOp \cite{cocoop}  &54.95	&58.59	&65.35	&68.02	&\cellcolor[gray]{0.9}61.73 &10.58	&8.87	&19.55	&55.73	&\cellcolor[gray]{0.9}23.68 \\

         FedTPG \cite{fedtpg} &88.92	&61.48	&62.98	&65.69	&\cellcolor[gray]{0.9}69.77 &\textbf{46.11}	&19.14	&21.78	&20.12	&\cellcolor[gray]{0.9}26.79 \\

         PromptFL \cite{promptfl} &98.82	&55.71	&75.42	&69.54	&\cellcolor[gray]{0.9}74.87 &14.35	&14.62	&21.60	&49.51	&\cellcolor[gray]{0.9}25.02 \\
         
        FedKgCoOp \cite{kgcoop} &98.87	&55.80	&75.36	&69.57	&\cellcolor[gray]{0.9}74.90 &14.37	&14.66	&21.57	&49.52	&\cellcolor[gray]{0.9}25.03 \\
       
        FedMaPLe \cite{maple} &99.53	&52.70	&71.08	&63.86	&\cellcolor[gray]{0.9}71.79 &30.10	&33.64	&28.81	&52.66	&\cellcolor[gray]{0.9}36.30 \\

        FedVPT \cite{vpt} &99.45	&69.47	&89.87	&73.98	&\cellcolor[gray]{0.9}83.19 &20.46	&34.76	&26.08	&53.17	&\cellcolor[gray]{0.9}33.62 \\

        FedCLIP \cite{fedclip}  &100.00	&67.88	&87.48	&75.43	&\cellcolor[gray]{0.9}82.70 &25.60	&\textbf{35.21}	&29.25	&56.28	&\cellcolor[gray]{0.9}36.58 \\
       
        \method (Ours) &\textbf{100.00}	&70.15	&\textbf{90.93}	&\textbf{79.40}	&\cellcolor[gray]{0.9}\textbf{85.12} &23.95	&34.95	&\textbf{33.19}	&\textbf{57.34}	&\cellcolor[gray]{0.9}\textbf{37.36} \\
       
    \bottomrule
        
    \end{tabular}}
    \label{tab:vlcsterra_msst}
\end{table*}

\begin{table*}[!ht]
    \centering
    \caption{\textbf{Comparison of methods on the Multi-source Single-target (MSST) Domain Generalization setting}. The results are reported for the DomainNet dataset.}
    \scalebox{0.75}{
    \begin{tabular}{lcccccc|c}
    \toprule
        & \multicolumn{7}{c}{\textbf{DomainNet}} \\ \cmidrule(lr){2-8} 
        
        \multirow{-2}{*}{\textbf{Method}}  & {Clipart} & {Infograph} & {Painting} & {Quickdraw} & {Real} & {Sketch} & {Average} \\ 
        \midrule

        ZS-CLIP \cite{radford2019language}  &70.88	&45.94	&66.27	&14.19	&83.22	&62.25	&\cellcolor[gray]{0.9}57.13 \\ 
        \midrule
        
         FedOTP \cite{fedotp}  &64.33	&38.42	&54.60	&11.63	&73.45	&55.61	&\cellcolor[gray]{0.9}49.67 \\

         FedCoCoOp \cite{cocoop}  &70.84	&46.26	&65.90	&14.35	&83.13	&62.00	&\cellcolor[gray]{0.9}57.08 \\

         FedTPG \cite{fedtpg} &71.35	&46.03	&66.10	&13.96	&81.50	&61.97	&\cellcolor[gray]{0.9}56.82 \\

         PromptFL \cite{promptfl} &70.95	&45.98	&65.57	&13.80	&82.76	&62.15	&\cellcolor[gray]{0.9}56.87 \\
         
        FedKgCoOp \cite{kgcoop} &70.96	&46.00	&66.02	&13.83	&82.83	&61.90	&\cellcolor[gray]{0.9}56.92 \\
       
        FedMaPLe \cite{maple} &72.91	&49.93	&67.13	&16.20	&82.73	&64.36	&\cellcolor[gray]{0.9}58.88 \\

        FedVPT \cite{vpt} &71.74	&42.57	&61.47	&13.65	&83.66	&62.80	&\cellcolor[gray]{0.9}55.98 \\

        FedCLIP \cite{fedclip}  &71.88	&46.46	&67.09	&15.13	&83.56	&62.96	&\cellcolor[gray]{0.9}57.85 \\
       
        \method (Ours) &\textbf{73.93}	&\textbf{52.06}	&\textbf{69.16} &\textbf{18.50}	&\textbf{86.64}	&\textbf{66.72}	&\cellcolor[gray]{0.9}\textbf{61.17} \\
    \bottomrule
        
    \end{tabular}}
    \label{tab:domainnet_msst}
\end{table*}

\begin{table*}[!ht]
    \centering
    \caption{\textbf{Comparison of methods on the Single-source Multi-target (SSMT) Domain Generalization setting}. The results are reported for the PACS dataset. Here Ap, Cr, Ph, Sk denote to Art painting, Cartoon, Photo and Sketch domains respectively.}
    \scalebox{0.65}{
    \begin{tabular}{lccc|c||ccc|c||ccc|c||ccc|c}
    \toprule
        & \multicolumn{4}{c}{\textbf{Ap}} & \multicolumn{4}{c}{\textbf{Cr}} & \multicolumn{4}{c}{\textbf{Ph}} & \multicolumn{4}{c}{\textbf{Sk}} \\ \cmidrule(lr){2-5} \cmidrule(lr){6-9} \cmidrule(lr){10-13} \cmidrule(lr){14-17}
        
        \multirow{-2}{*}{\textbf{Method}} & {Cr} & {Ph} & {Sk} &{Avg.} & {Ap} & {Ph} & {Sk} &{Avg.} & {Ap} & {Cr} &{Sk} & {Avg.} &{Ap} &{Cr} &{Ph} & {Avg.} \\ 
        \midrule

        ZS-CLIP \cite{radford2019language} &98.72 &100.00	&88.38	&\cellcolor[gray]{0.9}95.70	&97.55	&100.00	&88.38	&\cellcolor[gray]{0.9}95.31	&97.55	&98.72	&88.38	&\cellcolor[gray]{0.9}94.88	&97.55	&98.72	&100.00	&\cellcolor[gray]{0.9}98.76\\ 
        \midrule
        
         FedOTP \cite{fedotp} &93.75	&98.21	&82.36	&\cellcolor[gray]{0.9}91.44	&91.69	&98.61	&76.42	&\cellcolor[gray]{0.9}88.91	&90.39	&94.74	&78.63	&\cellcolor[gray]{0.9}87.92	&94.46	&95.99	&98.81	&\cellcolor[gray]{0.9}96.42\\

         FedCoCoOp \cite{cocoop} &92.90	&80.88	&85.58	&\cellcolor[gray]{0.9}86.45	&79.97	&79.88	&84.73	&\cellcolor[gray]{0.9}81.53	&79.97	&90.77	&84.48	&\cellcolor[gray]{0.9}85.07	&81.76	&91.90	&81.87	&\cellcolor[gray]{0.9}85.18\\

        FedTPG \cite{fedtpg} &93.89	&84.46	&88.63	&\cellcolor[gray]{0.9}88.99	&89.90	&91.03	&88.04	&\cellcolor[gray]{0.9}89.66	&91.04	&95.31	&87.87	&\cellcolor[gray]{0.9}91.41	&90.88	&95.60	&91.83	&\cellcolor[gray]{0.9}92.77\\

        PromptFL \cite{promptfl} &98.72	&100.00	&92.11	&\cellcolor[gray]{0.9}96.94	&96.09	&98.60	&88.97	&\cellcolor[gray]{0.9}94.55	&94.13	&97.02	&92.36	&\cellcolor[gray]{0.9}94.50	&96.41	&96.31	&99.20	&\cellcolor[gray]{0.9}97.31\\
         
        FedKgCoOp \cite{kgcoop} &98.65	&99.89	&92.15	&\cellcolor[gray]{0.9}96.90	&96.15	&98.63	&88.91	&\cellcolor[gray]{0.9}94.56	&94.24	&96.87	&92.40	&\cellcolor[gray]{0.9}94.50	&96.35	&96.25	&99.27	&\cellcolor[gray]{0.9}97.29\\      
       
        FedMaPLe \cite{maple} &97.59	&99.80	&90.25	&\cellcolor[gray]{0.9}95.88	&90.39	&99.80	&90.16	&\cellcolor[gray]{0.9}93.45	&90.88	&97.73	&90.08	&\cellcolor[gray]{0.9}92.89	&90.23	&97.30	&96.81	&\cellcolor[gray]{0.9}94.78\\

        FedVPT \cite{vpt} &97.33	&99.53	&84.24	&\cellcolor[gray]{0.9}93.70	&95.44	&98.25	&85.89	&\cellcolor[gray]{0.9}93.19	&97.85	&98.67	&86.13	&\cellcolor[gray]{0.9}94.22	&96.56	&98.06	&99.47	&\cellcolor[gray]{0.9}98.03 \\

        FedCLIP \cite{fedclip} &98.72	&100.00	&88.89	&\cellcolor[gray]{0.9}95.87	&\textbf{97.56}	&100.00	&88.89	&\cellcolor[gray]{0.9}95.48	&\textbf{97.56}	&98.72	&88.89	&\cellcolor[gray]{0.9}95.06	&\textbf{97.56}	&98.72	&100.00	&\cellcolor[gray]{0.9}98.76\\
       
        \method (Ours) &\textbf{99.02}	&\textbf{100.00}	&\textbf{93.15}	&\cellcolor[gray]{0.9}\textbf{97.39}	&96.45	&\textbf{100.00}	&\textbf{92.19}	&\cellcolor[gray]{0.9}\textbf{96.21}	&96.87	&\textbf{98.91}	&\textbf{92.56}	&\cellcolor[gray]{0.9}\textbf{96.11}	&97.49	&\textbf{99.40}	&\textbf{100.00}	&\cellcolor[gray]{0.9}\textbf{98.96}\\
    \bottomrule
        
    \end{tabular}}
    \label{tab:pacs_ssmt}
\end{table*}

\begin{table*}[!ht]
    \centering
    \caption{\textbf{Comparison of methods on the Single-source Multi-target (SSMT) Domain Generalization setting}. The results are reported for the OfficeHome dataset. Here Ar, Cl, Pr, Rw denote to Art, Clipart, Product and RealWorld domains respectively.}
    \scalebox{0.65}{
    \begin{tabular}{lccc|c||ccc|c||ccc|c||ccc|c}
    \toprule
        & \multicolumn{4}{c}{\textbf{Ar}} & \multicolumn{4}{c}{\textbf{Cl}} & \multicolumn{4}{c}{\textbf{Pr}} & \multicolumn{4}{c}{\textbf{Rw}} \\ \cmidrule(lr){2-5} \cmidrule(lr){6-9} \cmidrule(lr){10-13} \cmidrule(lr){14-17}
        
        \multirow{-2}{*}{\textbf{Method}} & {Cl} & {Pr} & {Rw} &{Avg.} & {Ar} & {Pr} & {Rw} &{Avg.} & {Ar} & {Cl} &{Rw} & {Avg.} &{Ar} &{Cl} &{Pr} & {Avg.} \\ 
        \midrule

        ZS-CLIP \cite{radford2019language} &67.25 &87.99 &90.10 &\cellcolor[gray]{0.9}81.78 &80.63 &87.99 &90.10 &\cellcolor[gray]{0.9}86.24 &80.63 &67.25 &90.10 &\cellcolor[gray]{0.9}79.33 &80.63 &67.25 &87.99 &\cellcolor[gray]{0.9}78.62 \\ 
        \midrule
        
        FedOTP \cite{fedotp} &63.33 &80.97 &83.59 &\cellcolor[gray]{0.9}75.96 &69.78 &80.36 &80.75 &\cellcolor[gray]{0.9}76.97 &68.00 &62.33 &84.75 &\cellcolor[gray]{0.9}71.69 &71.29 &64.87 &84.37 &\cellcolor[gray]{0.9}73.51 \\

        FedCoCoOp \cite{cocoop}  &36.98 &55.21 &54.37 &\cellcolor[gray]{0.9}48.85 &52.20 &53.70 &53.99 &\cellcolor[gray]{0.9}53.30 &57.00 &39.98 &60.74 &\cellcolor[gray]{0.9}52.57 &52.20 &37.44 &56.65 &\cellcolor[gray]{0.9}48.76 \\

        FedTPG \cite{fedtpg} &68.41 &90.33 &91.18 &\cellcolor[gray]{0.9}83.31 &81.04 &90.78 &91.26 &\cellcolor[gray]{0.9}87.69 &79.94 &\textbf{69.10} &90.33 &\cellcolor[gray]{0.9}79.79 &80.35 &68.49 &90.03 &\cellcolor[gray]{0.9}79.62 \\

        PromptFL \cite{promptfl} &66.79 &89.95 &90.80 &\cellcolor[gray]{0.9}82.51 &80.49 &90.26 &90.64 &\cellcolor[gray]{0.9}87.13 &79.26 &66.25 &90.34 &\cellcolor[gray]{0.9}78.62 &80.77 &65.95 &89.12 &\cellcolor[gray]{0.9}78.61 \\

        FedKgCoOp \cite{kgcoop} &66.83 &90.02 &90.85 &\cellcolor[gray]{0.9}82.57 &80.41 &90.29 &90.56 &\cellcolor[gray]{0.9}87.09 &79.45 &66.12 &90.56 &\cellcolor[gray]{0.9}78.71 &80.61 &65.72 &89.02 &\cellcolor[gray]{0.9}78.45 \\

        FedMaPLe \cite{maple} &66.03 &89.35 &88.73 &\cellcolor[gray]{0.9}81.37 &75.14 &86.18 &85.20 &\cellcolor[gray]{0.9}82.17 &\textbf{90.08} &67.26 &89.19 &\cellcolor[gray]{0.9}\textbf{82.18} &\textbf{82.69} &68.03 &89.88 &\cellcolor[gray]{0.9}80.20 \\

        FedVPT \cite{vpt} &68.05	&88.25	&89.50	&\cellcolor[gray]{0.9}81.93	&78.46	&84.67	&90.89	&\cellcolor[gray]{0.9}84.67	&79.32	&68.94	&90.35	&\cellcolor[gray]{0.9}79.54	&80.87	&66.30	&86.29	&\cellcolor[gray]{0.9}77.82 \\

        FedCLIP \cite{fedclip} &65.95 &88.07 &90.49 &\cellcolor[gray]{0.9}81.50 &79.95 &88.90 &90.57 &\cellcolor[gray]{0.9}86.47 &80.91 &68.26 &91.41 &\cellcolor[gray]{0.9}80.19 &80.22 &66.72 &87.99 &\cellcolor[gray]{0.9}78.31 \\
       
        \method (Ours) &\textbf{70.08} &\textbf{91.76} &\textbf{92.14} &\cellcolor[gray]{0.9}\textbf{84.66} &\textbf{81.88} &\textbf{92.00} &\textbf{92.21} &\cellcolor[gray]{0.9}\textbf{88.70} &80.79 &69.03 &\textbf{92.35} &\cellcolor[gray]{0.9}80.72 &82.16 &\textbf{70.25} &\textbf{92.07} &\cellcolor[gray]{0.9}\textbf{81.49} \\
    \bottomrule
    \end{tabular}}
    \label{tab:office_ssmt}
\end{table*}

\begin{table*}[!ht]
    \centering
    \caption{\textbf{Comparison of methods on the Single-source Multi-target (SSMT) Domain Generalization setting}. The results are reported for the VLCS dataset. Here C, L, V, S denote to Caltech, LabelMe, Pascal-VOC and SUN domains respectively.}
    \scalebox{0.65}{
    \begin{tabular}{lccc|c||ccc|c||ccc|c||ccc|c}
    \toprule
        & \multicolumn{4}{c}{\textbf{C}} & \multicolumn{4}{c}{\textbf{L}} & \multicolumn{4}{c}{\textbf{V}} & \multicolumn{4}{c}{\textbf{S}} \\ \cmidrule(lr){2-5} \cmidrule(lr){6-9} \cmidrule(lr){10-13} \cmidrule(lr){14-17}
        
        \multirow{-2}{*}{\textbf{Method}} & {L} & {V} & {S} &{Avg.} & {C} & {V} & {S} &{Avg.} & {C} & {L} &{S} & {Avg.} &{C} &{L} &{V} &{Avg.} \\ 
        \midrule

        ZS-CLIP \cite{radford2019language} &68.88 &89.24 &75.02 &\cellcolor[gray]{0.9}77.71 &100.00 &89.24 &75.02 &\cellcolor[gray]{0.9}88.09 &\textbf{100.00} &68.88 &75.02 &\cellcolor[gray]{0.9}81.30 &\textbf{100.00} &68.88 &\textbf{89.24} &\cellcolor[gray]{0.9}\textbf{86.04} \\ 
        \midrule
        
         FedOTP \cite{fedotp} &58.09 &69.89 &63.49 &\cellcolor[gray]{0.9}63.83 &69.10 &54.00 &52.69 &\cellcolor[gray]{0.9}58.60 &98.82 &62.54 &71.70 &\cellcolor[gray]{0.9}77.69 &75.00 &49.69 &56.86 &\cellcolor[gray]{0.9}60.52 \\

         FedCoCoOp \cite{cocoop} &57.34 &67.42 &68.63 &\cellcolor[gray]{0.9}64.46 &50.94 &62.39 &67.21 &\cellcolor[gray]{0.9}60.18 &56.84 &55.83 &67.21 &\cellcolor[gray]{0.9}59.96 &56.84 &58.59 &65.75 &\cellcolor[gray]{0.9}60.39 \\

        FedTPG \cite{fedtpg} &70.89 &81.93 &62.84 &\cellcolor[gray]{0.9}71.89 &46.22 &56.56 &59.08 &\cellcolor[gray]{0.9}53.95 &98.35 &65.12 &72.49 &\cellcolor[gray]{0.9}78.65 &79.95 &51.69 &66.43 &\cellcolor[gray]{0.9}66.02 \\

        PromptFL \cite{promptfl} &61.94 &80.12 &67.05 &\cellcolor[gray]{0.9}69.70 &45.20 &60.97 &59.13 &\cellcolor[gray]{0.9}55.10 &98.78 &36.60 &56.09 &\cellcolor[gray]{0.9}63.82 &99.23 &59.76 &78.82 &\cellcolor[gray]{0.9}79.27 \\

        FedKgCoOp \cite{kgcoop} &61.98 &80.45 &67.11 &\cellcolor[gray]{0.9}69.85 &45.28 &61.10 &64.16 &\cellcolor[gray]{0.9}56.85 &98.82 &36.76 &56.04 &\cellcolor[gray]{0.9}63.87 &99.29 &59.85 &79.07 &\cellcolor[gray]{0.9}79.40 \\

        FedMaPLe \cite{maple} &66.00 &76.11 &65.99 &\cellcolor[gray]{0.9}69.37 &87.50 &57.65 &59.59 &\cellcolor[gray]{0.9}68.25 &99.29 &53.20 &66.90 &\cellcolor[gray]{0.9}73.13 &99.53 &56.34 &76.90 &\cellcolor[gray]{0.9}77.59 \\

        FedVPT \cite{vpt} &69.36	&89.68	&75.93	&\cellcolor[gray]{0.9}78.32	&98.45	&88.60	&75.34	&\cellcolor[gray]{0.9}87.46	&98.78	&68.60	&73.98	&\cellcolor[gray]{0.9}80.45	&96.59	&66.27	&87.67	&\cellcolor[gray]{0.9}83.51 \\

        FedCLIP \cite{fedclip} &68.63 &87.07 &76.04 &\cellcolor[gray]{0.9}77.25 &\textbf{100.00} &85.89 &75.74 &\cellcolor[gray]{0.9}87.21 &\textbf{100.00} &60.48 &75.03 &\cellcolor[gray]{0.9}78.50 &\textbf{100.00} &66.63 &86.18 &\cellcolor[gray]{0.9}84.27 \\

        \method (Ours) &\textbf{70.93} &\textbf{90.40} &\textbf{78.13} &\cellcolor[gray]{0.9}\textbf{79.82} &\textbf{100.00} &\textbf{90.54} &\textbf{78.37} &\cellcolor[gray]{0.9}\textbf{89.64} &\textbf{100.00} &\textbf{70.06} &\textbf{78.93} &\cellcolor[gray]{0.9}\textbf{83.00} &\textbf{100.00} &\textbf{69.44} &88.51 &\cellcolor[gray]{0.9}85.98 \\
    \bottomrule
    \end{tabular}}
    \label{tab:vlcs_ssmt}
\end{table*}

\begin{table*}[!ht]
    \centering
    \caption{\textbf{Comparison of methods on the Single-source Multi-target (SSMT) Domain Generalization setting}. The results are reported for the Terra Incognita dataset. Here L38, L43, L46, L100 denote to Location-38, Location-43, Location-46 and Location-100 domains respectively.}
    \scalebox{0.65}{
    \begin{tabular}{lccc|c||ccc|c||ccc|c||ccc|c}
    \toprule
        & \multicolumn{4}{c}{\textbf{L38}} & \multicolumn{4}{c}{\textbf{L43}} & \multicolumn{4}{c}{\textbf{L46}} & \multicolumn{4}{c}{\textbf{L100}} \\ \cmidrule(lr){2-5} \cmidrule(lr){6-9} \cmidrule(lr){10-13} \cmidrule(lr){14-17}
        
        \multirow{-2}{*}{\textbf{Method}} & {L43} & {L46} & {L100} &{Avg.} & {L38} & {L46} & {L100} &{Avg.} & {L38} & {L43} &{L100} & {Avg.} &{L38} &{L43} &{L46} &{Avg.} \\ 
        \midrule

        CLIP \cite{radford2019language} &33.64	&29.19	&52.93	&\cellcolor[gray]{0.9}38.59	&20.14	&29.19	&52.93	&\cellcolor[gray]{0.9}34.09	&20.14	&33.64	&52.93	&\cellcolor[gray]{0.9}35.57	&20.14	&33.64	&29.19	&\cellcolor[gray]{0.9}27.66 \\ 
        \midrule
        
         FedOTP \cite{fedotp} &27.42	&17.05	&1.84	&\cellcolor[gray]{0.9}15.44	&28.12	&19.34	&9.62 &\cellcolor[gray]{0.9}19.03 &0.38	&28.50	&3.82	&\cellcolor[gray]{0.9}10.90	&0.99	&27.76	&16.45	&\cellcolor[gray]{0.9}15.06 \\ 

         FedCoCoOp \cite{cocoop} &14.00	&14.71	&51.50	&\cellcolor[gray]{0.9}26.74	&6.59	&29.41	&54.98	&\cellcolor[gray]{0.9}30.33	&9.86	&7.87	&46.93	&\cellcolor[gray]{0.9}21.55	&10.27	&14.17	&16.50	&\cellcolor[gray]{0.9}13.65 \\ 

        FedTPG \cite{fedtpg} &20.21	&13.29	&10.98	&\cellcolor[gray]{0.9}14.83	&\textbf{38.46}	&15.68 &11.32	&\cellcolor[gray]{0.9}21.82	&\textbf{45.42} &12.34	&15.41	&\cellcolor[gray]{0.9}24.39	&\textbf{46.10}	&19.47	&21.40	&\cellcolor[gray]{0.9}28.99 \\ 

        PromptFL \cite{promptfl} &19.97	&13.40	&45.70	&\cellcolor[gray]{0.9}26.36	&21.81	&11.38	&24.21	&\cellcolor[gray]{0.9}19.13	&19.73	&14.08	&57.50	&\cellcolor[gray]{0.9}30.44	&29.73	&15.08	&17.43	&\cellcolor[gray]{0.9}20.75 \\ 

        FedKgCoOp \cite{kgcoop} &19.94	&13.44	&45.72	&\cellcolor[gray]{0.9}26.37	&21.83	&11.42	&24.22	&\cellcolor[gray]{0.9}19.16	&19.73	&14.09	&57.55	&\cellcolor[gray]{0.9}30.46	&29.78	&15.14	&17.48	&\cellcolor[gray]{0.9}20.80 \\       

        FedMaPLe \cite{maple} &\textbf{34.38}	&28.11	&44.61	&\cellcolor[gray]{0.9}35.70	&24.64	&20.75	&46.59	&\cellcolor[gray]{0.9}30.66	&30.38	&30.99	&48.09	&\cellcolor[gray]{0.9}36.48	&34.16	&33.55	&26.85	&\cellcolor[gray]{0.9}\textbf{31.52} \\ 

        FedVPT \cite{vpt} &32.75	&29.75	&48.64	&\cellcolor[gray]{0.9}37.05	&18.45	&30.56	&50.58	&\cellcolor[gray]{0.9}33.20	&18.93	&30.67	&48.18	&\cellcolor[gray]{0.9}32.59	&16.59	&33.08	&28.57	&\cellcolor[gray]{0.9}26.08 \\

        FedCLIP \cite{fedclip} &\textbf{34.38}	&27.56	&52.87	&\cellcolor[gray]{0.9}38.27	&25.67	&26.31	&\textbf{58.73}	&\cellcolor[gray]{0.9}\textbf{36.90}	&24.81	&\textbf{36.70}	&\textbf{57.78}	&\cellcolor[gray]{0.9}\textbf{39.76}	&27.13	&\textbf{38.61}	&26.74	&\cellcolor[gray]{0.9}30.83 \\ 

        \method (Ours) &33.90	&\textbf{33.08}	&\textbf{55.34}	&\cellcolor[gray]{0.9}\textbf{40.77}	&26.78	&\textbf{33.10}	&56.16	&\cellcolor[gray]{0.9}38.68	&23.64	&34.39	&55.67	&\cellcolor[gray]{0.9}37.90	&25.43	&34.28	&\textbf{31.32}	&\cellcolor[gray]{0.9}30.34 \\ 
    \bottomrule
    \end{tabular}}
    \label{tab:terra_ssmt}
\end{table*}

\begin{table*}[!ht]
    \centering
    \caption{\textbf{Comparison of methods on the Single-source Multi-target (SSMT) Domain Generalization setting}. The results are reported for the DomainNet dataset. Here Cl, Ig, Pt, Qd, Re, Sk denote to Clipart, Infograph, Panting, Quickdraw, Real and Sketch domains respectively.}
    \scalebox{0.65}{
    \begin{tabular}{lccccc|c||ccccc|c||ccccc|c}
    \toprule
        & \multicolumn{6}{c}{\textbf{Cl}} & \multicolumn{6}{c}{\textbf{Ig}} & \multicolumn{6}{c}{\textbf{Pt}} \\ \cmidrule(lr){2-7} \cmidrule(lr){8-13} \cmidrule(lr){14-19}
        
        \multirow{-2}{*}{\textbf{Method}} & {Ig} & {Pt} & {Qd} & {Re} & {Sk} &{Avg.} & {Cl} & {Pt} & {Qd} & {Re} & {Sk} &{Avg.} & {Cl} & {Ig} & {Qd} & {Re} & {Sk} &{Avg.} \\ 
        \midrule

        ZS-CLIP \cite{radford2019language} &45.94	&66.27	&14.19	&83.22	&62.25	&\cellcolor[gray]{0.9}54.37	&70.88	&66.27	&14.19	&83.22	&62.25	&\cellcolor[gray]{0.9}59.36	&70.88	&45.94	&14.19	&83.22	&62.25	&\cellcolor[gray]{0.9}55.30 \\ 
        \midrule
        
         FedOTP \cite{fedotp} &30.68	&44.32	&8.68	&58.81	&46.08	&\cellcolor[gray]{0.9}37.71	&51.16	&42.96	&7.56	&57.81	&45.28	&\cellcolor[gray]{0.9}40.95	&52.28	&30.93	&8.20	&59.04	&46.35	&\cellcolor[gray]{0.9}39.36 \\

         FedCoCoOp \cite{cocoop} &50.51	&67.71	&15.62	&83.81	&64.72	&\cellcolor[gray]{0.9}56.47	&71.87	&66.02	&14.56	&84.15	&63.81	&\cellcolor[gray]{0.9}60.08	&71.36	&48.02	&13.04	&82.76	&63.78	&\cellcolor[gray]{0.9}55.79 \\ 

        FedTPG \cite{fedtpg} &50.02	&67.06	&12.70	&84.24	&63.31	&\cellcolor[gray]{0.9}55.47	&71.92	&67.36	&13.12	&84.26	&63.79	&\cellcolor[gray]{0.9}60.09	&72.58	&50.12	&12.55	&84.24	&63.44	&\cellcolor[gray]{0.9}56.59 \\ 

        PromptFL \cite{promptfl} &50.27	&67.48	&15.60	&83.55	&64.52	&\cellcolor[gray]{0.9}56.28	&71.59	&65.95	&14.25	&83.89	&63.62	&\cellcolor[gray]{0.9}59.86	&71.52	&47.91	&14.67	&82.84	&63.71	&\cellcolor[gray]{0.9}56.13 \\ 

        FedKgCoOp \cite{kgcoop} &50.35	&67.52	&15.68	&83.70	&64.61	&\cellcolor[gray]{0.9}56.37	&71.72	&65.98	&14.27	&83.94	&63.67	&\cellcolor[gray]{0.9}59.92	&71.58	&47.78	&14.52	&82.88	&63.62	&\cellcolor[gray]{0.9}56.08\\

        FedMaPLe \cite{maple} &50.35	&66.69	&15.74	&82.99	&63.59	&\cellcolor[gray]{0.9}55.87	&71.37	&66.51	&15.74	&83.15	&63.67	&\cellcolor[gray]{0.9}60.09	&71.70	&48.82	&14.30	&82.79	&62.41	&\cellcolor[gray]{0.9}56.00 \\ 

        FedVPT \cite{vpt} &47.65	&66.50	&14.94	&83.65	&63.01	&\cellcolor[gray]{0.9}55.15	&71.41	&66.36	&14.63	&80.45	&60.70	&\cellcolor[gray]{0.9}58.71	&67.46	&42.84	&12.57	&80.82	&60.40	&\cellcolor[gray]{0.9}52.82 \\

        FedCLIP \cite{fedclip} &45.72	&66.62	&14.83	&83.17	&64.42	&\cellcolor[gray]{0.9}54.95	&71.34	&66.63	&14.81	&83.17	&62.42	&\cellcolor[gray]{0.9}59.67	&71.37	&45.72	&14.82	&83.17	&62.41	&\cellcolor[gray]{0.9}55.50 \\ 

        \method (Ours) &\textbf{51.26}	&\textbf{69.72}	&\textbf{18.96}	&\textbf{86.53}	&\textbf{65.92}	&\cellcolor[gray]{0.9}\textbf{58.48}	&\textbf{73.47}	&\textbf{69.74}	&\textbf{18.25}	&\textbf{85.94} &\textbf{65.48}	&\cellcolor[gray]{0.9}\textbf{62.58}	&\textbf{74.00}	&\textbf{51.38}	&\textbf{18.95}	&\textbf{85.73}	&\textbf{64.78}	&\cellcolor[gray]{0.9}\textbf{58.97} \\ 
    \midrule

    \midrule
        & \multicolumn{6}{c}{\textbf{Qd}} & \multicolumn{6}{c}{\textbf{Re}} & \multicolumn{6}{c}{\textbf{Sk}} \\ \cmidrule(lr){2-7} \cmidrule(lr){8-13} \cmidrule(lr){14-19}
        
        \multirow{-2}{*}{\textbf{Method}} & {Cl} & {Ig} & {Pt} & {Re} & {Sk} &{Avg.} & {Cl} & {Ig} & {Pt} & {Qd} & {Sk} &{Avg.} & {Cl} & {Ig} & {Pt} & {Qd} & {Re} &{Avg.} \\ 
        \midrule

        ZS-CLIP \cite{radford2019language} &70.88	&45.94	&66.27	&83.22	&62.25	&\cellcolor[gray]{0.9}65.71	&70.88	&45.94	&66.27	&14.19	&62.25	&\cellcolor[gray]{0.9}51.91	&70.88	&45.94	&66.27	&14.19	&83.22	&\cellcolor[gray]{0.9}56.10 \\ 
        \midrule
        
         FedOTP \cite{fedotp} &48.55	&29.38	&40.06	&55.21	&43.48	&\cellcolor[gray]{0.9}43.34	&51.93	&29.53	&43.81	&8.52	&45.23	&\cellcolor[gray]{0.9}35.80	&52.54	&30.88	&44.26	&8.45	&59.03	&\cellcolor[gray]{0.9}39.03 \\ 

         FedCoCoOp \cite{cocoop} &71.45	&49.26	&67.34	&81.87	&63.67	&\cellcolor[gray]{0.9}66.72	&71.66	&48.70	&67.29	&13.04	&63.78	&\cellcolor[gray]{0.9}52.89	&72.34	&49.98	&67.50	&14.89	&83.96	&\cellcolor[gray]{0.9}57.73 \\ 

        FedTPG \cite{fedtpg} &71.52	&50.00	&67.87	&83.47	&64.64	&\cellcolor[gray]{0.9}67.50	&72.48	&49.75	&66.97	&12.26	&62.88	&\cellcolor[gray]{0.9}52.87	&72.81	&50.14	&67.41	&13.00	&84.22	&\cellcolor[gray]{0.9}57.52 \\ 

        PromptFL \cite{promptfl} &71.29	&49.01	&67.24	&82.09	&63.76	&\cellcolor[gray]{0.9}66.68	&71.47	&48.96	&67.14	&13.25	&63.54	&\cellcolor[gray]{0.9}52.87	&72.16	&50.41	&67.35	&15.30	&83.87	&\cellcolor[gray]{0.9}57.82 \\ 

        FedKgCoOp \cite{kgcoop} &71.33	&49.15	&67.13	&82.14	&63.85	&\cellcolor[gray]{0.9}66.72	&71.52	&48.87	&67.19	&13.34	&63.65	&\cellcolor[gray]{0.9}52.91	&72.08	&50.49	&67.42	&15.37	&83.74	&\cellcolor[gray]{0.9}57.82\\ 

        FedMaPLe \cite{maple} &68.63	&45.32	&61.38	&78.03	&59.69	&\cellcolor[gray]{0.9}62.61	&72.07	&49.42	&67.60	&12.63	&63.24	&\cellcolor[gray]{0.9}52.99	&\textbf{72.96}	&49.64	&68.09	&16.21	&83.75	&\cellcolor[gray]{0.9}58.13 \\ 

        FedVPT \cite{vpt} &71.70	&44.28	&60.73	&79.44	&62.83	&\cellcolor[gray]{0.9}63.80	&66.58	&44.78	&61.22	&15.29	&60.57	&\cellcolor[gray]{0.9}49.69	&67.27	&46.37	&61.00	&12.78	&81.68	&\cellcolor[gray]{0.9}53.82 \\

        FedCLIP \cite{fedclip} &71.45	&45.78	&66.64	&83.20	&62.48	&\cellcolor[gray]{0.9}65.91	&71.32	&45.73	&66.61	&14.83	&62.43	&\cellcolor[gray]{0.9}52.18	&71.37	&45.72	&66.59	&14.81	&83.18	&\cellcolor[gray]{0.9}56.34 \\

        \method (Ours) &\textbf{73.69}	&\textbf{51.90}	&\textbf{68.72}	&\textbf{84.50}	&\textbf{65.14}	&\cellcolor[gray]{0.9}\textbf{68.79}	&\textbf{72.57}	&\textbf{51.72}	&\textbf{68.91}	&\textbf{18.59}	&\textbf{65.28}	&\cellcolor[gray]{0.9}\textbf{55.41}	&72.70	&\textbf{50.63}	&\textbf{68.37}	&\textbf{18.29}	&\textbf{85.59}	&\cellcolor[gray]{0.9}\textbf{59.12} \\
    \bottomrule
    \end{tabular}}
    \label{tab:domainnet_ssmt}
\end{table*}

\subsection{Detailed ablation studies}

\paragraph{Number of shots of images.} In Fig. \ref{fig:shots}, we present the performance of \method compared FL baselines across varying numbers of shots (or images) in base-to-new and MSST DG settings. Both of the results clearly demonstrate that FedMVP consistently outperforms others at all shot levels. Interestingly, even with as few as 2 samples per class, \method can outperform the baselines with four times the data, indicating better data efficiency.

\begin{figure}[h]
    \centering
        \includegraphics[width=0.7\linewidth]{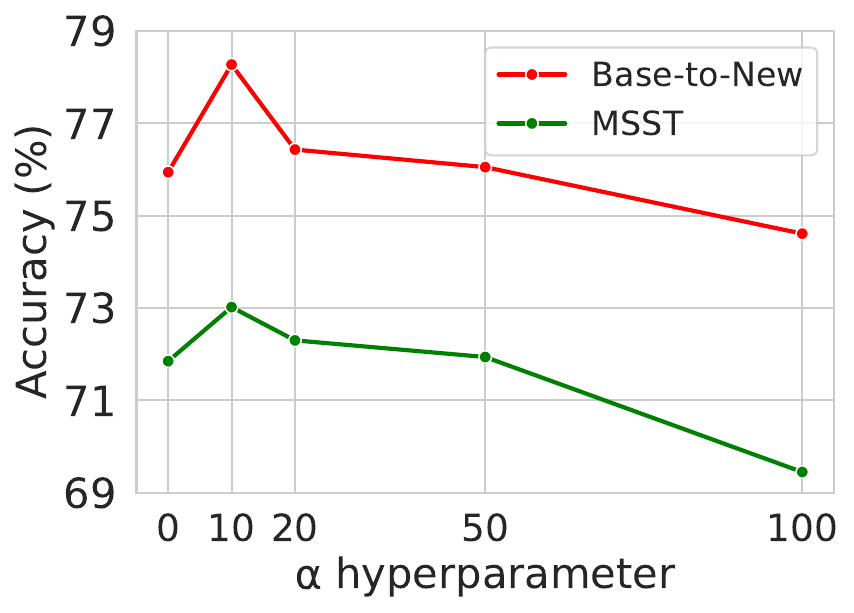}
        \caption{\textbf{Sensitivity to $\alpha$ hyperparamer in our \method} on Base-to-New and MSST DG (DomainBed) setting. }
        \label{fig:alpha}
    \vspace{-0.6cm}  
\end{figure}

\paragraph{Sensitivity to $\alpha$ hyperparamer in our \method.} In Fig. \ref{fig:alpha}, we demonstrate that maintaining a constant value of $\alpha = 10$ yields consistent results across both the \bton and MSST DG tasks. It is evident from the plot that as the value of $\alpha$ increases, the influence of the cross-entropy loss term, $\mathcal{L}_{ce}$, diminishes. This reduction in influence ultimately leads to less accurate backpropagation of loss functions for classification task, highlighting the delicate balance between $\alpha$ and the model's performance.

\paragraph{Effect of LLMs in our \method.} We evaluate the performance of our \method using four different large language models (LLMs), namely Llama-3.2-3B \cite{llama3}, Qwen2.5-14B \cite{qwen2}, Phi-4 \cite{phi4}, and GPT-4o \cite{gpt4}, and present their results in Table \ref{tab:llm_effect} for the MSST DG settings on both the PACS and OfficeHome datasets. The results demonstrate that GPT-4o outperforms the other models, achieving the highest performance overall. This superior performance can be attributed to its ability to capture more accurate and nuanced feature descriptions. Interestingly, all of the evaluated LLMs significantly outperform the second-best competitor, FedCLIP, on the MSST DG task. This highlights the importance of detailed feature representations in the success of our \method, further emphasizing the value of precise feature description for enhancing the performance.

\begin{table*}[!ht]
    \centering
    \caption{\textbf{Comparison of effects of different LLMs used in \method on the Multi-source Single-target (MSST) Domain Generalization setting}. The results are reported for the PACS and OfficeHome datasets.}
    \scalebox{0.75}{
    \begin{tabular}{lcccc|c||cccc|c}
    \toprule
        & \multicolumn{5}{c}{\textbf{PACS}} & \multicolumn{5}{c}{\textbf{OfficeHome}} \\ \cmidrule(lr){2-6} \cmidrule(lr){7-11}
        
        \multirow{-2}{*}{\textbf{Method}}  & {A. Painting} & {Cartoon} & {Photo} & {Sketch} & {Average} & {Art} & {Clipart} & {Product} & {RealWorld} & {Average} \\ 
        \midrule         
       
        Llama-3.2-3B \cite{llama3} &96.34	&98.92	&99.67	&92.21	&\cellcolor[gray]{0.9}96.79	&81.84	&69.94	&\textbf{91.84}	&92.30	&\cellcolor[gray]{0.9}83.98 \\

        Qwen2.5-14B \cite{qwen2} &96.20	&99.04	&100.00	&\textbf{93.08}	&\cellcolor[gray]{0.9}97.08	&82.05	&69.80	&91.63	&92.18	&\cellcolor[gray]{0.9}83.92 \\

        Phi-4 \cite{phi4}  &96.67	&98.78	&99.46	&92.17	&\cellcolor[gray]{0.9}96.77	&82.04	&70.02	&91.54	&92.45	&\cellcolor[gray]{0.9}84.01 \\
       
        GPT-4o \cite{gpt4} &96.92	&\textbf{99.35}	&\textbf{100.00} &92.86	&\cellcolor[gray]{0.9}\textbf{97.28} &\textbf{82.20}	&\textbf{70.05}	&91.78	&\textbf{92.56}	&\cellcolor[gray]{0.9}\textbf{84.15} \\
    \bottomrule
        
    \end{tabular}}
    \label{tab:llm_effect}
\end{table*}

\paragraph{Superiority of \method in non-federated offline setting.} In Table \ref{tab:offline_bton}, we present the performance of \method, demonstrating its potential to significantly improve vision-language alignment, even in non-federated offline settings. For comparison, we include a range of state-of-the-art methods, such as zero-shot CLIP (ZS-CLIP)~\cite{radford2019language} and several recent non-federated prompt learning techniques, including CoOp \cite{coop}, CoCoOp \cite{cocoop}, VPT \cite{vpt}, KgCoOp \cite{kgcoop}, MaPLe \cite{maple}, PromptSRC \cite{promptsrc}, StyLIP \cite{stylip}, CoPrompt \cite{coprompt}, TCP \cite{tcp}, DePT \cite{dept}, and DeKgTCP \cite{dekgtcp}. These methods are evaluated within the \bton setting. Our findings reveal that while DePT achieves the highest performance on the base classes, our \method outperforms all the competitors when it comes to unseen new classes, as well as the harmonic mean across both base and new classes. This highlights the robustness of \method, which not only demonstrates a reduced tendency to overfit but also exhibits superior adaptability to unseen samples during inference. These results underscore the effectiveness of \method in generalizing across both known and novel data distributions, making it a promising approach for real-world vision-language tasks.

\begin{table*}[!ht]
    \centering
    \caption{\textbf{Comparison of methods on the Base-to-new generalization task in non-federated offline setting.}}
    \scalebox{0.7}{
    \begin{tabular}{lc|cccccccccccc|c}
    \toprule
        & &\multicolumn{1}{c}{CLIP} &\multicolumn{1}{c}{CoOp} &\multicolumn{1}{c}{CoCoOp} &\multicolumn{1}{c}{VPT} &\multicolumn{1}{c}{KgCoOp} &\multicolumn{1}{c}{MaPLe} &\multicolumn{1}{c}{PromptSRC}	&\multicolumn{1}{c}{StyLIP} &\multicolumn{1}{c}{CoPrompt}	&\multicolumn{1}{c}{TCP} &\multicolumn{1}{c}{DePT}	&DeKgTCP & FedMVP \\
        
        \multirow{-2}{*}{\textbf{Method}}  & \multirow{-2}{*}{\textbf{Sets}} &ICML21 &IJCV22 &CVPR22 &ECCV22 &CVPR23 &CVPR23 &ICCV23 &WACV24 &ICLR24 &CVPR24 &CVPR24 &ICLR25& - \\ 
        \midrule

        &Base	&69.34	&82.69	&80.47	&82.11	&80.73	&82.28	&84.12	&83.22	&84.00	&84.13	&\textbf{85.18}	&84.96	&85.00 \\ 
        
        &New	&74.22	&63.22	&71.69	&71.73	&73.61	&75.14	&75.02	&75.94	&77.23	&75.36	&76.17	&76.38	&\textbf{77.85} \\ 
        
         \multirow{-3}{*}{\textbf{Average}} &H	&71.70	&71.66	&75.83	&76.57	&77.00	&78.55	&79.31	&79.41	&80.48	&79.51	&80.42	&80.44	&\textbf{81.27}  \\
         \midrule

         &Base	&72.43	&76.47	&75.98	&75.90	&75.83	&76.66	&77.75	&77.15	&77.67	&77.27	&\textbf{78.20}	&77.40	&78.11 \\
         
         &New	&68.14	&67.88	&70.43	&68.10	&69.96	&70.54	&70.70	&71.34	&71.27	&69.87	&70.27	&69.20	&\textbf{72.26} \\
         
         \multirow{-3}{*}{\textbf{ImageNet}} &H	&70.22	&71.92	&73.10	&71.79	&72.78	&73.47	&74.06	&74.13	&74.33	&73.38	&74.02	&73.07	&\textbf{75.07} \\
         \midrule

         &Base	&96.84	&98.00	&97.96	&98.03	&97.72	&97.74	&98.13	&98.23	&98.27	&98.23	&98.57	&\textbf{98.64}	&98.48 \\
         
         &New	&94.00	&89.81	&93.81	&94.30	&94.39	&94.36	&93.90	&94.91	&94.90	&94.67	&94.10	&\textbf{95.20}	&94.43 \\
         
         \multirow{-3}{*}{\textbf{Caltech101}} &H	&95.40	&93.73	&95.84	&96.13	&96.03	&96.02	&95.97	&96.54	&96.55	&96.42	&96.28	&\textbf{96.89}	&96.41 \\
         \midrule

         &Base	&91.17	&93.67	&95.20	&95.13	&94.65	&95.43	&95.50	&\textbf{95.96}	&95.67	&94.67	&95.43	&94.47	&95.70 \\
            
         &New	&97.26	&95.29	&97.69	&96.47	&97.76	&97.76	&97.40	&98.14	&98.10	&97.20	&97.33	&97.76	&\textbf{98.45} \\
        
	    \multirow{-3}{*}{\textbf{OxfordPets}} &H	&94.12	&94.47	&96.43	&95.80	&96.18	&96.58	&96.44	&97.04	&96.87	&95.92	&96.37	&96.09	&\textbf{97.06} \\
        \midrule

        &Base	&63.37	&78.12	&70.49	&71.63	&71.76	&72.94	&78.40	&75.19	&76.97	&80.80	&80.80	&\textbf{81.18}	&80.95 \\
        
        &New	&74.89	&60.40	&73.59	&72.20	&75.04	&74.00	&74.73	&74.46	&74.40	&74.13	&\textbf{75.00}	&74.75	&74.67 \\
        
        \multirow{-3}{*}{\textbf{StanfordCars}} &H	&68.65	&68.13	&72.01	&71.92	&73.36	&73.47	&75.52	&74.82	&75.66	&77.32	&77.79	&\textbf{77.83}	&77.68 \\
        \midrule

        &Base	&72.08	&97.60	&94.87	&95.93	&95.00	&95.92	&97.90	&96.54	&97.27	&97.73	&98.40	&\textbf{98.58}	&98.51 \\
        
        &New	&77.80	&59.67	&71.75	&70.37	&74.73	&72.46	&76.77	&73.08	&76.60	&75.57	&77.10	&75.18	&\textbf{78.76} \\
        
        \multirow{-3}{*}{\textbf{Flowers102}} &H	&74.83	&74.06	&81.71	&81.18	&83.65	&82.56	&86.06	&83.19	&85.71	&85.23	&86.46	&85.30	&\textbf{87.53} \\
        \midrule

        &Base	&90.10	&88.33	&90.70	&89.80	&90.50	&90.71	&90.63	&91.20	&90.73	&90.57	&90.87	&90.73	&\textbf{91.35} \\
        
        &New	&91.22	&82.26	&91.29	&90.37	&91.70	&92.05	&91.50	&92.48	&92.07	&91.37	&91.57	&91.55	&\textbf{93.04} \\
        
        \multirow{-3}{*}{\textbf{Food101}} &H	&90.66	&85.19	&90.99	&90.08	&91.09	&91.38	&91.06	&91.84	&91.40	&90.97	&91.22	&91.14	&\textbf{92.19} \\
        \midrule

        &Base	&27.19	&40.44	&33.41	&35.90	&36.21	&37.44	&42.30	&37.65	&40.20	&41.97	&\textbf{45.70}	&45.20	&42.38 \\
        
        &New	&36.29	&22.30	&23.71	&30.37	&33.55	&35.61	&36.97	&35.93	&39.33	&34.43	&36.73	&35.09	&\textbf{39.82} \\
        
        \multirow{-3}{*}{\textbf{FGVCAircraft}} &H	&31.09	&28.75	&27.74	&32.90	&34.83	&36.50	&39.46	&36.77	&39.76	&37.83	&40.73	&39.51	&\textbf{41.06} \\
        \midrule

        &Base	&69.36	&80.60	&79.74	&79.50	&80.29	&80.82	&82.83	&82.12	&82.63	&82.63	&83.27	&82.52	&\textbf{83.41} \\
        
        &New	&75.35	&65.89	&76.86	&76.17	&76.53	&78.70	&79.00	&79.95	&80.03	&78.20	&78.97	&78.30	&\textbf{79.50} \\
        
        \multirow{-3}{*}{\textbf{SUN397}} & H	&72.23	&72.51	&78.27	&77.80	&78.36	&79.75	&80.87	&81.02	&81.31	&80.35	&81.06	&80.35	&\textbf{81.41} \\
        \midrule

        &Base	&53.24	&79.44	&77.01	&80.90	&77.55	&80.36	&82.60	&81.57	&83.13	&82.77	&\textbf{84.80}	&83.80	&83.28 \\
        
        &New	&59.90	&41.18	&56.00	&52.73	&54.99	&59.18	&57.50	&61.72	&\textbf{64.73}	&58.07	&61.20	&59.66	&61.94 \\
        
        \multirow{-3}{*}{\textbf{DTD}} &H	&56.37	&54.24	&64.85	&63.85	&64.35	&68.16	&67.80	&70.27	&\textbf{72.79}	&68.25	&71.09	&69.70	&71.04 \\
    \midrule
													
        &Base	&56.48	&92.19	&87.49	&95.83	&85.64	&94.07	&92.40	&\textbf{94.61}	&94.60	&91.63	&93.23	&94.02	&94.33 \\
        
        &New	&64.05	&54.74	&60.04	&65.03	&64.34	&73.23	&68.43	&74.06	&78.57	&74.73	&77.90	&\textbf{81.69}	&81.59 \\
        
        \multirow{-3}{*}{\textbf{EuroSAT}} &H	&60.03	&68.69	&71.21	&77.48	&73.48	&82.30	&78.63	&83.08	&85.84	&82.32	&84.88	&87.42	&\textbf{87.50} \\
    \midrule
	
       &Base	&70.53	&84.69	&82.33	&84.63	&82.89	&83.00	&86.93	&85.19	&86.90	&87.13	&87.73	&88.06	&\textbf{88.46} \\
       
       &New	&77.50	&56.05	&73.45	&72.90	&76.67	&78.66	&78.33	&79.22	&79.57	&80.77	&77.70	&81.77	&\textbf{81.92} \\
       
       \multirow{-3}{*}{\textbf{UCF101}} &H	&73.85	&67.46	&77.64	&78.33	&79.65	&80.77	&82.41	&82.10	&83.07	&83.83	&82.46	&84.80	&\textbf{85.06} \\
    
    \bottomrule
        
    \end{tabular}}
    \label{tab:offline_bton}
\end{table*}

\paragraph{Effect of different prompting methods.} While using the multi-modal prompts for multi-modal prompting (\ie, through both vision and text encoder) is an interesting proposal, we find that it hurts performance (Table \ref{tab:multimodal}). We postulate that tuning prompts through both encoders is redundant as long as the prompts are multi-modal. While we expected the textual prompting with multi-modal prompts to work equally well as FedMVP, we find it further degrades performance. This can be attributed to overfitting, as shown in \cite{kgcoop}. In addition, we can cache the text representations once and backpropagate through the vision encoder alone in FedMVP, which is computation friendlier.

\end{document}